\documentclass[a4paper]{article}

\usepackage[pages=all, color=black, position={current page.south}, placement=bottom, scale=1, opacity=1, vshift=5mm]{background}
\SetBgContents{}      % copyright

\usepackage[margin=1in]{geometry} % full-width

\usepackage{amsmath}
\usepackage{amsthm}
\usepackage{amssymb}

\usepackage[utf8]{inputenc}

\usepackage{graphicx}
\usepackage{amsfonts}
\usepackage{amsmath}
\usepackage{bm}
\usepackage{mathtools}
\usepackage{caption}
\usepackage{subcaption}
\usepackage{booktabs}
\usepackage[misc]{ifsym}
\usepackage[T1]{fontenc}
\usepackage{makecell}
\usepackage{diagbox}
\usepackage{pifont}
\usepackage{colortbl}
\usepackage{algorithm}
\usepackage{algpseudocode}
\usepackage{pgffor}
\usepackage{xstring}
\usepackage{multirow}

\usepackage{amssymb}
\usepackage{latexsym}
\usepackage{marvosym}

\usepackage{siunitx}

\usepackage{url}
\usepackage{xcolor}

\usepackage{hyperref}

\definecolor{newcolor}{rgb}{.8,.349,.1}
\definecolor{Gray}{gray}{0.85}
\definecolor{LightGray}{gray}{0.95}

\newcolumntype{a}{>{\columncolor{Gray}}c}
\newcolumntype{b}{>{\columncolor{LightGray}}c}

\algnewcommand{\LineComment}[1]{\State \(\triangleright\) #1}

\usepackage{algorithm, algpseudocode} % use algorithm and algorithmicx for typesetting algorithms
\usepackage{mathrsfs} % for \mathscr command

\usepackage{lipsum}

\usepackage{orcidlink}

\definecolor{orcidlogocol}{HTML}{A6CE39}
\newcommand{\lforcid}{\orcidlink{https://orcid.org/0009-0003-3667-2001}}
\newcommand{\riorcid}{\orcidlink{https://orcid.org/0000-0003-4898-040X}}
\newcommand{\kyorcid}{\orcidlink{https://orcid.org/0000-0002-2318-1460}}

\makeatletter
\newcommand\footnoteref[1]{\protected@xdef\@thefnmark{\ref{#1}}\@footnotemark}
\makeatother

\title{TriDeNT \Neptune: Triple deep network training for privileged knowledge distillation in histopathology}
\author{Lucas Farndale\thanks{Co-corresponding authors: \{lucas.farndale,ke.yuan\}@glasgow.ac.uk}
 $^{,1,2,3,4}$ \lforcid \and Robert Insall$^{1,2,5}$ \riorcid \and Ke Yuan$^{*,1,2,3}$ \kyorcid}

\date{$^1$ School of Cancer Sciences, University of Glasgow \\
$^2$ Cancer Research UK Scotland Institute \\
$^3$ School of Computing Science, University of Glasgow \\
$^4$ School of Mathematics and Statistics, University of Glasgow \\
$^5$ Division of Biosciences, University College London \\}

\providecommand{\keywords}[1]{\textbf{\textit{Keywords:}} #1}

\begin{document}

\sisetup{detect-weight,mode=text}
% for avoiding siunitx using bold extended
\renewrobustcmd{\bfseries}{\fontseries{b}\selectfont}
\renewrobustcmd{\boldmath}{}
% abbreviation
\newrobustcmd{\B}{\bfseries}
\newrobustcmd{\U}{\underline}
\newcommand{\xmark}{\ding{55}}
\newcommand{\cmark}{\ding{51}}
\newcommand{\greycell}{\cellcolor{lightgray}}

\noexpandarg\exploregroups
\newcommand\ReplaceUnderscore[1]{\StrSubstitute{#1}{\_}{\_}}
\newcommand\ReplaceK[1]{\StrSubstitute{#1}{K}{k}}

\maketitle

\begin{abstract}

Computational pathology models rarely utilise data that will not be available for inference. This means most models cannot learn from highly informative data such as additional immunohistochemical (IHC) stains and spatial transcriptomics. We present TriDeNT \Neptune, a novel self-supervised method for utilising privileged data that is not available during inference to improve performance. We demonstrate the efficacy of this method for a range of different paired data including immunohistochemistry, spatial transcriptomics and expert nuclei annotations. In all settings, TriDeNT \Neptune ~outperforms other state-of-the-art methods in downstream tasks, with observed improvements of up to 101\%. Furthermore, we provide qualitative and quantitative measurements of the features learned by these models and how they differ from baselines. TriDeNT \Neptune ~offers a novel method to distil knowledge from scarce or costly data during training, to create significantly better models for routine inputs.

\end{abstract}

%% Keywords

% \linenumbers

%% main text

\begin{figure}[t]
    \centering
    \includegraphics[width=\textwidth]{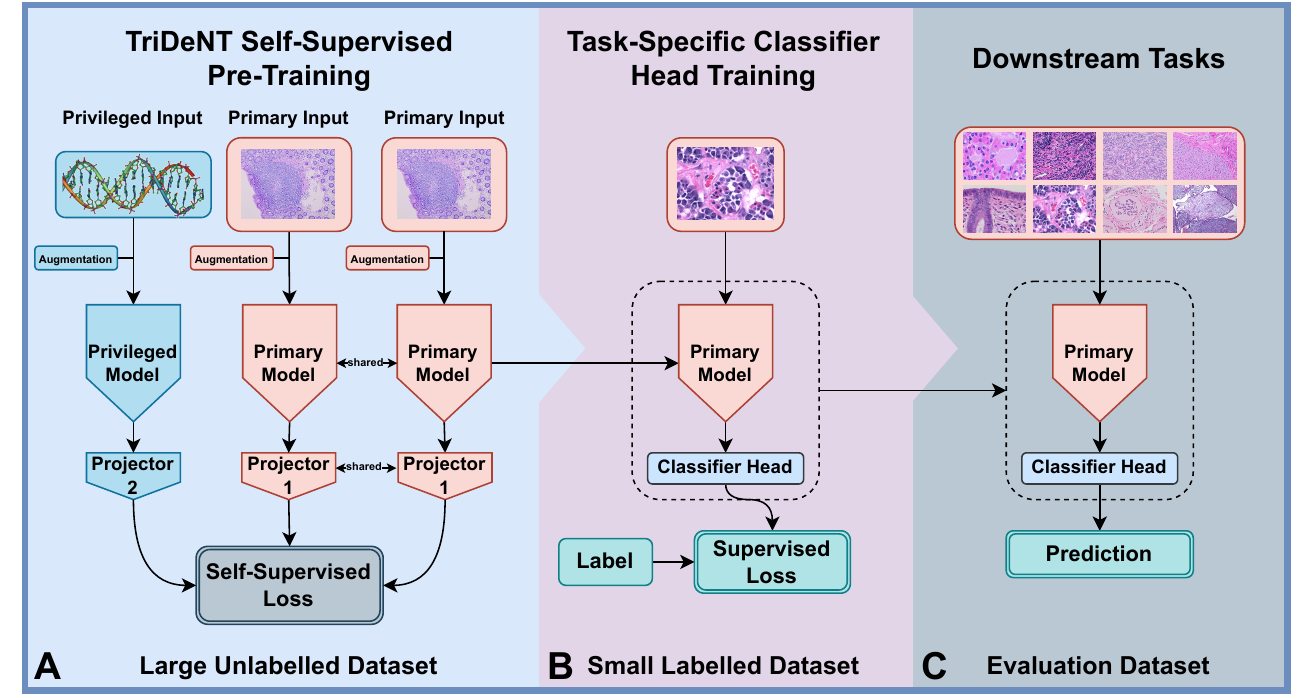}
    \caption{A: TriDeNT \Neptune ~architecture. TriDeNT \Neptune ~incorporates information from privileged input data to complement a primary data source. There are two encoder/projector pairs, one for the primary input (e.g. H\&E patches), and one for the secondary input (e.g. transcriptomics). The primary patches are augmented and passed to the primary encoder, followed by the projector, to output a representation. The privileged data are similarly passed to the privileged data encoder and projector. All representations are then used to calculate the self-supervised loss, which enforces invariance between representations. B: Classifier head training. Following this pre-training, the primary encoder is then used as a backbone for a downstream task, with a small classifier head appended. This is then trained in a supervised manner, requiring only a small amount of data. C: Use for downstream tasks. Finally, this trained model with a classifier head can be rolled out for use.}
    \label{fig:datapipeline}
\end{figure}

\section{Introduction}

Humans are able to easily transfer knowledge gained from studying one imaging technique into another. A clinician working with a rare type of histological staining who discovers a morphological change indicating disease will easily leverage this knowledge when they see the same change in routine stains from new patients, such as H\&E (Haematoxylin and Eosin)\footnote{See Figure \ref{tab:abbreviations} for a full list of abbreviations used in the text.} staining. For a deep learning model, this information would be useless, as it would have to be retrained from scratch for the new type of staining. There exist methods to enable deep learning algorithms to shift domains, however, the generality typically comes at the cost of performance on the primary domain \cite{rusu2016progressive}.

Deep learning approaches are quickly becoming pre-eminent in computational pathology, as they are able to make fast and accurate predictions at scale. Furthermore, research interest is beginning to turn to methods which do not require manual labelling of data, finding features in data without supervision. Despite this, deep learning models for pathology images are often only suitable for the task on which they were trained, and cannot meaningfully transfer to new domains without significant and costly re-training.  

Histology has long been the focus of a large amount of research attention in deep learning, and as a result there exist large datasets, such as TCGA (The Cancer Genome Atlas) \cite{tcga} and HTAN \cite{htan} containing data from routine examinations, such as H\&E stains, CT scans and x-rays. This has enabled very powerful models to be trained for these modalities, as these datasets are large, well-curated, and often cover many different demographics. What these datasets typically lack, however, is strong labels for most features present in these images. This means supervised models can only be trained on these large training datasets to predict slide-level labels such as survival, rather than clinically relevant features that require more extensive annotation. 

Despite not usually having strong labels, many datasets contain data from multiple sources and modalities, ranging from common techniques such as immunohistochemistry (IHC) \cite{Liu_2022_CVPR} to cutting-edge technologies such as super-resolution microscopy \cite{qiao2021evaluation}, spatial transcriptomics \cite{maniatis2019spatiotemporal}, and multiplex IHC \cite{ghahremani2023ai}. For example, studies using spatial transcriptomics typically also obtain H\&E stains alongside the genetic data. 

While models utilising multiple sources of data have been shown to be highly effective \cite{song2021effective, arevalo2017gated, kiela2014learning}, the abounding issue with these approaches is that obtaining additional data sources in practice is extremely prohibitive to the models' usefulness. State-of-the-art techniques are typically prohibitively expensive or impractical to be routinely used until long after their invention, and consequently their use is limited to a few research activities in exceptionally well-resourced labs. Even some more routine techniques, such as many immunohistochemical (IHC) stains, are arduous and expensive to obtain, register and align with existing data, and the available data will vary between samples or patients. It is therefore an important research direction to find methods which can use these additional privileged data sources during training to build better models of routine data, as these can be collected at scale and with fewer resources.

Deep learning offers an opportunity to improve patient outcomes and fundamental research using \emph{Learning Using Privileged Information} (LUPI) methods \cite{vapnik2009new}, which seek to improve performance by utilising additional data during training that is not available during inference. By training models to learn from \emph{privileged data}, we can develop models which make use of multiple sources to better analyse routine medical imaging without supervision. Furthermore, if they are available, manual annotations can be used as an additional input source for models to learn from during training without being restricted to only learn to output these annotations, as in supervised learning.

Primarily motivated by text/image retrieval tasks, there have been many LUPI methods developed. In general, these have been in supervised settings, however recently several unsupervised and self-supervised approaches have been developed. 

In some cases, existing unimodal self-supervised architectures have been shown to be amenable to LUPI, for example SimCLR \cite{chen2020simple}, Barlow Twins \cite{zbontar2021barlow} and VICReg \cite{bardes2021vicreg}, while others have been explicitly designed to cater to this problem setting, notably CLIP \cite{radford2021learning}, DeCLIP \cite{li2021supervision} and ALIGN \cite{jia2021scaling}, which use a contrastive objective similar to SimCLR to train models to predict the correct image/text pairings, and VSE++ \cite{faghri2017vse++} which uses hard-negative mining to improve representations. 

\begin{figure*}[p]
    \centering
    \begin{subfigure}{\textwidth}
    \centering
        \includegraphics[width=\textwidth]{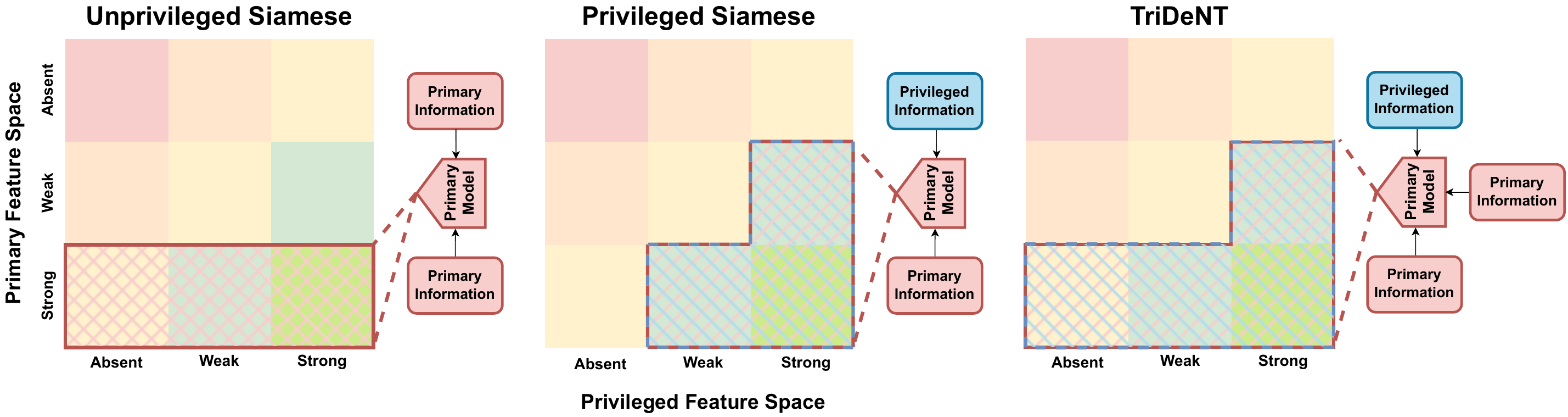}
        \caption{}
        \label{fig:latent_space}
    \end{subfigure}

    \begin{subfigure}{\textwidth}
    \centering
        \includegraphics[width=\textwidth]{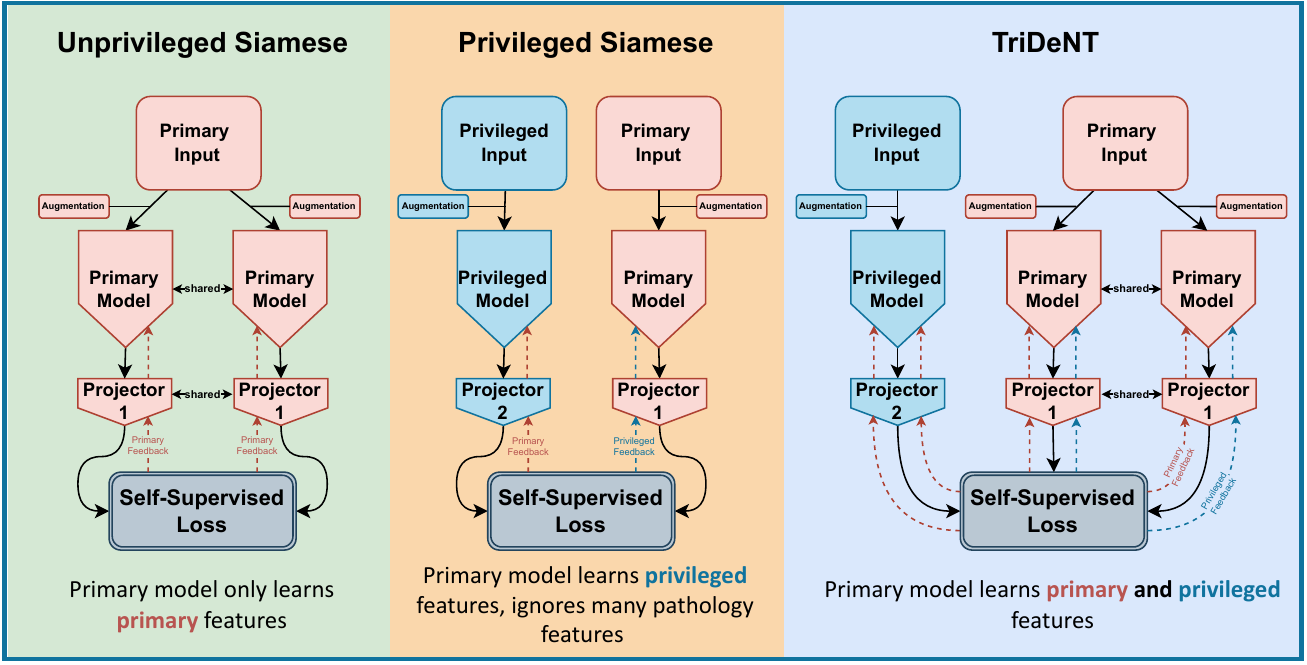}
        \caption{}
        \label{fig:siamese-methods}
    \end{subfigure}
    
    \caption{(a) Abstract description of the features which will be learned by different types of self-supervised models. The colour of the lines reflects the information being leveraged by privileged and unprivileged primary models. Features are either strongly present, weakly present, or absent in the primary and privileged data. Unprivileged Siamese models learn only features strongly present in the primary input, and are unlikely to learn any features which are only weakly present. Privileged models are likely to only learn features strongly or weakly present in both primary or privileged inputs. TriDeNT \Neptune ~combines the benefits of both methods to learn all features strongly present in the primary data, even those absent in the privileged data, while also learning features weakly present in the primary data that are strongly present in the privileged data. (b) Schematic for the learning process of these models. Black arrows indicate the forward flow of information through the network, and dashed lines indicate the signals which are received during backpropagation. Each branch effectively acts as a supervisory signal for the other branches, backpropagating feedback on the best features to learn. The primary model in the unprivileged Siamese setting only receives supervisory feedback from the primary data, so only learn primary features. Primary models in the privileged Siamese setting only receive supervisory feedback from the privileged data, so neglect many primary features. With TriDeNT \Neptune, primary models receive feedback from both data types, leading to features from both inputs being learned.}
    \label{fig:intro-figures}
\end{figure*}

Self-supervised LUPI training has been shown to be a highly effective method of improving the performance of models whose privileged data contains more task-relevant information than the primary data \cite{farndale2023more, girdhar2023imagebind}. These methods are designed to minimise the difference between the representations of each input by mapping all embeddings into a shared latent space (\emph{joint embedding}) \cite{lecun2022path}. However, in the case where we are only interested in the output of one branch, this can be restrictive. For example, if a feature is not shared between both inputs, these methods will neglect it, leading to worse performance \cite{farndale2023more} (Figure \ref{fig:latent_space}, see Section \ref{sec:privilegedinfo}). This was apparent in \cite{girdhar2023imagebind}, where, despite impressive retrieval performance, the proposed joint embedding model significantly underperformed supervised models on classification tasks, implying that important features were neglected in the primary domain. In this work we present TriDeNT \Neptune ~ (Figure \ref{fig:datapipeline}, Section \ref{sec:trident}), a new method designed to enable features which are only present in the primary data to be learned in addition to those shared between inputs. The main contributions of this work are:

\begin{itemize}
    \item We develop a new three-branch self-supervised model architecture, TriDeNT \Neptune, which utilises privileged information without compromising the features learned from the primary data;
    \item Using standard computational pathology tasks, we find that the previous state of the art -- standard Siamese self-supervised joint embedding architectures (e.g. \cite{girdhar2023imagebind}) -- embeds only information shared between views, meaning performance is reduced where not all task-relevant information is present in the privileged input;
    \item We show that TriDeNT \Neptune ~can incorporate features from additional stains, spatial transcriptomics, or nuclei annotations, for unprivileged downstream data analysis, and learns considerably more biologically relevant information from H\&E images.
\end{itemize}

\section{Methodology}
\label{sec:methodology}

\subsection{Self-Supervised Learning}
\label{sec:ssl}

In contrast to supervised learning which requires manual labels, and unsupervised learning which utilises task agnostic methods to find structure in data, \emph{self-supervised learning} (SSL) seeks to extract supervisory signals from data that it can leverage to produce meaningful representations of its inputs. These methods differ from unsupervised methods as they require manually engineered architectures for the specific data of interest, such as choosing appropriate data augmentations.

There are two types of SSL: i) generative, e.g., imputation of missing data, where the missing data provides the supervisory signal, such as a word masked from a sentence, and ii) discriminative, e.g., \emph{Siamese} models which map of multiple inputs into the same latent space, using the representation of each input as a supervisory signal for the other source. 

In the typical supervised setting, training consists of passing input/label pairs $(\bm{x}, \bm{y})$ to a model $\bm{y}=\psi(\bm{x})$ and optimising $\psi$ for some loss function comparing $y$ with a ground truth. Siamese self-supervised models instead take as inputs pairs $(\bm{x},\bm{x}')$, and use models $\bm{z}=\phi(\bm{x})$, $\bm{z}'=\phi'(\bm{x}')$ with the aim of minimising the difference $d(\bm{z},\bm{z}')$ between $\bm{z}$ and $\bm{z}'$. Typically this is implemented with $\phi=\phi'$, with $\bm{x}$ and $\bm{x}'$ both augmentations of the same input.

This method is amenable to a trivial constant solution where $\bm{z}=\bm{c}$ for some constant $c$ for all inputs. Therefore, Siamese methods require some regularisation to avoid such collapse. There are two approaches: contrastive and non-contrastive. Contrastive methods such as \cite{chen2020simple, radford2021learning, caron2020unsupervised} use both \emph{positive} (matching) and \emph{negative} (non-matching) pairs, and seek to pull together positive pairs while pushing apart negative pairs, either in embedding space \cite{chen2020simple,radford2021learning} or through cluster assignment \cite{caron2020unsupervised}. Non-contrastive methods such as \cite{zbontar2021barlow, bardes2021vicreg, chen2021exploring, grill2020bootstrap} instead use only positive pairs, and regularise the representations to avoid collapse by using architectural constraints such as momentum encoders \cite{grill2020bootstrap}, stop gradients \cite{chen2021exploring} and covariance constraints \cite{zbontar2021barlow, bardes2021vicreg}.

\subsection{Knowledge Distillation}

Knowledge distillation is the transfer of knowledge from a \emph{teacher} model to a \emph{student} model \cite{hinton2015distilling}, usually with the goal of either teaching a smaller student model to emulate the performance or learn some desirable property of the teacher model. This is usually achieved by passing the same input to both models, and minimising the difference between their outputs, using either their representations or possibly a projection head. The student model is usually smaller, and the teacher model may be pre-trained. The objective is to teach the student model to produce the same or a related output to the teacher, possibly using fewer resources.

In this work, we utilise this approach not to distil knowledge not on the level of the model, but on the level of the data. We consider settings where there are multiple sources of data about the same input, such as different histological stains from the same sample, and we wish to train models to classify only the primary data. By mapping both inputs into a joint embedding space, the models' objective is to produce the same representation for each, with the goal of improving the quality of representations of the primary data.

\subsection{Privileged Information}
\label{sec:privilegedinfo}

Privileged information is information which is available during training but not during inference. In a supervised setting this is defined, using the notation of Section \ref{sec:ssl}, as having data $(\bm{x}, \bm{x}^*, \bm{y})$ during training, and optimising a model $\bm{y}=\psi(\bm{x})$, which will then be used in inference without the privileged information $\bm{x}^*$. Most existing work on LUPI is focused on understanding supervised learning dynamics using support vector machines (SVMs), however, much of this has been extended to neural networks \cite{vapnik2017knowledge}, unsupervised learning \cite{feyereisl2012privileged,karaletsos2015bayesian}, and knowledge distillation \cite{lopez2015unifying}. The original framework \cite{vapnik2009new} was defined for SVMs, with the privileged information being used to estimate the slack values. The slack values can be understood equivalently for neural networks as loss values. This was leveraged by \cite{yang2017miml} to use privileged information to estimate loss values for a neural network for multiple instance learning, integrating privileged information at both the instance level and the bag level. Rather than directly using privileged information to inform predictions, \cite{lambert2018deep} use privileged information to determine dropout variance during training, leading to greater sample efficiency.

Despite the apparent advantage of providing privileged information to a model, it has been shown that training with privileged information does not satisfy a \emph{no-harm guarantee} \cite{lambert2018deep}. This can be due to a variety of factors, such as because estimating properties of the privileged information can be more difficult than estimating the same properties of the primary data. It was shown in \cite{farndale2023more} that Siamese LUPI leads to improved performance on tasks where the privileged input contains more task-relevant information than the primary input, e.g., a low-resolution image paired with a high-resolution image. However, it is also observed that if the privileged input contains less task-relevant information, it can reduce performance. This is because mapping both inputs into the same latent space causes task-relevant information in the primary input to be lost if it is not shared between branches, as is visualised in Figure \ref{fig:latent_space}. Consequently, non-LUPI learning can lead to better performance in these scenarios, despite the loss of additional task-relevant information which could be gained from an privileged input.

Extending the supervised setting to the Siamese self-supervised setting, we have inputs pairs $(\bm{x},\bm{x}^*)$, and use models $\bm{z}=\phi(\bm{x})$, $\bm{z}^*=\phi^*(\bm{x}^*)$ with only the model $\phi$ being used for inference. For example, we may have a set of H\&E images $\bm{x}$ and privileged paired IHC images $\bm{x}^*$ which are only available during training.

As Siamese joint-embedding models minimise the difference between representations in the shared embedding space, any features which are not shared between branches will be neglected. There is no way to predict a feature in the privileged input from the primary input if no information exists about that feature in the primary input. On the other hand, features which are weakly present in the primary input but strongly present in the privileged input may be learned, as there is a strong supervisory signal from the privileged data. In the non-LUPI setting (Siamese learning without privileged inputs), such features are unlikely to be learned due to the absence of the strong signal from the privileged input. Formally, following \cite{jing2021understanding}, we consider features which have variance that is nonzero but lower than the augmentation regime to be weakly present, and those with greater variance than the augmentation regime to be strongly present.

\subsection{TriDeNT \Neptune}
\label{sec:trident}

The goal of TriDeNT \Neptune ~method is to combine the benefits of both LUPI and non-LUPI methods in such a way that the primary encoder can make best use of signals from all inputs. We use a three-pronged approach, with two branches acting on the primary input and a third acting on the privileged input. Our method can be considered a generalisation of the standard Siamese self-supervised architecture. We take inputs $\bm{X}=(\bm{x}, \bm{x}^*)\in(\mathcal{X},\mathcal{X}^*)$, where we assume each input contains some information about their shared source. The inputs could represent any type of input array, such as images, -omics data, or patient information. We assume $\bm{x}^*$ contains some mutual information with $\bm{x}$. We aim to obtain representations $\bm{z},\bm{z}^*\in\mathcal{Z}$, such that the $\bm{z}$ is a \emph{sufficient} representation of $\bm{x}$ for some task $T$, that is to say we have mutual information $I(\bm{z};T)=I(\bm{x};T)$. Note that, in contrast to comparable approaches, we are only interested in optimising $\bm{z}^*$ insofar as this benefits $\bm{z}$, as only $\bm{z}$ is to be used for inference.

Inputs $\bm{x}, \bm{x}^*$ are augmented by stochastic operators
\begin{equation}
    a:\mathcal{X}\rightarrow\mathcal{X}, \quad a^*:\mathcal{X}^*\rightarrow\mathcal{X}^*
\end{equation}
respectively, and mapped to representations $\bm{z}^i\in\mathcal{Z}$ by encoders $f^i:\mathcal{X}^i\rightarrow\mathcal{Z}$ according to the rule
\begin{equation}
    \bm{z}^i=f^i(\hat{a}(\bm{x})), \quad i=1,2,*.
\end{equation}
We have defined $\hat{a}$ to be $a$ if its input is $\bm{x}$ and $a^*$ if its input is $\bm{x}^*$, as in general there is no reason for augmentations to be the same for primary and privileged data. This yields three representations, $\bm{z}^1, \bm{z}^2$, and $\bm{z}^*$, where $\bm{z}^1$ and $\bm{z}^2$ are representations of each augmentation of the primary data, and $\bm{z}^*$ is the representation of the privileged data. Representations are then mapped to embeddings $\bm{e}^i\in\mathcal{E}$ by a projector $g^i:\mathcal{Z}\rightarrow\mathcal{E}$ with the rule $\bm{e}^i=g^i(\bm{z}^i)$. In general we will have primary encoder $f=f^1=f^2$ and projector $g=g^1=g^2$.

Note that the spaces $\mathcal{Z}$ and $\mathcal{E}$ are not dependent on $i$, as these are shared latent spaces. Projections into an embedding space are used in keeping with existing approaches \cite{zbontar2021barlow, bardes2021vicreg, chen2020simple}, as this has been shown to improve generalisation and feature learning. For inference, augmentations are not applied, so $\hat{a}$ is set to $\hat{a}(x)=x$. In general, we use the same encoder for both branches taking $\bm{x}$ as input, as sharing weights has been shown to improve performance on unprivileged tasks \cite{farndale2023more}. Typically, we will have $\mathcal{Z}=\mathbb{R}^{n\times d}$ where $n$ is the batch size and $d$ is the dimension of the representation. For pseudocode, see Algorithm \ref{alg:pseudocode}.

\subsection{Objective Function}

Consider a setting where $N$ is the number of branches with the primary input and $M$ is the number of branches with the privileged input. We generalise a two-branch self-supervised loss $\mathcal{L}_2(\bm{z}^i,\bm{z}^j)$ to $N+M$ branches by summing over the losses between representations such that the $N+M$ branch loss is defined as 
\begin{equation}
\mathcal{L}_{N,M}(\bm{z}^1,\ldots,\bm{z}^N,\bm{z}^{*1},\ldots,\bm{z}^{*M})\coloneqq\sum_{i\neq j}^{N+M}\mathcal{L}_2(\bm{z}^i,\bm{z}^j).
\end{equation}
We investigate the case where $N=2$ and $M=1$ (giving three branches overall) however the method could easily be generalised to more branches. Siamese unprivileged learning is the case with $N=2$ and $M=0$, and Siamese privileged learning is the case with $N=M=1$. A short discussion of the use of additional privileged branches is presented in Appendix \ref{sec:additional-privileged-branches}, where we see that using more than one privileged branch could deteriorate performance. In the present work we use both contrastive and non-contrastive choices of $\mathcal{L}_2$ to demonstrate that TriDeNT \Neptune ~is robust to the choice of self-supervised loss. We illustrate this using the VICReg (Variance Invariance Covariance Regularisation) \cite{bardes2021vicreg} objective and the InfoNCE ([Mutual Information] Noise Contrastive Estimation) objective \cite{oord2018representation}, which have both been used extensively in self-supervised architectures (e.g. \cite{lee2022vnibcreg,chen2020simple,radford2021learning,girdhar2023imagebind}).

For brevity we focus only on architectures which can be summed in this way, although this setting can be easily extended to architectures requiring more complex structuring of their loss function. For example, self-predictive architectures such as BYOL \cite{grill2020bootstrap} and SimSiam \cite{chen2021exploring} would require designation of \emph{online} and \emph{target} branches and pairings between them.

\subsubsection{VICReg}

The VICReg objective is defined as
\begin{equation}
    \mathcal{L}_{VICReg}(\bm{z}^1, \bm{z}^2)\coloneqq\lambda s(\bm{z}^1, \bm{z}^2) + \sum_{i=1}^2\left[\mu v(\bm{z}^i) + \nu c(\bm{z}^i)\right]
\end{equation}
where $s(\cdot,\cdot)$ is an invariance regularisation term, $v(\cdot)$ is a variance regularisation term, and $c(\cdot)$ a covariance regularisation term, with $\bm{z}^i$ being the embedding of branch $i$, and $\lambda, \mu, \nu$ weighting coefficients. These functions are
\begin{align}
    s(\bm{z}^i, \bm{z}^j)\coloneqq & \frac{1}{n}\sum_{k=1}^{n}\lVert \bm{z}_k^{i}-\bm{z}_k^{j}\rVert_2^2,\\
    v(\bm{z}^i) \coloneqq & \frac{1}{D}\sum_{d=1}^{D}\max\left(0, \gamma-\sqrt{\mathrm{Var}([\bm{z}^i]_d)+\epsilon}\right),\\
    c(\bm{z}^i) \coloneqq & \frac{1}{D}\sum_{d\neq \delta}\left[C(\bm{z}^i)\right]_{d,\delta}^2,
\end{align}
where
\begin{equation}
    C(\bm{z}^i)\coloneqq \frac{1}{n-1}\sum_{k=1}^{n}(\bm{z}_k^i-\bar{\bm{z}}^{i})(\bm{z}_k^{i}-\bar{\bm{z}}^{i})^\mathrm{T}, \quad \bar{\bm{z}}^i \coloneqq \frac{1}{n}\sum_{k=1}^{n} \bm{z}_k^{i},
\end{equation}
$n$ is the batch size with $[z_a^i]_j\in\bm{z}^i$ being dimension $j$ of element $a$ in the batch of representations $\bm{z}^i$, $\gamma$ is a term determining the desired variance of the representations, $D$ is the dimension of the representation, and $\epsilon$ is a small constant to ensure numerical stability. Unlike many Siamese networks (e.g. \cite{chen2021exploring,grill2020bootstrap}) VICReg can admit distinct inputs and architectures on each branch. This is because both branches are regularised separately by the covariance term, and consequently has been shown to work better than VSE++ \cite{faghri2017vse++} and Barlow Twins \cite{zbontar2021barlow} for multi-modal data \cite{bardes2021vicreg}.

Both the variance and covariance functions $v$ and $c$ are applied to each branch independently, meaning that the invariance between branches is achieved simply through the distance function $s$. In the original description, these functions were implemented with all parameters shared between both branches, but this is not a necessary restriction.

\subsubsection{InfoNCE}

We use the variant of the InfoNCE/NT-Xent/N-pairs losses used in SimCLR \cite{chen2020simple}, ImageBind \cite{girdhar2023imagebind}, etc., which is defined as
\begin{equation}
    \mathcal{L}_{InfoNCE}(\bm{z}^1, \bm{z}^2)\coloneqq \frac{1}{2n}\sum_{i=1}^{n}\left(l(\bm{z}_i^1, \bm{z}_i^2)+l_i(\bm{z}_i^2, \bm{z}_i^1)\right)
\end{equation}
with
\begin{equation}
    l(\bm{z}_i^a, \bm{z}_i^b)\coloneqq -\log\frac{\exp{(\mathrm{sim}(\bm{z}_i^a,\bm{z}_i^b)/\tau)}}{\sum_{k=1}^{n} \exp{(\mathrm{sim}(\bm{z}_i^a,\bm{z}_k^b)/\tau})},
\end{equation}
where $\mathrm{sim}(\cdot)$ is the cosine similarity, $\tau$ is a temperature parameter, and $n$ is the batch size.

\subsection{Primary and Privileged Features}

For an intuitive understanding of the method, it is helpful to consider the representation of each branch as a supervisory signal for the others. Our model can therefore be considered a multi-objective setting, where the primary encoder $f$ aims to balance the information extracted from each augmentation of $\bm{x}$ which is shared with $\bm{x}^*$, and that which is shared with the other augmentation of $\bm{x}$. In turn, the supervisory signals for $\bm{x}^*$ are $\bm{z}^1$ and $\bm{z}^2$, and consequently they will only extract features which can also be found in $\bm{x}$. In our typical setting, this corresponds to balancing information which is only weakly present in the primary input $\bm{x}$, but strongly present in the privileged input $\bm{x}^*$, with information which is strongly present in primary input $\bm{x}$. The result of this trade-off is that privileged features with a strong supervisory signal from $\bm{z}^1$ and $\bm{z}^2$ are learned, but primary features with a strong supervisory signal from from $\bm{z}^*$ are also learned. This is in contrast to the dichotomy between only learning strong features or only learning shared features presented by the standard 2-branch approaches.

\begin{table}[!p]
    \centering
    \caption{Datasets and Tasks}
    \label{tab:datasets}
    \scalebox{0.65}{\begin{tabular}{ll}
    \toprule
    \multicolumn{2}{c}{Training Datasets} \\
    \midrule
        \makecell[l]{Name: \textbf{SegPath}\\Reference: \cite{komura2023restaining}\\Tissue: Pan-Cancer \\Split: See Table \ref{tab:segpath-stains}\\Task: None} & \makecell[l]{Contains eight subsets of H\&E slides, each with a different paired IF stain (see Table \ref{tab:segpath-stains}). Features 1,583 patients and 18\\different tissue types, with no overlap between the H\&E images in each subset. IF images are only\\released as binarised images using a threshold value determined in the original study \cite{komura2023restaining}. Only used for training.} \\
    \midrule
        \makecell[l]{Name: \textbf{BCI}\\Reference: \cite{Liu_2022_CVPR}\\Tissue: Breast \\Split: 62,336/15,632\\Task: HER2 Status Prediction} & \makecell[l]{HER2 (Human Epidermal growth factor Receptor 2) is a protein which has been found to be prognostic for breast cancer.\\It is tested for using IHC staining, and classified into 4 grades (0, 1+, 2+, 3+). BCI contains paired H\&E/IHC\\patches from 51 breast cancer patients, which have been registered for precise correspondence between the H\&E/IHC patches.} \\
    \midrule
        \makecell[l]{Name: \textbf{PanNuke}\\Reference: \cite{gamper2019pannuke}\\Tissue: Pan-Cancer\\Split: 4295/2283\\Task: Neoplastic Cell Detection} & \makecell[l]{PanNuke contains H\&E patches paired with exhaustive nuclei segmentations from 19 tissue types. The associated task is\\neoplastic cell detection, following \cite{huang2023visual}, where the model must determine whether a patch\\contains an abnormal, excessive growth of tissue, whether benign or malignant.} \\
    \midrule
        \makecell[l]{Name: \textbf{ALS-ST}\\Reference: \cite{maniatis2019spatiotemporal}\\Tissue: Mouse/Human Spinal Cord\\Split: See Appendix \ref{sec:als-st-dataset}\\Task: Genotype Prediction \&\\White/Grey Matter Classification} & \makecell[l]{Dataset containing 80 human and 331 mouse spinal cord sections from 7 humans and 67 mice who have ALS, a\\neurodegenerative disease affecting the motor neurons. All samples feature a H\&E slide with matched and aligned spatial transcriptomics.\\ The tasks are to predict the mouse SOD1 genotype from SOD1-G93A (ALS),SOD1-WT (Wildtype), and Knockout, and to classify white\\matter and grey matter.} \\
    \midrule
    \multicolumn{2}{c}{Evaluation Datasets} \\
    \midrule
         \makecell[l]{Name: \textbf{NCT}\\Reference: \cite{kather_dataset}\\Tissue: Colorectal\\Split: 100,000/7,177\\Task: Tissue Classification} & \makecell[l]{Manually annotated patches of nine tissue types: adipose (ADI), background (BACK), debris (DEB), lymphocytes (LYM),\\mucus (MUC), smooth muscle (MUS), normal colon mucosa (NORM), cancer-associated stroma (STR), colorectal\\adenocarcinoma epithelium (TUM). Patches extracted from H\&E slides from 86/50 patients in the train/test sets respectively.\\This task assesses the models' ability to differentiate features which are primarily determined by the H\&E image, but can be\\enhanced by paired information, such as presence of immune cells helping classify lymphocytes.} \\
    \midrule
        \makecell[l]{Name: \textbf{Camelyon}\\Reference: \cite{bandi2018detection}\\Tissue: Lymph Node\\Split: 179,394/146,722\\Task: Out-of Distribution\\Metastasis Detection} & \makecell[l]{There is a large degree of variation between different scanners, staining protocols and sample collection methods, so H\&E images can\\look very different depending on how, when, and where they were collected. The WILDS distribution of Camelyon features 1399 breast\\lymph node whole slide images from 5 different hospitals, with centres 1,2, and 3 comprising the train set, 4 being the validation set,\\and 5 being the test set. There is a large difference between sets, so Camelyon assesses models' generalisation ability.} \\
    \midrule
        \makecell[l]{Name: \textbf{MHIST}\\Reference: \cite{wei2021petri}\\Tissue: Colorectal Polyps\\Split: 2175/977\\Task: Polyp Classification} & \makecell[l]{Features patches from 328 whole slide images of colorectal growths, \emph{polyps}, which can become cancerous.\\MHIST contains two classes: \emph{serrated} polyps, which can become cancerous, and \emph{hyperplastic} polyps, which are typically benign.\\Note that these images are at $8\times$ magnification, so this task assesses the models' generalisation performance across magnification scales.} \\
    \midrule
        \makecell[l]{Name: \textbf{Singapore}\\Reference: \cite{oner2022ai}\\Tissue: Prostate\\Split: 3843/4261\\Task: Prostate Gland\\Malignancy Classification} & \makecell[l]{Classification dataset with samples from 46 patients who underwent a prostate core needle biopsy. Patches are centred\\on a prostate gland labelled as benign or malignant. This task assesses the models' ability to make\\classifications based on a specific biological feature which was uncommon or absent during training.} \\
    \midrule
        \makecell[l]{Name: \textbf{TIL}\\Reference: \cite{kaczmarzyk10dataset}\\\cite{abousamra2022deep}\\\cite{saltz2018spatial}\\Tissue: Pan-Cancer\\Split: 209,221/56,275\\Task: Tumour Infiltrating\\ Lymphocyte Detection} & \makecell[l]{Features patches from 7983 whole slide images from 23 cancer types. The task is to detect \emph{tumour infiltrating lymphocytes} (TILs),\\which are an important biomarker for cancer prognosis, with increased TIL density being associated with positive clinical outcomes.\\Assesses models' ability to detect features which co-occur with more prominent labels such as tumour. Also assesses relative performance\\of different paired data, as immune-related paired data is far more relevant to performance than other stains.} \\
    \midrule
        \makecell[l]{Name: \textbf{PANDA}\\Reference: \cite{bulten2022artificial}\\Tissue: Prostate\\Split: 7962/2654\\Task: Prostate Biopsy\\ISUP Grading} & \makecell[l]{Large dataset featuring 10616 whole-slide images of prostate biopsies, weakly labelled with ISUP grades from 1 to 5. Results are\\reported as Cohen's $\kappa$. Assesses performance of models' representations for aggregated slide level predictions on a difficult,\\clinically relevant task.} \\
    \midrule
        \makecell[l]{Name: \textbf{IMP 1K/4K}\\Reference: \cite{Oliveira2021}\\\cite{Neto2022}\\\cite{Neto2024}\\Tissue: Colorectal\\Split: 1132(1K)/4433(4K)/900\\Task: Colorectal Dysplasia\\Detection} & \makecell[l]{Dataset with 1132 whole slide images (1K) from colorectal biopsies and polypectomies, with an extension to 4433 slides (4K). Labels\\are: non-neoplastic, low-grade lesions (conventional adenomas with low-grade dysplasia), and high-grade lesions (conventional\\adenomas with high-grade dysplasia, intra-mucosal carcinomas and invasive adenocarcinomas).} \\
    \midrule
        \makecell[l]{Name: \textbf{IMP Cervix}\\Reference: \cite{Oliveira2023}\\Tissue: Cervix\\Split: 480/120 \\Task: Cervical Dysplasia\\Detection} & \makecell[l]{Features 600 whole slide images of cervical Loop Electrosurgical Excision Procedure (LEEP) samples, with 4 classes: non-neoplastic,\\low-grade, and high-grade squamous intraepithelial lesion.} \\
    \bottomrule
    \end{tabular}
    }
\end{table}

\subsection{Datasets and Tasks}
\label{sec:main-datasets-and-tasks}

While H\&E staining is the routine protocol for tissue analysis, pathologists usually rely on IHC or IF staining to obtain information about the locations of individual proteins, which may aid further investigation or confirm their diagnoses. IHC and IF stains contain highly specific information about a particular protein, so add useful information beyond that which can be readily identified with H\&E staining. While this is necessary for human pathologists to identify features which cannot be easily identified by eye in the generic H\&E stains, for example separating the identities of individual epithelial, endothelial, myeloid cells and lymphocytes, it has been shown that neural networks can accurately reproduce many of these stains from only H\&E \cite{xu2019gan}. It would be tempting to conclude, then, that this is a solved problem, and researchers should simply use these models to produce representations of H\&E patches which contain features relevant to IHC or IF stains. Unfortunately, this has been shown to perform poorly \cite{farndale2023more,balestriero2024learning}, as image to image translation models are restricted to learning very fine grained information, such as the exact locations of nuclei, at the expense of the types of low-redundancy coarse grained features which are learned by self-supervised \emph{Siamese} networks. 

Nevertheless, the results of these works on image translation imply that much or all of the useful information in IHC and IF stains can be predicted from H\&Es. We investigate how representation are affected by distilling information from IHC and IF stains into models of H\&E stains, evaluating both brightfield IHC images and thresholded IF images. We also investigate whether distilling manual nuclei segmentations can improve performance, as this is a painstaking process for pathologists to perform, but there are now large datasets of (semi-) manually generated segmentations, with accurate models able to perform segmentation from H\&E stains. We finally investigate the use of spatial transcriptomics as privileged information, as this contains rich complementary information to H\&E stains, and is an example of how cutting-edge data could be distilled into models of routine data.

In Table \ref{tab:datasets} we provide a short overview of the datasets used in this work. For a full description, please see Appendix \ref{sec:datasets}.

\begin{figure*}[!p]
    \centering
    \begin{subfigure}{\textwidth}
    \centering
        \includegraphics[width=0.9\textwidth]{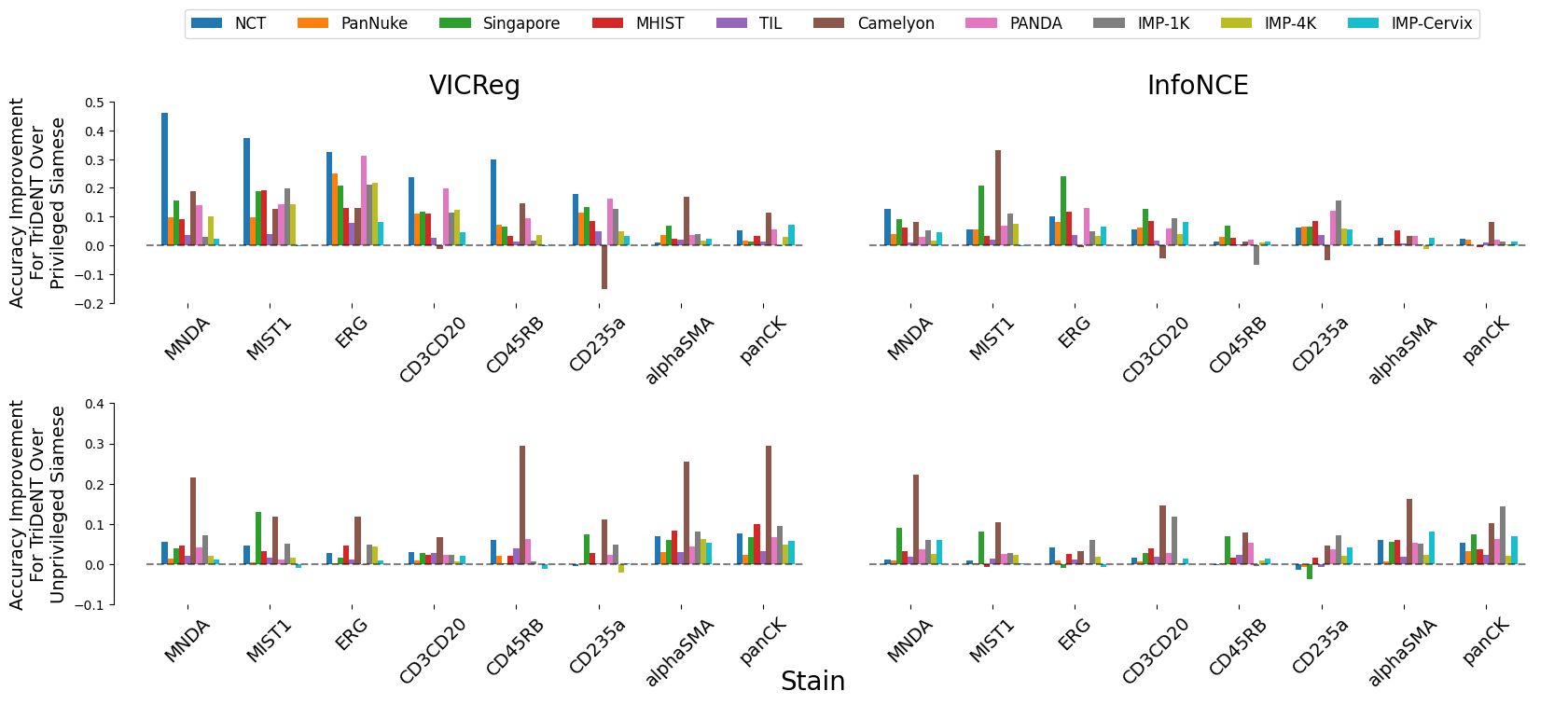}
        \caption{}
        \label{fig:segpath_res_tri_priv}
    \end{subfigure}

    % \begin{subfigure}{\textwidth}
    % \centering
    %     \includegraphics[width=0.9\textwidth]{fig/segpath_tri-unpriv.png}
    %     \caption{}
    %     \label{fig:segpath_res_tri_unpriv}
    % \end{subfigure}
    \begin{subfigure}{\textwidth}
    \centering
    \scalebox{0.9}{\begin{tabular}{ccccc|ccc|c}
    \toprule
        \multirow{3}{*}{\diagbox[height=3\line]{Task}{}}& Loss: & \multicolumn{3}{c}{VICReg} & \multicolumn{3}{c}{InfoNCE} & \multirow{3}{*}{\makecell{Supervised\\Baseline}}\\
        & Method: & \greycell TriDeNT \Neptune & \multicolumn{2}{c}{Siamese} & \greycell TriDeNT \Neptune & \multicolumn{2}{c}{Siamese} & \\
        & Privileged: & \greycell\cmark & \cmark & \xmark & \greycell\cmark & \cmark & \xmark \\
    \midrule
        NCT & & \greycell\B 0.9031 & 0.6618	& 0.8582 & \greycell\B 0.8979 & 0.8405 & 0.8762 & 0.9245 \\
        PanNuke & & \greycell\B 0.9324 & 0.8332 & 0.9190 & \greycell\B 0.9372 & 0.8926 & 0.9299 & 0.8901 \\
        Singapore & & \greycell\B 0.8359 & 0.7173 & 0.7839 & \greycell\B 0.8488 & 0.7486 & 0.8044 & 0.9103 \\
        MHIST & &\greycell\B 0.7656 & 0.6786 & 0.7179 & \greycell\B 0.7802 & 0.7236 & 0.7534 & 0.7042 \\
        TIL & & \greycell\B 0.9270 & 0.8934 & 0.9048 & \greycell\B 0.9300 & 0.9127 & 0.9154 & 0.9216 \\
        Camelyon & & \greycell\B 0.7067 & 0.6179 & 0.5229 & \greycell\B 0.6689 & 0.6147 & 0.5570 & 0.8440$^\dagger$ \\
        PANDA & & \greycell\B 0.7099 & 0.5675 & 0.6763 & \greycell\B 0.7243 & 0.6648 & 0.6873 & - \\
        IMP 1K & & \greycell\B 0.7186 & 0.6415 & 0.6389 & \greycell\B 0.7471 & 0.6850 & 0.6832 & - \\
        IMP 4k & & \greycell\B 0.8554 & 0.7657 & 0.8332 & \greycell\B 0.8632 & 0.8356 & 0.8454 & - \\
        IMP Cervix & & \greycell\B 0.7439 & 0.7289 & 0.7301 & \greycell\B 0.7622 & 0.7247 & 0.7346 & - \\
    \bottomrule
    \end{tabular}}
    \caption{}
    \label{tab:short_segpath_results}
    \end{subfigure}

    \begin{subfigure}{\textwidth}
        \centering
        \includegraphics[width=0.9\textwidth]{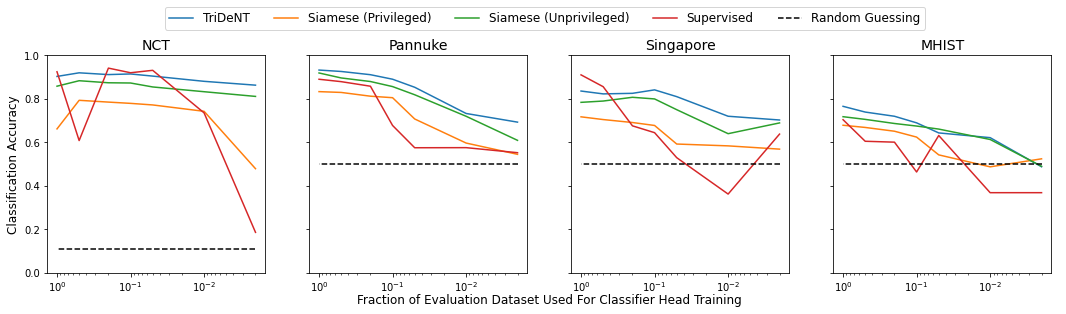}
        \caption{}
        \label{fig:dataset-sizes}
    \end{subfigure}
    
    \caption{Difference in accuracy between TriDeNT \Neptune~ and privileged/unprivileged Siamese training on SegPath. Values greater than zero (above the dashed line) indicate a higher accuracy for TriDeNT \Neptune. (b) Results for ten evaluation tasks averaged across all eight stains. Supervised baseline is provided for comparison, bold indicates best performance for the given self-supervised loss function. Supervised comparisons are only given for patch-level tasks, as train-time patch aggregation for slide-level tasks cannot be comparably achieved. Higher values indicate better performance. For full results see Table \ref{tab:segpath}. Value marked $\dagger$ from \cite{farndale2024synthetic}. (c) Classification training dataset size performance comparison. Models were all pretrained on SegPath and evaluated using the full test set of each dataset. Training was carried out on 100\%, 50\%, 20\%, 10\%, 5\%, 1\%, and 0.2\% of each classifier training dataset, and averaged over all SegPath stains. Supervised comparisons are trained in the same fashion, but not averaged.}
    \label{fig:segpath_results}
\end{figure*}

\section{Results}

\subsection{Embedding Knowledge From Privileged Image Modalities}
 
We first demonstrate that TriDeNT \Neptune ~is highly effective for improving the quality of representations in the primary encoder by distilling privileged information from immunofluorescence (IF) images to H\&E stained images (Figure \ref{fig:segpath_results} and Table \ref{tab:segpath}). Models are trained on the SegPath dataset \cite{komura2023restaining}, which consists of eight subsets of H\&E images paired with an image derived from the IF stain of a consecutive slice for one of eight antibodies. Evaluation is performed on four standard computational pathology tasks (see Appendix \ref{sec:datasets} for full details). We find that the model significantly increases performance by up to 101\% compared to a privileged baseline model. TriDeNT \Neptune ~ retains not only the useful features shared between inputs, but also the features which are only present in the primary data, leading to better performance on all evaluated tasks. Even in cases where the privileged data does not appear to significantly improve performance, TriDeNT \Neptune ~still achieves comparable performance, as it obtains a strong supervisory signal from the additional H\&E branch. This is in contrast with the privileged Siamese setting, where it is clear that the pairing can cause a seismic drop in classification accuracy if the privileged data is not informative for the task being evaluated (see Section \ref{sec:harmful-privileged-info} for a more detailed analysis).

We see that there are significant performance gains of up to 101\% (0.4566 to 0.9169, see Table \ref{tab:segpath}) in the NCT tissue type classification task for TriDeNT \Neptune ~against the baseline privileged method. The improved performance from using TriDeNT \Neptune ~is seen across the board, with average performance improvements on all tasks, and only a handful of cases where individual stains underperform baseline models. We generally observe that when TriDeNT \Neptune ~performs worse than a baseline, it is only marginal, however, there are many cases where the performance difference between TriDeNT \Neptune ~and a baseline is enormous. For example, on Camelyon, performance is improved from effectively random guessing at 51\% up to 81\% with pan-CK as privileged information.

Perhaps unsurprisingly given the diagnostic importance of cytokeratin stains for detecting tumours, the greatest increases in performance against the unprivileged baseline method were generally achieved for the pan-CK model, with similar gains for $\alpha$SMA. Notably, for the TIL task the immune-related stains CD3CD20 and CD45RB achieved the best performance, as this privileged information was more task relevant than others. Compared to the baseline unprivileged method, there was less benefit for pairing CD235a or ERG, perhaps because red blood cells (stained by CD235a) and the endothelium (stained by ERG) were less relevant to the tasks being assessed. Still, compared to the baseline privileged method, performance on CD235a and ERG was significantly improved. 

We see that in Siamese models, some stains help improve prediction accuracy while others hinder it. For example, in the PanNuke neoplastic cell detection task, privileged Siamese training is considerably less accurate for MIST1 and ERG stains, which stain plasma and endothelial cells respectively, while it is more accurate for $\alpha$SMA and pan-CK, which stain smooth muscle cells/myofibroblasts and epithelial cells respectively. We show in Figure \ref{tab:segpath-characteristics} that this difference in performance is associated with differences in the proportion of empty space in the privileged information (see Section \ref{sec:harmful-privileged-info} for further discussion). While all stainings provide valuable information about their specific cell types, some have very few features which can be learned (see Appendix \ref{sec:segpath-characteristics} for more details). This causes the primary encoder's representations to collapse and perform poorly on downstream tasks, as Siamese models can only learn features which are shared between branches. TriDeNT \Neptune, in contrast, can retain the features from both primary and privileged information, leading to improved performance over unprivileged models. In Section \ref{sec:harmful-privileged-info} we will further discuss how TriDeNT \Neptune ~can mitigate the effects of harmful or uninformative privileged information, compared to Siamese methods. We also note that there are some small differences in the distributions and sample sizes of the tissue samples used for different stains. For example, in the most extreme case the unprivileged models have a range of 0.1399 for their accuracies on the Singapore gland malignancy detection task.

The performance improvements observed in patch-level tasks are found to carry over into slide-level tasks, which are shown in Table \ref{tab:segpath-2}. These tasks are generally considered to be of more clinical relevance than patch-level tasks, as they involve making predictions on the level of the patient. We consider here the detection and grading of dysplasia, and the ISUP grading of tumours. We find that a simple aggregation of the representations from models trained with TriDeNT \Neptune ~is an effective predictor on these tasks, with strong performance across all tasks.

\begin{table*}[t]
    \centering
    \caption{Results for models trained on the BCI dataset containing H\&E patches paired with privileged brightfield IHC.}
    \label{tab:rawihc}
    \scalebox{1}{\begin{tabular}{cccccc}
    \toprule
         Loss & Method & Privileged & BCI & NCT & PanNuke \\
    \midrule
         \multirow{3}{*}{VICReg} & \greycell TriDeNT \Neptune & \greycell\cmark & \greycell\B 0.8559 & \greycell\B 0.8347 & \greycell 0.8966 \\
         & \multirow{2}{*}{Siamese} & \cmark & 0.8552 & 0.8019 & \B 0.9071 \\
         & & \xmark & 0.6863 & 0.8103 & 0.8506 \\
    \midrule
         \multirow{3}{*}{InfoNCE} & \greycell TriDeNT \Neptune & \greycell\cmark & \greycell\B 0.8800 & \greycell\B 0.8267 & \greycell\B 0.9115 \\
         & \multirow{2}{*}{Siamese} & \cmark & 0.8319 & 0.7961 & 0.8677 \\
         & & \xmark & 0.7034 & 0.8045 & 0.9023 \\
    \midrule
        CrossEntropy & Supervised & - & 0.6331 & 0.9245 & 0.8901 \\
     \bottomrule
    \end{tabular}}
\end{table*}

\subsubsection{TriDeNT \Neptune ~Pretrained Models Outperform Supervised Models on Small Datasets}
\label{sec:classifier-dataset-size}

In Fig. \ref{fig:dataset-sizes} we demonstrate that TriDeNT \Neptune~ consistently retains a higher level of performance as less classifier training data is used. Notably, there are dataset sizes where supervised and privileged Siamese models collapse to a trivial solution, while TriDeNT \Neptune ~continues to perform well. The ability to learn well from tiny, few-shot classification datasets is evidence of the utility of models trained with TriDeNT \Neptune ~for a variety of downstream applications, as in many biomedical settings there are very few samples available for a given topic of interest. TriDeNT \Neptune ~can allow researchers and clinicians to make use of these few-shot datasets to enable the study of previously unworkable datasets.

\subsection{Embedding Knowledge from Additional Brightfield Images}

To demonstrate the generality of the method, we train models on the BCI dataset of paired H\&E and brightfield IHC patches. We only perform evaluation on the BCI, NCT and PanNuke datasets, as the BCI dataset is a breast cancer dataset, while the Singapore and MHIST datasets are prostate and colorectal polyp specimens respectively, which are far out of the training distribution. We include the NCT dataset, despite comprising only colorectal tissue, as these patches are well curated into different tissue type classes which mostly bear a strong resemblance to those in breast cancer samples. Strikingly, TriDeNT \Neptune ~outperforms the supervised baseline by a large margin on the BCI task. We propose that there may be features weakly present in H\&E stains which are highly predictive of HER2 status, and the pairing with IHC stains which contain those features very strongly results in this improved performance.

As Table \ref{tab:rawihc} shows, we find that TriDeNT \Neptune ~is also highly effective on all tasks compared to the unprivileged Siamese baseline. As there is more information in the privileged paired data, the privileged baseline is considerably higher for this task. Despite this, TriDeNT \Neptune ~still outperforms both comparable baselines on all tasks but one, achieving improvements of up to 25.1\% compared to the unprivileged baseline and up to 5.8\% compared to the privileged baseline. There is only a single task where TriDeNT \Neptune ~does not improve performance: evaluation on the PanNuke dataset of a model trained with the VICReg loss, performing 1.6\% less than the privileged baseline. The brightfield IHC stains contain considerably more task-relevant information for cell segmentation, so this is unsurprising. This effect can be understood visually as the IHC `weak' quadrant in Figure \ref{fig:latent_space} being very narrow and containing very few features. Most task-relevant features are strongly present in the IHC and therefore there is less to be gained by adding the few missing weak features.

\begin{table*}[t]
    \centering
    \caption{Results for models trained on the PanNuke dataset containing H\&E patches paired with nuclear segmentation masks.}
    \label{tab:pannuke_results}
    \scalebox{1}{\begin{tabular}{cccccccc}
    \toprule
         Loss & Method & Privileged & NCT & PanNuke & Singapore & MHIST \\
    \midrule
         \multirow{3}{*}{VICReg} & \greycell TriDeNT \Neptune & \greycell\cmark & \greycell\B 0.7337 & \greycell\B 0.9106 & \greycell\B 0.7975 & \greycell\B 0.7523 \\
         & \multirow{2}{*}{Siamese} & \cmark & 0.6000 & 0.8274 & 0.7106 & 0.6530 \\
         & & \xmark & 0.7301 & 0.8682 & 0.7754 & 0.7421 \\
    \midrule
         \multirow{3}{*}{InfoNCE} & \greycell TriDeNT \Neptune & \greycell\cmark & \greycell\B 0.7530 & \greycell\B 0.9115 & \greycell\B 0.8226 & \greycell 0.7369 \\
         & \multirow{2}{*}{Siamese} & \cmark & 0.5289 & 0.7403 & 0.6951 & 0.6264 \\
         & & \xmark & 0.7199 & 0.8668 & 0.8015 & \B 0.7451 \\
    \midrule
        CrossEntropy & Supervised & - & 0.9245 & 0.8901 & 0.9103 & 0.7042 \\
     \bottomrule
    \end{tabular}}
\end{table*}

\begin{figure*}[!t]
    \centering
    \begin{subfigure}{\textwidth}
        \includegraphics[width=\textwidth]{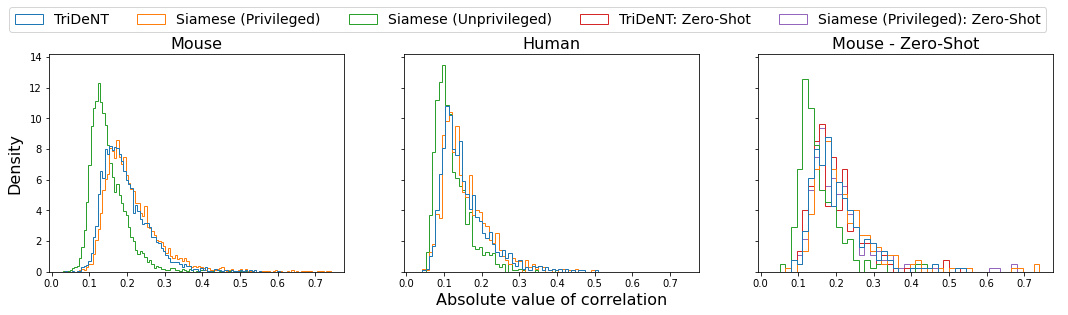}
    \caption{}
    \label{fig:als-st-correlations}
    \end{subfigure}

    \begin{subfigure}{0.7\textwidth}
    \centering
        \scalebox{0.7}{\begin{tabular}{cccccc}
    \toprule
         & & & Human & \multicolumn{2}{c}{Mouse} \\
         Loss & Method & Privileged & White/Grey & White/Grey & Genotype \\
    \midrule
         \multirow{3}{*}{VICReg} & \greycell TriDeNT \Neptune & \greycell\cmark & \greycell\B 0.8395 & \greycell\B 0.8861 & \greycell\B 0.5634 \\
         & \multirow{2}{*}{Siamese} & \cmark & 0.8072 & 0.8833 & 0.5518 \\
         & & \xmark & 0.8179 & 0.8741 & 0.5513 \\
    \midrule
         \multirow{3}{*}{InfoNCE} & \greycell TriDeNT \Neptune & \greycell\cmark & \greycell\B 0.8401 & \greycell\B 0.8870 & \greycell 0.5614 \\
         & \multirow{2}{*}{Siamese} & \cmark & 0.8109 & 0.8856 & 0.5383 \\
         & & \xmark & 0.8045 & 0.8782 & \B 0.5662 \\
    \midrule
        MSE & Direct Gene Prediction & - & 0.7527 & 0.8669 & 0.5426\\
    \midrule
        CrossEntropy & Supervised & - & 0.8590 & 0.8720 & 0.4933 \\
     \bottomrule
    \end{tabular}}
    \vspace{5mm}
    \caption{}
    \label{tab:st}
    \end{subfigure}
    ~
    \begin{subfigure}{0.28\textwidth}
        \includegraphics[width=\textwidth]{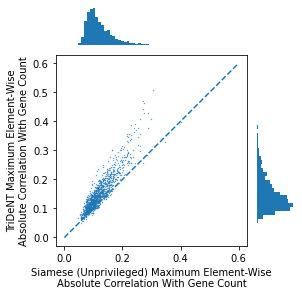}
        \caption{}
        \label{fig:counts-corr}
    \end{subfigure}
    \caption{(a) Correlation histograms between representations and gene count arrays for mouse and human ALS-ST data. Bins are chosen using the maximum of the Sturges \cite{sturges1926choice} and Freedman-Diaconis \cite{freedman1981histogram} estimators. In the third histogram, zero-shot models are evaluated on genes which were not seen during training, while other models which did see those genes in training are evaluated on the same genes for comparison (of course, unprivileged models never see any genes). Comparison with models which saw these genes during training. (b) Spatial transcriptomics results for white/grey matter classification with both VICReg and InfoNCE losses. Baselines provided are `Direct Gene Prediction', where a supervised model is trained to predict the gene counts for that patch directly and the representation is then fine-tuned on the white/grey classification task, and a standard supervised model. (c) Greater correlation strengths between gene counts and representations of TriDeNT \Neptune ~models than unprivileged Siamese models. For each gene, the maximum absolute correlation between the TriDeNT \Neptune ~representations for each patch and the corresponding gene counts are plotted against those for unprivileged Siamese representations, with TriDeNT \Neptune ~almost always achieving greater correlation strength. Dashed line is the identity. Appended histograms show distribution of data. Mouse data only, see Figure \ref{fig:counts-correlation-plot-human} for human data, which shows a similar pattern, and Figure \ref{fig:counts-correlation-plot-mouse} for extended comparisons of mouse data, also including privileged Siamese and supervised results.}
    \label{fig:corr-hist}
\end{figure*}

\subsection{Image Annotations are an Effective Source of Privileged Information}
\label{sec:pannuke-results}

We find that TriDeNT \Neptune ~is effective not only for integrating additional sources of data, but also for manually determining the most useful aspects of the data which should be learned, where the user has some prior knowledge to incorporate into the dataset. This is intuitively the opposite of traditional machine learning approaches, where the user has to handcraft inputs to be passed to the model, and the model only learns from those features. With our approach, the user can manually handcraft inputs, such as the segmentation masks in this example, while still giving the model the flexibility to learn other features not known a priori to the user. The results in Table \ref{tab:pannuke_results} demonstrate that TriDeNT \Neptune ~is able to train encoders which retain the features of both the nuclei and the background/connective tissue. We see performance improvements of up to 42.4\% compared to the privileged baseline, and up to 5.2\% compared to the unprivileged baseline.

These results also suggest that, in the privileged Siamese case, the features that are learned are those relating to the shape of the nuclei, rather than any sub-nuclear features or features relating to the connective tissue which would enable better identification of tissue and cell types.

\subsection{Vision Models with Privileged Spatial Transcriptomics Data Learn More Biologically Relevant Features}
\label{sec:st-results}

A key application of TriDeNT \Neptune ~is the distillation of information from privileged sources beyond images. As TriDeNT \Neptune ~does not require the architecture of each branch to be the same, it is possible to utilise any input type on any branch. We investigate the use of spatial transcriptomics (gene expression counts from an array of spatial points on a slide) as privileged data to train models for H\&E inputs. These data have been shown to be highly informative and enable the study of the relationship between gene expression and tissue morphology, however, they are very expensive to generate, and as such are far from routine use. The difficulty of this task is compounded by the established poor performance of deep learning methods on tabular data \cite{shwartz2022tabular,borisov2022deep}.

Despite this, we see consistent improvements of up to 4.4\% for TriDeNT \Neptune ~over other methods for the spatial transcriptomics white matter/grey matter classification task, as shown in Figure \ref{tab:st}. It is likely the case that there are some mislabelled examples due to the processes involved in alignment, so higher accuracy on this task may simply not be possible, which could explain the saturation of performance around 89\% in the mouse example. We observe a similar improvement for VICReg on the genotype prediction task, with an improvement of up to 2.2\%. InfoNCE shows a similar performance for TriDeNT \Neptune ~and unprivileged Siamese models, which both outperform privileged Siamese.

To assess the level of information shared between the transcriptomic results and the representations of the H\&E patches, we investigate the cross-correlation between elements of the representations and the gene counts for each matching patch. We calculate the cross-correlation across the validation set between each element in the representations and the count for each gene, and for each gene take the correlation of the corresponding element with the maximum correlation or minimum anti-correlation, whichever has the greater absolute value. This maximum/minimum is chosen because the vast majority of elements will not correlate with any given gene, and the absolute value is taken because the sign of the element is arbitrary, so correlation and anti-correlation are equivalent. We use the absolute value of the correlation for the element selected for each gene, and use these to generate the histograms in Figure \ref{fig:als-st-correlations}. It is clear that privileged training obtains representations which are far more correlated to the gene counts than unprivileged training, with minimal differences in the correlations between TriDeNT \Neptune ~and Siamese approaches. This implies that the models have learned equivalently informative representations about the coarse-grained features of the genes. Figure \ref{fig:counts-corr} demonstrates that the correlation strength is significantly greater for TriDeNT \Neptune ~compared to an unprivileged Siamese model, and Figures \ref{fig:counts-correlation-plot-mouse} and \ref{fig:counts-correlation-plot-human} show the relationships between the gene correlations of representations from TriDeNT \Neptune, Siamese methods, and supervised learning. Figures \ref{fig:wiki-heatmap} and \ref{fig:kegg-heatmap} show the geneset enrichment for each method, demonstrating that TriDeNT \Neptune ~captures more meaningful interrelationships that are more informative about the relationship between tissue morphology and gene expression than unsupervised Siamese models. This is especially important for scientific discovery, as these analyses are used to generate hypotheses for further research. Figure \ref{fig:als-st-umaps} shows UMAP projections of the representation space coloured by genotype and gene, to illustrate that TriDeNT \Neptune ~identifies distinct morphological clusters which are not found by unprivileged Siamese models. Figure \ref{fig:als-st-correlations} also shows that the findings are robust to human and mouse datasets, indicating the generality of the method.

\begin{figure*}[t]
    \centering
    \begin{subfigure}{0.45\textwidth}
       \includegraphics[width=\textwidth]{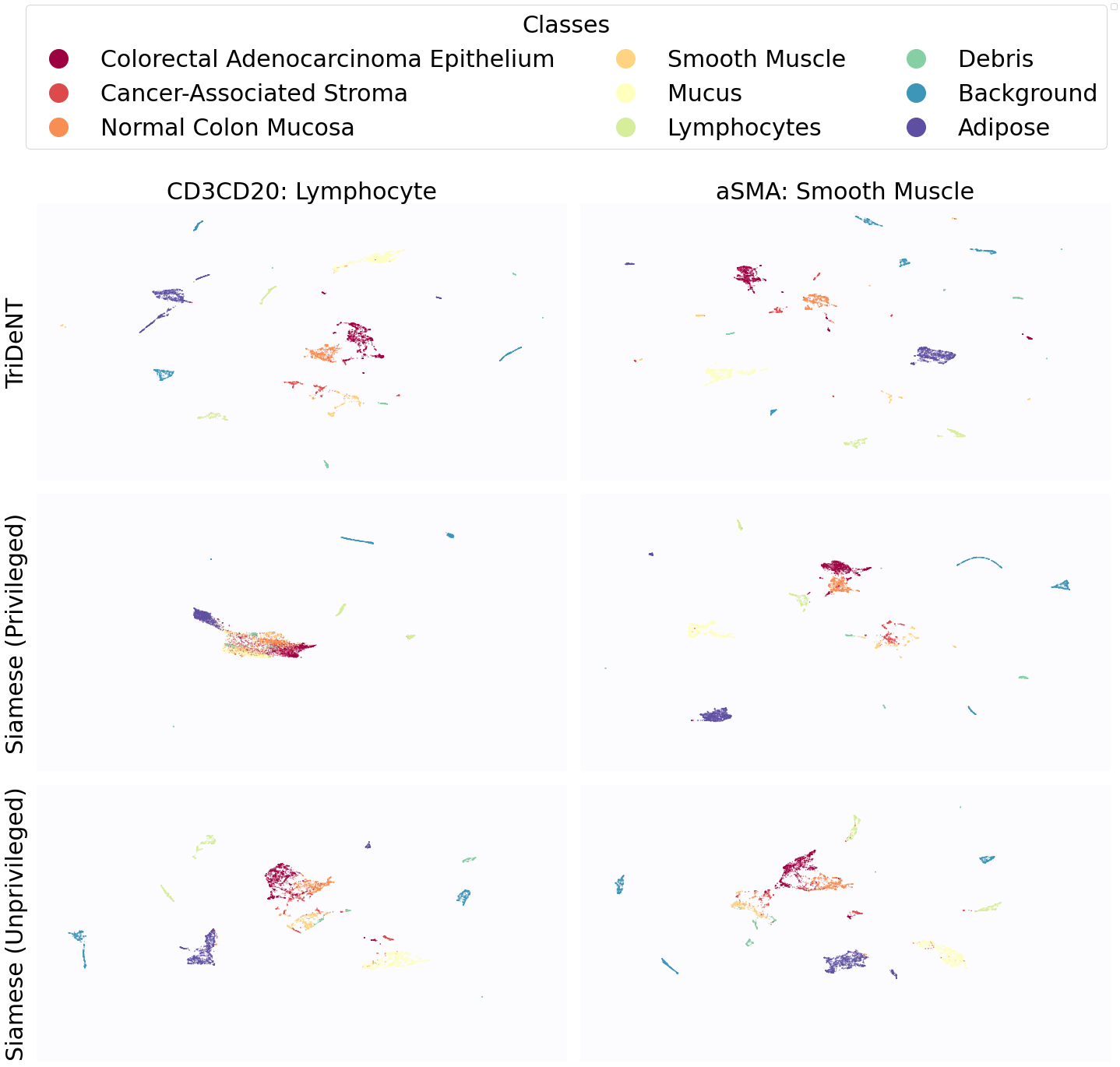}
        \caption{}
        \label{fig:main-umaps} 
    \end{subfigure}
    ~
    \centering
    \begin{subfigure}{0.45\textwidth}
        \includegraphics[width=\textwidth]{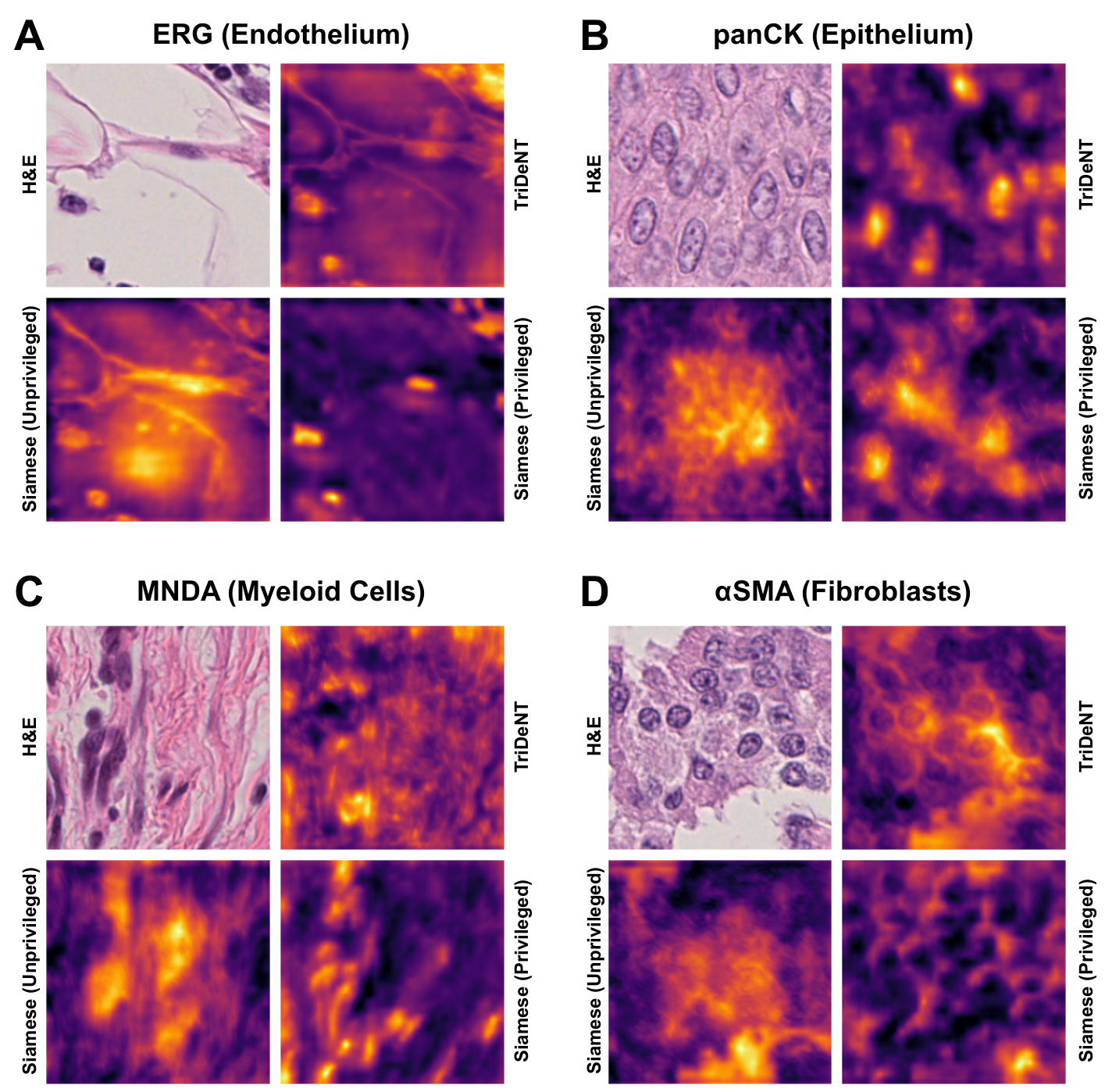}
        \caption{}
        \label{fig:GradCAM}
    \end{subfigure}
    \caption{(a) Sample UMAP projections from 2048 dimensions into 2 for models trained on the SegPath CD3CD20 and $\alpha$SMA subsets, evaluated on the NCT test dataset. Points are coloured by tissue type. Note that accuracies for these tasks were i) TriDeNT \Neptune: CD3CD20 0.8982, $\alpha$SMA 0.9273; Siamese (Privileged): CD3CD20 0.6625, $\alpha$SMA 0.9186; Siamese (Unprivileged): CD3CD20 0.8694, $\alpha$SMA 0.8570. (b) GradCAM heatmaps for selected images from the SegPath dataset. Evaluated with VICReg loss. Brighter colours represent greater activation strengths. For a larger selection, including heatmaps for InfoNCE models, see Figures \ref{fig:supp-grad-mnda-vicreg} to \ref{fig:supp-grad-panCK-infonce}.}
    \label{fig:umap-gradcam}
\end{figure*}

We also demonstrate the correlations of the model's representations with genes unseen during the self-supervised phase of training (Figure \ref{fig:als-st-correlations}). We find that the model is not simply overfitting on the given genes, as these genes are not present in the training data, yet the privileged models still demonstrate greater correlation with their counts than an unprivileged approach.

Existing approaches for integrating spatial gene expression data with tissue morphology \cite{he2020integrating} have focused on directly generating the transcriptomics from the H\&E patches. While effective for predicting the expression of a given gene, this is highly ineffective for learning useful representations of the coarse-grained tissue features. Generative models have been shown to produce representations which are contain less semantic information than joint-embedding architectures and do not perform as well on downstream tasks  \cite{assran2022masked,assran2023self}. This has also been shown specifically for pathology images \cite{farndale2023more}. We confirm this by directly predicting the gene counts from the H\&E patches, and show in Table \ref{tab:st} (Supervised (Transfer)) that transfer task performance is inferior to both the TriDeNT \Neptune ~and Siamese approaches. We note that this leads to representations which do not generalise well, as is also true for other supervised methods and image-to-image translation methods.

\subsection{TriDeNT \Neptune~ Identifies Features of Both Primary and Privileged Inputs from Primary Input Alone}

To further analyse the learned representations, we produce UMAP projections of the latent space labelled with the tissue types for the NCT tissue type classification task, as shown for CD3CD20 and $\alpha$SMA in Figure \ref{fig:main-umaps}, and for all SegPath stains in Figures \ref{fig:supp_umaps_1} and \ref{fig:supp_umaps_2}. These figures make the reasons for the varying performance of the privileged Siamese model more apparent. For stains with better performing privileged Siamese models, such as $\alpha$SMA, the UMAPs are very similar between Siamese methods and TriDeNT, with well-differentiated tissue type clusters. In those with worse performance, such as ERG, the tissue types are poorly differentiated, often with only adipose and background forming distinct clusters from the other classes. On closer inspection, it is notable in these projections that TriDeNT \Neptune~ produces more well-defined and separated clusters in general than Siamese networks. This is further evidenced in Figure \ref{fig:als-st-umaps}, where TriDeNT \Neptune ~is shown to identify clusters with overexpression of a given gene significantly more effectively than an unprivileged Siamese model. Interstingly, we find that the privileged Siamese model for CD3CD20 forms distinct clusters for the \emph{lymphocytes} class, which corresponds well to the privileged information. In contrast, ERG and MNDA appear similarly but without the presence of this cluster, suggesting that the privileged information impacts the presence of certain clusters.

We also analyse the activation maps for each model using GradCAM as described in Appendix \ref{sec:training_implementation}. This offers more insight into the areas of the image which are contributing most heavily to the models' representations. In Figure \ref{fig:GradCAM} we present some representative examples, however, a larger selection which was chosen at random is presented in Figures \ref{fig:supp-grad-mnda-vicreg} to \ref{fig:supp-grad-panCK-infonce}. The larger selection makes it easier to see the emergent patterns. We see that unprivileged Siamese models tend to focus primarily on image features such as textures and colour, particularly when the image contains white background. Privileged Siamese models tend to focus on regions associated with their privileged information, primarily cell nuclei for panels A,B, and C, and smooth muscle for D. TriDeNT \Neptune ~occupies an intermediate position, incorporating both features specific to the privileged data and more the general features associated with unprivileged Siamese networks.

We can see in Figures \ref{fig:supp-grad-erg-vicreg} and \ref{fig:supp-grad-erg-infonce} that for ERG, the privileged Siamese model focuses almost exclusively on nuclei. As there are very few endothelial cells in the dataset, it could be an effective strategy to identify anything that could potentially be an endothelial cell to minimise the difference between the representations of the H\&E model and the IF mask model (see Appendix \ref{sec:segpath-characteristics} for more details). In the corresponding unprivileged Siamese image, we see that the model identifies some of these nuclei, albeit less strongly, but also focuses heavily on the other tissue and even the background, while strongly fixating on two spots of debris in the center of the image. This model has less `incentive' to learn the weak features related to endothelial cells as these occur rarely and are not easy to detect, while more generic strong features such as the presence of connective tissue and the prevalence of background are more common and predictable from augmented images. We see that the TriDeNT \Neptune ~ERG model also largely ignores nuclei, primarily focusing on the connective tissue, supporting the argument that TriDeNT \Neptune ~learns to ignore the privileged information when it is not useful. We note that no VICReg model appears to focus on the endothelial cells in the images we have tested, however the InfoNCE TriDeNT \Neptune ~and privileged Siamese models do successfully identify this cell (e.g. row 2, column 4 in Figures \ref{fig:supp-grad-erg-vicreg} and \ref{fig:supp-grad-erg-infonce}).

In panel C we see a similar pattern, with the privileged Siamese model fixating solely on the nuclei, while the TriDeNT \Neptune ~model takes a more balanced approach. The unprivileged Siamese model appears to focus on a single cluster of nuclei while neglecting others, and similarly identifies an area of fibroblasts with its distinctive pattern but does not others.

In contrast to panels A and C which represent models with poor privileged Siamese results, panels B and D represent models whose privileged Siamese results were comparable to both TriDeNT \Neptune ~and even the supervised baseline. It is therefore interesting to note that there are far more similarities between the privileged Siamese and TriDeNT \Neptune ~models in both cases. Particularly in panel B, TriDeNT \Neptune ~and the privileged Siamese model return virtually identical heatmaps, with both strongly identifying epithelial nuclei and neglecting the same areas of connective tissue. The unprivileged model in this case appears to focus solely on the centre of the image, giving a significantly different heatmap to the other panels.

Panel D again shows the previous pattern, with the privileged Siamese model identifying the features strongly present in the privileged data -- fibroblasts -- while neglecting the nuclei present. TriDeNT \Neptune ~also strongly identifies the connective tissue, but, unlike the privileged Siamese model, does not completely neglect the nuclei. The unprivileged Siamese model primarily identifies background, and does not appear to identify the nuclei in this example.

\begin{table*}[t]
    \centering
    \caption{Results for models trained on the PanNuke dataset containing H\&E patches paired with redundant or harmful privileged information -- blank patches or randomly shuffled nuclear segmentation masks -- to assess whether TriDeNT \Neptune ~can mitigate the impact of detrimental privileged information. We denote the best performance in a category in bold, and the second best with an underline.}
    \label{tab:pannuke_blank_shuffled_results}
    \begin{tabular}{ccccccccc}
    \toprule
         Paired Data & Loss & Method & Privileged & NCT & PanNuke & Singapore & MHIST \\
    \midrule
         \multirow{6}{*}{Blank Patches} & \multirow{3}{*}{VICReg} & \greycell TriDeNT \Neptune & \greycell\cmark & \greycell \U{0.6943} & \greycell\U{0.8125} & \greycell \U{0.7592} & \greycell \U{0.7257} \\
         & & \multirow{2}{*}{Siamese} & \cmark & 0.5378 & 0.7227 & 0.6142 & 0.5670 \\
         & & & \xmark & \B 0.7301 & \B 0.8682 & \B 0.7754 & \B 0.7421 \\
    \cmidrule{2-8}
         & \multirow{3}{*}{InfoNCE} & \greycell TriDeNT \Neptune & \greycell\cmark & \greycell\B 0.7680 & \greycell\U{0.8092} & \greycell \U{0.7875} & \greycell \U{0.7345} \\
         & & \multirow{2}{*}{Siamese} & \cmark & 0.6466 & 0.7063 & 0.6200 & 0.6034 \\
         & & & \xmark & \U{0.7199} & \B 0.8668 & \B 0.8015 & \B 0.7451 \\
    \midrule
        \multirow{6}{*}{Shuffled Patches} & \multirow{3}{*}{VICReg} & \greycell TriDeNT \Neptune & \greycell\cmark & \greycell\B 0.7429 & \greycell\U{0.8235} & \greycell\B 0.7932 & \greycell\U{0.7277} \\
         & & \multirow{2}{*}{Siamese} & \cmark & 0.5075 & 0.7254 & 0.6313 & 0.5865 \\
         & & & \xmark & \U{0.7301} & \B 0.8682 & \U{0.7754} & \B 0.7421 \\
    \cmidrule{2-8}
         & \multirow{3}{*}{InfoNCE} & \greycell TriDeNT \Neptune & \greycell\cmark & \greycell\B 0.7598 & \greycell \U{0.7751} & \greycell \U{0.7803} & \greycell \U{0.7316} \\
         & & \multirow{2}{*}{Siamese} & \cmark & 0.6267 & 0.7084 & 0.6269 & 0.6689 \\
         & & & \xmark & \U{0.7199} & \B 0.8668 & \B 0.8015 & \B 0.7451 \\
     \bottomrule
    \end{tabular}
\end{table*}

\subsection{TriDeNT \Neptune ~Mitigates Harmful and Uninformative Privileged Information}
\label{sec:harmful-privileged-info}

TriDeNT \Neptune ~is designed to integrate privileged information, and the normal assumption is that this privileged information is useful. However, the types of information found in real medical data are highly heterogeneous and can contain both information that highly useful and information that is completely irrelevant. This is studied with two scenarios: blank privileged information, and randomly shuffled privileged information. Blank privileged information provides no information and can only be detrimental to performance, and randomly shuffled privileged information contains information that has no correspondence to the primary H\&E patch it is paired with. The objective is to assess the ability of TriDeNT \Neptune, and comparable baselines, to mitigate the influence of this irrelevant or harmful privileged information.

Table \ref{tab:pannuke_blank_shuffled_results} shows that TriDeNT \Neptune ~achieves comparable performance to unprivileged models, implying that TriDeNT \Neptune ~is able to ignore the irrelevant and harmful privileged information, and only learn features from the primary input. In fact, in some cases with randomly shuffled patches TriDeNT \Neptune ~marginally improves performance. This could be because the irrelevant information still contains some common features between patches, such as the shapes of nuclei, which encourage the primary encoder to learn to detect similar round shapes in the H\&E.

In contrast, the performance of privileged Siamese models is very poor, with this baseline achieving the worst performance on every test dataset. This is because by mapping into a single shared latent space, the primary and privileged models can only learn features which are shared between inputs, and nothing can be learned from vacuous inputs.

Despite TriDeNT employing no explicit feature selection mechanisms, we see that this method can dynamically select features which optimise its objective. When privileged information is useful, features are selected which are correlated with the privileged information. When the privileged information is irrelevant or harmful, it is ignored in favour of the primary features that would be learned by unprivileged methods. This is critical for the real-world usefulness and general applicability of TriDeNT \Neptune, which can be used to effectively distil any source of privileged information without the potential for damaging performance, as routinely happens with privileged models.

\begin{figure}[t]
\centering
    \begin{subfigure}{0.6\textwidth}
        \includegraphics[width=\textwidth]{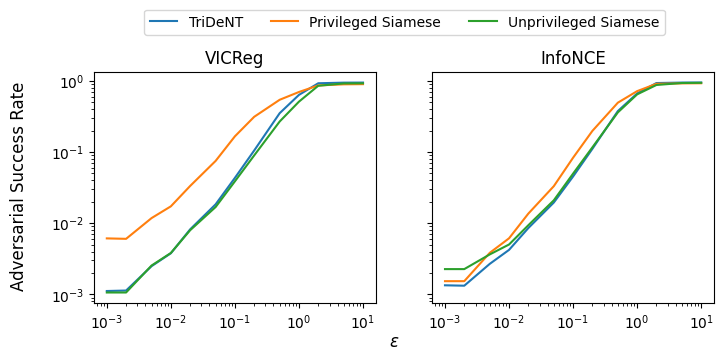}
        \caption{}
        \label{fig:adversarial-robustness-plot}
    \end{subfigure}
    \hspace{5mm}
    \begin{subfigure}{0.3\textwidth}
        \scalebox{0.8}{\begin{tabular}{cccc}
        \toprule
             Loss & Method & Privileged & SSR $\downarrow$ \\
        \midrule
             \multirow{3}{*}{VICReg} & \greycell TriDeNT \Neptune & \greycell\cmark & \greycell \U{0.6850} \\
             & \multirow{2}{*}{Siamese} & \cmark & 1.6903 \\
             & & \xmark & \B 0.6248 \\
        \midrule
             \multirow{3}{*}{InfoNCE} & \greycell TriDeNT \Neptune & \greycell\cmark & \greycell\B 0.8679 \\
             & \multirow{2}{*}{Siamese} & \cmark & 1.1392 \\
             & & \xmark & \U{0.9929} \\
         \bottomrule
        \end{tabular}
        }
        \vspace{12mm}
        \caption{}
        \label{tab:adversarial-robustness-table}
    \end{subfigure}
\caption{(a) Average adversarial robustness to PGD attacks for models trained on Segpath stains. For full results see Tables \ref{tab:full-adversarial-0.001-0.05} and \ref{tab:full-adversarial-0.1-10}. (b) Standardised Success Rate (SSR) for these models. Lower values are better, with bold indicating best performance and underline indicating second best.}
\label{fig:adversarial-robustness}
\end{figure}

\subsection{Robustness of TriDeNT \Neptune ~to Domain Shift and Adversarial Attacks}
\label{sec:adversarial-robustness}

A growing concern with computational pathology models is the robustness of models to discrepancies between their training data and evaluation data. This can be either in the form of unintentional differences caused by different image acquisition methods, causing a domain shift, or in the form of intentional adversarial attacks \cite{ghaffari2022adversarial}. These are malicious attacks that are designed to augment the input data in a way that causes the model to make an incorrect classification while being imperceptible to humans. The robustness of models to these attacks is of interest for translation of models to clinical use due to the criticality of their predictions to human health. These models must not be vulnerable to small perturbations, either as a result of malicious actors or unintentional variation in measurement. 

A concern with TriDeNT \Neptune ~and other models using privileged information could be that the successful distillation of information leads models to be less robust. Privileged models are learning at least some weaker features than unprivileged models, and consequently could in theory be easier to exploit than unprivileged models which learn only strong features in the primary information (see Section \ref{sec:privilegedinfo} and Figure \ref{fig:latent_space} for definitions and discussion of weak/strong features).

In Figure \ref{fig:segpath_results} we have presented results on the Camelyon test set, which assesses models' domain transfer performance when trained on images from three hospitals and evaluated on images from another, with large differences between hospitals. This shows that TriDeNT \Neptune ~can achieve very strong performance on different domains, implying that the features learned with the TriDeNT \Neptune ~training regime are more robust. This is in contrast to unprivileged models, which achieve little better than random guessing.

We present results for the adversarial robustness of all SegPath models in Figure \ref{fig:adversarial-robustness}. We used a white-box adversarial attack -- Projected Gradient Descent (PGD) \cite{madry2017towards} -- to assess the robustness of each model, and find that TriDeNT \Neptune ~has a similar robustness to unprivileged models, while privileged Siamese models are considerably less robust. This is because TriDeNT \Neptune ~retains the strong primary features as well as the weaker, privileged features, and consequently is less affected than the privileged Siamese models which are reliant solely on the weaker privileged features. This is an important finding for the clinical relevance of TriDeNT \Neptune ~compared to unprivileged models, as it implies that not only do these models learn features which are more robust to the overt feature shift of domain transfer, they also learn features which are similarly robust to potential adversarial attacks. The results in Table \ref{tab:adversarial-robustness-table} show a standardised success rate metric, which is defined as the mean standardised value for each model
\begin{equation}
    SSR_m\coloneqq\left.\frac{1}{|E|}\sum_{\epsilon\in E} \frac{SR_{m\epsilon}}{||SR_\epsilon||}\right.,
\end{equation}
where $m$ is a model in the set of models (TriDeNT \Neptune, privileged Siamese, and unprivileged Siamese), $E$ is the set of perturbation strengths $\epsilon$, and $SR_{m\epsilon}$ is the success rate of the attack with perturbation strength $\epsilon$ on model $m$. This determines the magnitude of the difference between models across all values of $\epsilon$, while accounting for the different scales of these values, with a smaller value indicating better adversarial robustness.

\section{Discussion and Conclusions}

In this work, we have proposed TriDeNT \Neptune, a modelling approach which has been demonstrated to effectively integrate privileged sources of data into single-source models during training to improve performance. The model works by providing two supervisory signals to the primary encoder, which dynamically respond to the features which the primary encoder can extract. Experiments have shown that this approach can greatly outperform standard Siamese privileged and unprivileged methods, and even supervised learning, without significantly increasing the computational overheads. There are a vast number of biomedical datasets which contain paired data, such as paired -omics datasets \cite{tcga,schorn2021community}, different imaging methods (e.g. PESO \cite{bulten2019epithelium}), and even multiple images of the same source, such as the 7-pt skin lesion dataset \cite{kawahara2018seven}, CheXpert \cite{irvin2019chexpert} and CheXphoto \cite{phillips2020chexphoto}.

\subsection{Integrating Privileged Data is Invaluable for Research and Discovery}

The utility of TriDeNT \Neptune ~for research applications can be found not only in increasing the efficacy of primary data models for prediction accuracy, but also in training models to extract coarse-grained features which are relevant to the privileged input. Our results demonstrate that models trained with TriDeNT \Neptune ~will perform better on tasks where the privileged information is more task relevant. This is demonstrated, for example, where H\&E prediction of HER2+ status is greatly improved by pairing with HER2 IHC stains (Table \ref{tab:rawihc}), where immune-related privileged SegPath stains lead to better performance on the TIL task (Figure \ref{fig:segpath_res_tri_priv}), and where performance on metastasis- and malignancy-related tasks are most improved by privileged pan-CK stains (Figure \ref{fig:segpath_res_tri_priv}). This is also shown qualitatively with the GradCAM activation heatmaps in Figure \ref{fig:GradCAM}. A typical use case is that a scientist with a paired dataset could train a model to then evaluate an unpaired dataset, without needing to acquire more paired data. We have shown that the features which are found by privileged methods are significantly different from those found by unprivileged methods. This means that TriDeNT \Neptune ~could enable the identification of novel morphological clusters that are functionally important, such as those in our analyses in Figures \ref{fig:main-umaps} and \ref{fig:als-st-umaps}, which might not emerge from other methods of training or training on the new dataset alone.

\subsection{TriDeNT \Neptune ~Does Not Need Large Datasets}

Self-supervised methods typically require very large datasets \cite{reed2022self}, however our results, especially those for PanNuke (Section \ref{sec:pannuke-results}) and ALS-ST (Section \ref{sec:st-results}), demonstrate that TriDeNT \Neptune ~ offers improvements over comparable baselines for comparatively small pretraining datasets. We also studied the effect of evaluation dataset size in Section \ref{sec:classifier-dataset-size}, showing that TriDeNT \Neptune ~continues to achieve strong performance even when the classifier head is trained on a tiny dataset. This performance can only be expected to improve further if pretrained models are used, either from a general source such as ImageNet \cite{russakovsky2015imagenet} or from more specific pretraining tasks. We expect that this would be particularly useful in cases where the privileged paired model is pretrained, as teacher-student distillation would likely lead to greater performance in the student (primary) model.

\subsection{TriDeNT \Neptune ~ Can Incorporate Image Annotations Into Representation Learning}

Our experiments have demonstrated that models trained using TriDeNT \Neptune ~learn significantly different features to those trained in standard self-supervised settings, and that this can be leveraged to manually encode information by the user. For example, we demonstrated in Section \ref{sec:pannuke-results} that the model can be made to learn features related to nuclear segmentation masks, without requiring human prior knowledge of what those precise features might be. This offers new opportunities to make better use of the manual annotations which are provided with many datasets but typically only used as target labels for supervised learning. We have shown that these annotations can be used to create more generalist, robust and effective models when transferred to other tasks, either related or unrelated to the annotations.

Of course, not all annotations are manual, and machine-generated annotations, such as those from HoVer-Net \cite{graham2019hover}, could be incorporated into training procedures. Currently there exist a huge number of models which have been trained for one specific task, such as nuclear segmentation (e.g. \cite{graham2019hover}), tissue type annotation (e.g. \cite{kather2016multi}), virtual restaining (e.g. \cite{xu2019gan}), feature detection (e.g. \cite{aubreville2023mitosis}), etc., and all of these could be incorporated into new generalist models using TriDeNT \Neptune.

\subsection{Future Research}

While TriDeNT \Neptune ~offers a new capability for multi-modal distillation in medical imaging, further improvements can be made. Model weights were always frozen for downstream tasks, so the tasks detailed in this work are all zero-shot, meaning fine-tuning these models could lead to improved performance. Design choices were primarily made for simplicity and parity with previous work, and hyperparameters were chosen based on previous work on Siamese networks (e.g. \cite{bardes2021vicreg,chen2020simple}), so it is highly likely that the results are skewed in favour of these Siamese networks. Training was also only carried out for 100 epochs (200 for the spatial transcriptomics examples) on a batch size of 128, so models could potentially improve further with longer training times and larger batch sizes. Despite this, we have shown that TriDeNT \Neptune ~outperforms these methods, often by a considerable margin. Improvements could be made by adjusting the loss function, or by implementing more elaborate interactions between branches. The scope of this study was limited to histopathology, however TriDeNT \Neptune ~could be broadly applicable to other domains in both imaging and other modalities. We expect TriDeNT \Neptune ~to have extensive applications for multiplexed imaging, as the best way of integrating these multiple sources of information has not been established.

We also anticipate that utilising different network architectures on different branches could yield interesting results, such as pairing convolutional neural networks (CNNs) with graph neural networks, transformers or simply a larger CNN. We showed this is a possibility in Section \ref{sec:st-results}, where a primary CNN is paired with an multilayer perceptron for the privileged spatial transcriptomics data. This would enable different features and patterns to be identified and could lead to models which utilise the efficiency of CNNs with the power of these additional methods.

\section*{Declaration of Competing Interest}

The authors declare that they have no known competing financial interests or personal relationships that could have appeared to influence the work reported in this paper.

\section*{CRediT Authorship Contribution Statement}
\textbf{Lucas Farndale:} Conceptualisation, Software, Formal analysis, Methodology, Validation, Data curation, Visualisation, Writing - original draft. \textbf{Robert Insall:} Resources, Funding acquisition, Project administration, Supervision, Writing - review \& editing. \textbf{Ke Yuan:} Conceptualisation, Resources, Funding acquisition, Project administration, Supervision, Writing - review \& editing.

\section*{Code and Data Availability}

Code will be made available upon publication. All datasets used are publicly available from the following links: 
\begin{itemize}
    \item ALS-ST -- \href{https://als-st.nygenome.org/}{\url{als-st.nygenome.org}};
    \item BCI -- \href{https://bupt-ai-cz.github.io/BCI/}{\url{bupt-ai-cz.github.io/BCI/}};
    \item Camelyon -- \href{https://wilds.stanford.edu/datasets/#camelyon17}{\url{wilds.stanford.edu/datasets/}};
    \item IMP 1K/4K -- \href{https://rdm.inesctec.pt/km/dataset/nis-2023-008}{\url{rdm.inesctec.pt/km/dataset/nis-2023-008}};
    \item IMP Cervix -- \href{https://rdm.inesctec.pt/km/dataset/nis-2024-003}{\url{rdm.inesctec.pt/km/dataset/nis-2024-003}};
    \item MHIST -- \href{https://bmirds.github.io/MHIST/}{\url{bmirds.github.io/MHIST}}; 
    \item NCT Colorectal Cancer -- \href{https://zenodo.org/record/1214456}{\url{10.5281/zenodo.1214455}};
    \item PANDA -- \href{https://www.kaggle.com/c/prostate-cancer-grade-assessment/data}{\url{kaggle.com/c/prostate-cancer-grade-assessment/data}};
    \item PanNuke -- \href{https://warwick.ac.uk/fac/cross_fac/tia/data/pannuke}{\url{warwick.ac.uk/fac/cross_fac/tia/data/pannuke}}; 
    \item SegPath -- \href{https://dakomura.github.io/SegPath/}{\url{dakomura.github.io/SegPath}};
    \item Singapore Prostate Cancer -- \href{https://zenodo.org/record/7152243}{\url{10.5281/zenodo.7152243}}.
    \item TIL -- \href{https://zenodo.org/records/6604094}{\url{10.5281/zenodo.6604094}};
\end{itemize}
The exact patchings and dataset splits are available from the authors upon reasonable request where this is permitted by the dataset's license.

\section*{Acknowledgments}
Lucas Farndale is supported by the MRC grant MR/W006804/1, Robert Insall is supported by EPSRC grant EP/S0300875/1 and Wellcome grant 221786/Z/20/Z. Ke Yuan acknowledges support from EP/R018634/1, BB/V016067/1.

The authors would like to extend our gratitude to Adalberto Claudio-Quiros and Kai Rakovic for the helpful feedback and discussion.

\bibliographystyle{unsrt}
\bibliography{refs}

\clearpage

\noindent \textbf{Lucas Farndale} is a 2nd year PhD student at the CRUK Scotland Institute and the University of Glasgow. His research focuses on developing new computational and mathematical methods for biomedical imaging, with a focus on multimodal data integration. He received an MSci Mathematics degree from the University of Glasgow in 2022.
\subsection*{  } % This subsection (with no heading) is added to give more space between two biographies
\noindent \textbf{Robert Insall} is Professor of Computational Cell Biology at University College London and Professor of Mathematical and Computational Cell Biology at the University of Glasgow. A true cross-disciplinary researcher, trained in biochemistry and cell biology but with publications in mathematics, physics and computer science. Approaches include biochemistry and quantitative microscopy, mathematical and computational modelling, cell biology and machine learning. Recently funded by Wellcome Trust, MRC, EPSRC, InnovateUK, Datalabs Edinburgh, and CRUK.
\subsection*{  } % This subsection (with no heading) is added to give more space between two biographies
\noindent \textbf{Ke Yuan} is a Senior Lecturer in Machine Learning and Computational Biology jointly appointed at the Schools of Computing Science, the School of Cancer Sciences at the University of Glasgow and the Cancer Research UK Scotland Institute. Before Glasgow, he was a postdoctoral research fellow at the Cancer Research UK Cambridge Institute from 2012 to 2016. He received an MSc and PhD from the University of Southampton in 2008 and 2013, respectively.

% \setcounter{figure}{0}
% \renewcommand{\figurename}{Fig.}
% \renewcommand{\thefigure}{S\arabic{figure}}
% \setcounter{table}{0}
% \renewcommand{\tablename}{Table}
% \renewcommand{\thetable}{S\arabic{table}}
% \setcounter{algorithm}{0}
% % \renewcommand{\algorithmname}{Algorithm}
% \renewcommand{\thealgorithm}{S\arabic{algorithm}}
% \setcounter{section}{0}
% % \renewcommand{\sectionname}{Section}
% \renewcommand{\thesection}{S\arabic{section}}

\clearpage

% \part*{Supplementary Information for TriDeNT: Triple Deep Network Training for Privileged Knowledge Distillation in Histopathology}

% \title{Supplementary Information for TriDeNT: Triple Deep Network Training for Privileged Knowledge Distillation in Histopathology}

% \author{Lucas Farndale
%  $^{*,1,2,3,4}$ \lforcid \and Robert Insall$^{1,2,5}$ \riorcid \and Ke Yuan$^{*,1,2,3}$ \kyorcid}

% \date{$^1$ School of Cancer Sciences, University of Glasgow \\
% $^2$ Cancer Research UK Scotland Institute \\
% $^3$ School of Computing Science, University of Glasgow \\
% $^4$ School of Mathematics and Statistics, University of Glasgow \\
% $^5$ Division of Biosciences, University College London \\}

% \maketitle
\appendix

% \part*{Supplementary Information}

\section{Biological Background}
\subsection{Histology}

To analyse tissue samples, pathologists take slices of tissue around 5$\mu$m thick to be analysed under a microscope. To make it easier to identify different structures, the slide is typically \emph{stained} with chemicals that bind tightly to different components of the sample, dying them different colours. By far the most common staining method used in histopathology is haematoxylin and eosin (H\&E) staining. Haematoxylin stains components that are rich in nucleic acids, such as nuclei and ribosomes, while Eosin stains common protein structures pink, such as connective tissue, collagen, and the cytoplasm \cite{kierszenbaum2015histology}.

\emph{Immunohistochemistry} is an ancillary staining technique often used in medical diagnosis, which works by using an antibody to target specific proteins in tissue. Secondary chemical processes are coupled to the antibody to produce a colour, making it much easier to identify their presence and location within tissue. There are many different antibodies used which each target a specific protein or protein type. Common examples of antibody targets include cytokeratins, which are found in epithelial cells, CD3, CD4, and CD20, which are found in various types of immune cells, and smooth muscle actin (SMA), which is found in myofibroblasts \cite{buchwalow2010immunohistochemistry}.

Typically these stains are highly informative about one particular protein of instance, but lack the generality of H\&E staining. Consequently, they are primarily used as secondary sources of information to assist with research or diagnosis. 

\subsection{Spatial Transcriptomics}

Understanding the spatial context of gene expression is critical to interpreting the functions that underlie tissue architectures. Until recently, analysis of RNA could only be performed in bulk, on whole tissue samples, which made studying the relationships between tissue morphology and gene expression very difficult. \emph{Spatial transcriptomics} \cite{staahl2016visualization} is a state-of-the art set of techniques which enable the transcriptomic reads to be spatially resolved to a point on the tissue, population of cells, or possibly even within a single cell. This has enabled the study of spatial patterns of gene expression, the behaviour of rare cell types within tissue \cite{kleshchevnikov2022cell2location}, and cell-cell interaction and variability \cite{tian2023expanding}.  The method has been refined to allow analysis down to the level of single molecules, but remains prohibitively expensive for routine clinical or research use.

The standard 10x Visium technology \cite{rao2020bridging} obtains both an H\&E slide and spatial transcriptomic array for consecutive sections of each sample, where each transcriptomic read corresponds to a hexagon on the H\&E slide. This enables us to distil privileged information from gene expression into representations of tissue morphology.

\section{Datasets and Tasks}
\label{sec:datasets}

\begin{figure*}
    \centering
    \begin{subfigure}[t]{0.18\textwidth}
    \centering
        \includegraphics[width=\textwidth]{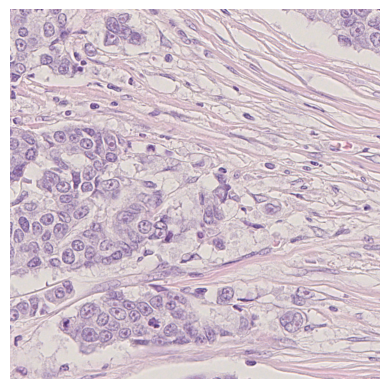}
        \caption{SegPath -- H\&E}
        \label{fig:segpathhe}
    \end{subfigure}
    \begin{subfigure}[t]{0.18\textwidth}
    \centering
        \includegraphics[width=\textwidth]{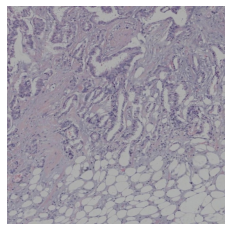}
        \caption{BCI -- H\&E}
        \label{fig:bcihe}
    \end{subfigure}
    \begin{subfigure}[t]{0.18\textwidth}
    \centering
        \includegraphics[width=\textwidth]{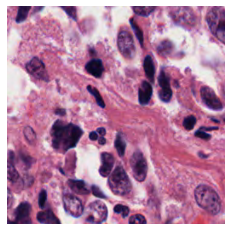}
        \caption{PanNuke -- H\&E}
        \label{fig:pannukehe}
    \end{subfigure}
    \hspace{10mm}
    \begin{subfigure}[t]{0.18\textwidth}
    \centering
        \includegraphics[width=\textwidth]{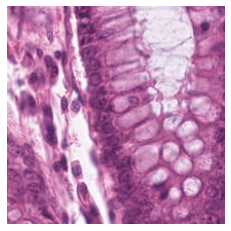}
        \caption{NCT -- H\&E}
        \label{fig:ncthe}
    \end{subfigure}
    \begin{subfigure}[t]{0.18\textwidth}
    \centering
        \includegraphics[width=\textwidth]{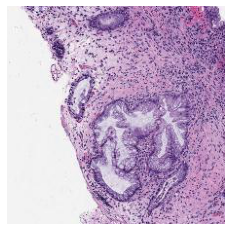}
        \caption{MHIST -- H\&E}
        \label{fig:mhisthe}
    \end{subfigure}

    \begin{subfigure}[t]{0.18\textwidth}
    \centering
        \includegraphics[width=\textwidth]{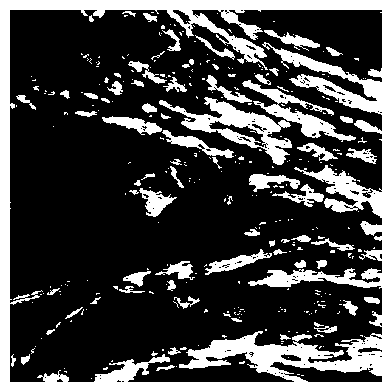}
        \caption{SegPath -- $\alpha$SMA Thresholded Fluorescence Image}
        \label{fig:segpathif}
    \end{subfigure}
    \begin{subfigure}[t]{0.18\textwidth}
    \centering
        \includegraphics[width=\textwidth]{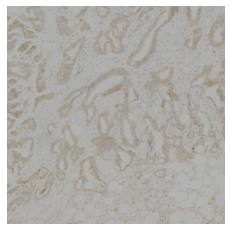}
        \caption{BCI -- IHC}
        \label{fig:bciihc}
    \end{subfigure}
    \begin{subfigure}[t]{0.18\textwidth}
    \centering
        \includegraphics[width=\textwidth]{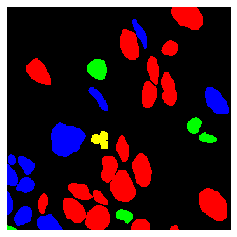}
        \caption{PanNuke -- Cell Segmentation Mask}
        \label{fig:pannukemask}
    \end{subfigure}
    \hspace{10mm}
    \begin{subfigure}[t]{0.18\textwidth}
    \centering
        \includegraphics[width=\textwidth]{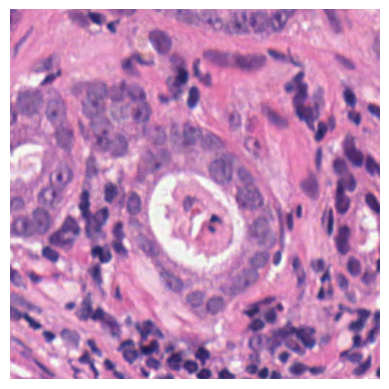}
        \caption{Singapore -- H\&E}
        \label{fig:singaporehe}
    \end{subfigure}
    \begin{subfigure}[t]{0.18\textwidth}
    \centering
        \includegraphics[width=\textwidth]{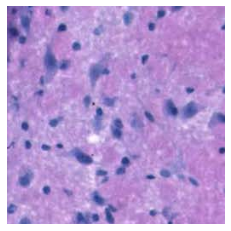}
        \caption{ALS-ST Mouse -- H\&E}
        \label{fig:mousehe}
    \end{subfigure}
    \caption{(a-c, f-h) Samples from each training dataset with privileged paired data. Colours in (h) correspond to different cell types, as described in Appendix \ref{sec:pannuke}. (d,e,i,j) Samples from evaluation datasets.}
    \label{fig:datasets}
\end{figure*}

\begin{table}[t]
    \centering
    \caption{SegPath stains and tissue structures stained by each antibody. The number of patches given is the number of $256\times256$px patches obtained following preprocessing.}
    \label{tab:segpath-stains}
    \scalebox{1}{\begin{tabular}{llc}
    \toprule
        Stain & Staining Target & No. Patches \\
    \midrule
        $\alpha$SMA & Smooth Muscle/Myofibroblasts & 498 848 \\
        CD235a & Red Blood Cells & 414 544 \\
        CD3CD20 & Lymphocytes & 196 368 \\
        CD45RB & Leukocytes & 396 880 \\
        ERG & Endothelium & 170 352 \\
        MIST1 & Plasma Cells & 211 696 \\
        MNDA & Myeloid Cells & 226 160 \\
        pan-CK & Epithelium & 424 144 \\
    \bottomrule
    \end{tabular}}
\end{table}

\subsection{SegPath}

SegPath \cite{komura2023restaining} is a large pan-cancer dataset containing 8 subsets corresponding to 8 different IF stains with paired H\&E stains, which are detailed in Table \ref{tab:segpath-stains}. Stains were performed individually on each slide, not multiplexed, and H\&Es were obtained from the same section as the IF by initially staining with H\&E, then washing the slides and restaining with the IF stain. Different slides were used for each stain, so there is inter-subset variability between the image features present in the subsets. Slides were tissue microarrays featuring samples from cancer types including glioma, meningioma, ependymoma, kidney renal clear cell carcinoma, lung adenocarcinoma, lung squamous cell carcinoma, breast adenocarcinoma, gastric adenocarcinoma, colon adenocarcinoma, pancreatic adenocarcinoma, cholangiocarcinoma, hepatocellular carcinoma, esophageal squamous cell carcinoma, head and neck squamous cell carcinoma, urothelial tumors, bladder cancer, prostate adenocarcinoma, sarcoma, melanoma, uterine cancer, ovarian tumors, and testicular germ cell tumors.

We paired H\&E images with corresponding thresholded IF images of the same (restained) patch from the SegPath dataset \cite{komura2023restaining}. IF images were obtained by thresholding IF intensity, such that pixels with lower intensity than the cutoff were labelled 0 and those higher than the cutoff were labelled 1. For ease of computing and consistency of size with evaluation datasets, we resized each $984\times 984$px image to $1024\times1024$px with linear interpolation, and split these into 16 $256\times256$px patches, and did the same for each corresponding mask. Each input was a tuple $(\bm{x}_{HE}, \bm{x}_{HE}, \bm{x}_{IF})$, with both $\bm{x}_{HE}$ and $\bm{x}_{IF}$ being $256\times256$px (pixel) patches from the same location on both the H\&E and restained slides. Given that no evaluation was to be performed directly on the SegPath dataset, the train, validation and test sets were combined to increase the dataset size, with the number of patches for each stain after preprocessing detailed in Table \ref{tab:segpath-stains}.

\subsection{BCI: Breast Cancer Immunohistochemisty}

Between 15-20\% of breast cancers will overexpress the protein HER2 (Human Epidermal growth factor Receptor 2), which promotes the growth of cancer cells, and typically have worse prognosis than HER2-negative cancers \cite{burstein2005distinctive}. Testing for HER2 is carried out using IHC staining, with 4 grades: 0, 1+, 2+, and 3+, with 0 being negative and 3+ being the highest positive grade. The BCI dataset \cite{Liu_2022_CVPR} contains 4871 $1024\times1024$px patches from 51 H\&E stained slices obtained from breast cancer patients, paired with corresponding spatially registered IHC patches. We split these into patches of size $256\times256$px for consistency with the evaluation datasets. We retained the same train/test split for the new patches, giving train/test datasets of size 62,336/15,632 respectively.

We primarily used BCI as a pretraining dataset to demonstrate the efficacy of TriDeNT \Neptune~ for raw IHC paired with H\&E patches, however, we also evaluated the pretrained model based on 4-class classification of HER2 status. The dataset was highly imbalanced (0: 5.18\%, 1+: 23.59\%, 2+: 43.56\%, 3+: 27.67\%) so we used proportional class weighting during classifier head training, but not self-supervised pretraining. For this reason we only used BCI as an evaluation set for the BCI pretrained models.

\subsection{NCT-CRC-HE-100K: National Center for Tumor Diseases ColoRectal Cancer H\&E Dataset (NCT)}

Commonly used as a benchmark for assessing model classification performance, the NCT (National Center for Tumour diseases) dataset \cite{kather_dataset} consists of 100 000 manually extracted, Macenko-normalised \cite{macenko2009method}, $224\times224$px patches from 86 colorectal cancer patients, with pathologists' annotations from the classes \emph{Adipose} (ADI), \emph{Background} (BACK), \emph{Debris} (DEB), \emph{Lymphocytes} (LYM), \emph{Mucus} (MUC), \emph{Smooth Muscle} (MUS), \emph{Normal Colon Mucosa} (NORM), \emph{Cancer-Associated Stroma} (STR), and \emph{Colorectal Adenocarcinoma Epithelium} (TUM). This dataset also contains an additional test set, obtained under the same conditions, featuring 7180 patches from 50 patients who had no overlap with those in the training set.

The evaluation task was the classification of the nine tissue types. We also used this dataset to illustrate the models' latent spaces using UMAP \cite{mcinnes2018umap}. This reduced the dimensionality of the representations from 2048 to 2, showing an approximate graphical illustration of the distribution of points in high-dimensional space.

\subsection{PanNuke}
\label{sec:pannuke}

We next evaluated the models' abilities to detect neoplastic cells using the \emph{PanNuke} dataset \cite{gamper2019pannuke}. A neoplasm is a type of abnormal and excessive growth of tissue, which can be either benign or malignant. PanNuke is a pan-cancer dataset of $256\times256$px H\&E patches with paired nuclear segmentation masks, with six classes: \emph{Neoplastic}, \emph{Non-Neoplastic Epithelial}, \emph{Inflammatory}, \emph{Connective}, \emph{Dead} and \emph{Background}. We evaluated the model on classifying patches containing neoplastic and non-neoplastic cells. Following \cite{huang2023visual}, we denoted patches containing no neoplastic cells as non-neoplastic, and patches containing at least 10 neoplastic cells where at least 30\% of all cells in the patch are neoplastic as neoplastic patches. This yielded a dataset with three folds of sizes 2205, 2090, and 2283 respectively. We used the first two folds for training and the third fold for testing, and provide three-fold cross-validation in Table \ref{tab:pannuke_cross_validation} to justify this.

Segmentation models typically learn fine-grained details such as the exact locations of features, which comes at the cost of identifying features in the images. In contrast, the augmentations used in self-supervised models discourage them from learning to identify the exact locations of features and similarly precise image properties, as these can change under augmentation. This enables self-supervised models to transfer more robustly to different tasks, as was shown in \cite{farndale2023more}, where a U-Net segmentation model was demonstrated to perform significantly worse when transferred to a classification task than a comparable self-supervised model. Given that the objective of feature extraction models is to learn these more abstract features, rather than the precise properties found by segmentation models, we evaluated only on the neoplastic classification task, rather than the segmentation task.

\subsection{Singapore Prostate Gland Dataset}

We refer to the prostate gland classification dataset described in \cite{oner2022ai} as the \emph{Singapore Dataset}, as the data were collected from 46 prostate cancer patients at Tan Tock Seng Hospital, Singapore. The dataset consists of patches of size $512\times512$px, at 20x, centered on a potentially cancerous prostate gland, which we center cropped to $256\times256$px. Train, validation and test splits were constructed on the patient level, leaving a train/valid/test split of 3843/2557/4261 respectively, with an overall balance of 4316 benign patches to 6336 malignant.

The task associated with this dataset is the binary classification between benign and malignant glands. We initially observed collapse to the trivial local optimum of only predicting malignant in many cases, so we used proportional class weights during classifier head training.

\subsection{MHIST: A Minimalist Histopathology Image Analysis Dataset}

Colorectal polyps are common precancerous growths which are usually benign but in a small number of cases develop into colorectal cancer. In general, it is difficult to predict which polyps will develop into cancer, but one known case is the \emph{sessile serrated adenoma}, which is known to turn cancerous if left untreated. In contrast, \emph{hyperplastic} polyps are typically benign.

MHIST \cite{wei2021petri} contains 3152 $224\times224$px patches obtained from 328 whole slide images at 8x magnification, featuring either a hyperplastic or serrated polyp as determined by a majority vote of 7 pathologists. We resized these patches to $256\times256$px. There was a train/test split of 2175/977, and there are 2162 hyperplastic patches compared to 990 serrated polyps.

The task associated with MHIST is binary classification between serrated and hyperplastic polyps. Due to the imbalance of the dataset, we used proportional class weights during classifier head training. While there is no overlap of patients between any evaluation datasets and SegPath, there are colorectal and prostate cancer samples in the dataset. MHIST contains images of polyps, so was a zero-shot task which was unseen for the model during training, in addition to being $8\times$ magnification compared to SegPath's $40\times$.

\subsection{Camelyon}

If a machine learning model is to have any clinical relevance, a critical feature is its ability to generalise to different domains. There can be a large difference in image appearance between different hospitals, scanners, staining protocols, and lengths of time since the tissue was collected, among other factors. The WILDS distribution \cite{koh2021wilds} of the Camelyon17 dataset \cite{bandi2018detection} is designed to evaluate models' ability to generalise beyond their training data to other datasets. Strong performance on this task indicates that the features learned by the model are generalisable and robust, while weak performance suggests that the model has overfitted on features specific to its training dataset. Camelyon consists of whole-slide-images of breast lymph nodes from five different hospitals, of which three comprise the training set, with one for the validation set and a final hospital for the test set. These images are $96\times96$px, with a binary label indicating whether the central $32\times32$px region contains any tumour tissue. The aim is to detect metastasis to the lymph node. We resize these images to $256\times256$px.

\subsection{Tumour Infiltrating Lymphocytes (TIL)}

As a departure from previous tasks which primarily assess the presence of malignancy, metastasis, and direct features of tumours, we use this task to assess a secondary feature of cancer: the presence of tumour infiltrating lymphocytes (TILs). These are a highly prognostic immune feature that can be well-predicted from H\&E images \cite{saltz2018spatial} and require models to have learned a different feature set to those used in previous tasks. The spatial distribution of TILs is also the subject of active research, and lends itself well to computational approaches due to the scalability of this type of analysis to large H\&E datasets \cite{abousamra2022deep}.

To assess models' ability to identify TILs, we use the dataset provided by \cite{abousamra2022deep}, which contains 304,097 patches from 7983 whole slide images of 23 cancer types taken from TCGA \cite{tcga}. Images are natively $100\times10$px at 0.5 mpp and we resize them to $256\times256$px, and an image is considered TIL-positive if there are at least two TILs present in the image.

\subsection{PANDA: Prostate cANcer graDe Assessment}

As one of the largest publicly available whole-slide datasets, PANDA is one of the most widely used datasets in computational pathology. The dataset features 10616 images of prostate biopsies with accompanying Gleason and ISUP scores, which assess different growth patterns in the tumours. This is a key evaluation of models' clinical applicability, as tumour grading is a critical task for pathologists when determining a patient's prognosis. This task is somewhat dependent on the method used to aggregate the patches to assign a single score for the whole slide, and we use a simple min-mean-max aggregation method to enable a fair comparison, as detailed in Appendix \ref{sec:training_implementation}. As is standard for this task, performance is reported as quadratically weighted Cohen's kappa ($\kappa$), as slides are scored out of 5, meaning misclassifying by one point is better than misclassifying by two, etc.

\subsection{IMP Colorectal Dysplasia 1K/4K}
\label{sec:imp-colorectal}

Colorectal cancer is characterised by a progression from healthy tissue, to mild dysplasia (minor changes to cells), through high-grade dysplasia (major changes to cells), before finally becoming an invasive carcinoma. If detected early, colorectal cancer usually be cured by surgery, however, late stage cancer has poor survival rates \cite{ahmed2014advances}. Therefore, the ability to stratify pre-cancerous polyps based on the grade of their dysplasia is an important task in the management of cancer and prognosis of patients. The IMP dataset contains 5333 whole slide images with a label of either non-neoplastic, low-grade lesions (conventional adenomas with low-grade dysplasia), or high-grade lesions (conventional
adenomas with high-grade dysplasia, intra-mucosal carcinomas and invasive adenocarcinomas). There is a test set featuring 900 slides, with a training set of 1132 (1K) slides and an extension to 4433 slides (4K). We present results for both datasets.

\subsection{IMP Cervical Dysplasia}

As with colorectal cancer as described in Appendix \ref{sec:imp-colorectal}, the management of cervical cancer is highly dependent on early detection and diagnosis, with a large negative survival gradient depending on the stage of the patient's cancer \cite{ward2020role}. This dataset contains 600 whole-slide images obtained from cervical loop electrosurgical excision procedures. There is no dedicated test set, so all results are from 5-fold cross-validation.

\subsection{ALS-ST: Amyotrophic Lateral Sclerosis Spatial Transciptomics}
\label{sec:als-st-dataset}

We next investigated whether models of H\&E patches can be improved by pairing them with the data obtained for that patch using spatial transcriptomics. We constructed a dataset from \cite{maniatis2019spatiotemporal} by obtaining patches of size $256\times256$px, centered on the centroid of each hexagon, paired with the gene count array from the corresponding Illumina transcriptomic read \cite{rao2020bridging}. This was used as input to a model with $\bm{x}$ being H\&E patches, and $\bm{x}^*$ the privileged gene count array. The dataset features 80 human and 331 mouse spinal cord sections from 7 humans and 67 mice, each with one H\&E stained image and corresponding spatially resolved transcriptomics. All human samples are from patients with Amyotrophic Lateral Sclerosis (ALS), the most common form of Motor Neuron Disease. ALS is a neurodegenerative condition that causes muscles to gradually weaken and waste, with a normal life expectancy of 2-4 years from symptom onset \cite{chio2009prognostic}.

For each sample, we aligned the spatial read locations with the H\&E images using the mappings from the original work, and extracted patches of size $256\times256$px centered on each location, which were paired with the corresponding gene count array. This resulted in datasets of size 62561 and 50017 for mouse and human samples respectively. We constructed train/test splits on the sample level for the mice, with the samples from 13 out of 67 mice reserved for the test dataset. Due to the small number of human samples, we reserved 20\% of patches as a test dataset.

To reduce redundancy and noise in the count data, genes were only included if they appeared at least 50 times with a count greater than 5, such that there were 5020 genes measured in each mouse sample, and 1136 for each human sample. Patches were also categorised into either white or grey matter based on their spatial location, as described in \cite{maniatis2019spatiotemporal}, with the labels \emph{Dorsal Horn}, \emph{Ventral Horn} and \emph{Medial Grey} being classified as grey matter and \emph{Ventral Medial White}, \emph{Ventral Lateral White}, \emph{Medial Lateral White} and \emph{Dorsal Medial White} being classified as white matter. The labels \emph{Central Canal}, \emph{Ventral Edge}, \emph{Lateral Edge}, \emph{Dorsal Edge} and \emph{Undefined} were excluded.

To compare performance on a H\&E-based analysis task, we performed binary classification between white and grey matter in the mouse dataset, based on the labels described above. This yielded a white/grey dataset with a train/test split of 44928/14848 patches. The balance of white/grey patches was 17444/27522 in the train set and 4433/7215 in the test set. We also classified patches based on the SOD1 genotype of the mouse the sample was extracted from, with the same train/test split as before and label distribution of 16415/26174/19972 (train) and 6671/5488/4294 (test) for SOD1-G93A (ALS)/SOD1-WT (Wildtype)/Knockout respectively.

To further analyse the performance of these models and the features they learn, we evaluated how elements of the models' representations of H\&E patches correlated with the gene counts corresponding to them. To do this, we calculated the cross-correlation matrix of all genes with all representation elements, and for each gene considered only the maximum correlation of any representation element with that gene's count. This is because a good representation will have limited redundancy, meaning most elements will not encode information about any given gene. We also considered only the absolute value of each correlation, as the sign of each representation element is arbitrary in this setting. The correlation score $c_i$ of each gene $i$ can therefore be written as
\begin{equation}
    c_i \coloneqq \max_{j=1,...,D}\lvert C_{ij}\rvert
\end{equation}
where $D$ is the dimension of the representation and $C_{ij}$ is the correlation between the count of gene $i$ and element $j$ of the representations.

\section{Training and Implementation Details}
\label{sec:training_implementation}

Models were trained for 100 epochs, with a batch size of 128. Models for ALS-ST examples were trained for 200 epochs as these were slower to converge. The backbone encoder was a ResNet-50 (output size 2048) with average pooling and random initialisation, and a 3-layer dense projection head was used with layer size 8192, batch normalisation between layers, ReLU activations, and a linear final layer. In ALS-ST examples, the privileged encoder had the same architecture as the projector, with layer sizes 8192, 8192, 2048. A warmup-cosine learning rate schedule was used, starting from 0 with the warmup period being one tenth of the number of epochs, and a maximum learning rate of $10^{-4}$ ($10^{-3}$ for spatial transcriptomics). We compared two different types of loss -- contrastive (InfoNCE, \cite{oord2018representation}) and non-contrastive (VICReg, \cite{bardes2021vicreg}) -- to show that the results are robust to different self-supervised approaches. In keeping with the original implementations, we used the parameters $\lambda=25, \mu=25, \nu=1$ for the VICReg loss, the temperature parameter $\tau=0.5$ for the InfoNCE loss. Augmentations are detailed in Tables \ref{tab:image_augmentations} (images) and \ref{tab:gene_augmentations} (gene counts). Pseudocode is provided in Algorithm \ref{alg:pseudocode}.

For classification tasks, model weights were frozen and a linear classification head consisting of a batch normalisation and a single dense layer with softmax activation was trained for the same number of epochs as the encoder was trained for. The same augmentation regime was used for training the classification head as was used for training the encoder. For slide-level tasks, representations are aggregated by taking the minimum, maximum, and mean for each representation element and concatenating these vectors, yielding a slide representation of length 6144. This simple aggregation method was chosen due to its minimal footprint, enabling a fair evaluation of the representation quality of each encoder. An Adam optimiser \cite{kingma2014adam} is used and all code was written in Python using TensorFlow and trained on either Nvidia RTX6000, A6000 or A100 GPUs. See Appendix \ref{sec:comp_times} for a full analysis of computational times.

Self-supervised baseline comparisons are privileged and unprivileged Siamese networks trained in exactly the same conditions as TriDeNT \Neptune. These approaches are the existing state of the art in LUPI \cite{farndale2023more,girdhar2023imagebind,xu2023asymmetric}.

Supervised comparisons were trained using the same architecture and hyperparameters as above, with the same classifier head architecture used for evaluation. Supervised models were pretrained, then their classifier heads were discarded, and a new classifier head was trained, to ensure a fair comparison with self-supervised models. For direct gene prediction models, the same procedure was followed, except in the initial phase the number of outputs of the classifier head was the number of genes whose counts are being predicted, with a linear activation. The loss used was mean squared error.

To analyse the learned representations, we used GradCAM \cite{selvaraju2017grad}, with the modification that instead of only calculating the gradients for one class, as in the standard implementation, we treated each element of the representation as a class and sum over these to produce a final composite figure. These could be analysed on a more granular level, however, we leave this interesting avenue for future work.

% LF: Normalise
\begin{table}[t]
    \centering
    \caption{Computational Speed (ms/step). SegPath training is split across two Nvidia RTX6000 GPUs, and ALS-ST training is carried out on a single Nvidia A100 GPU.}
    \label{tab:compute_time}
    \scalebox{0.8}{\begin{tabular}{ccccccc}
        \toprule
        & & \multicolumn{2}{c}{SegPath} & \multicolumn{2}{c}{ALS-ST} \\
        Method & Privileged & VICReg & InfoNCE & VICReg & InfoNCE \\
        \midrule
        \greycell TriDeNT \Neptune & \greycell \cmark & \greycell 566 & \greycell 542 & \greycell 525 & \greycell 536 \\
        \multirow{2}{*}{Siamese} & \cmark & 438 & 427 & 270 & 278 \\
        & \xmark & 352 & 340 & 464 & 511 \\
        \bottomrule
    \end{tabular}}
\end{table}

\subsection{Comparison of Computational Efficiency}
\label{sec:comp_times}

TriDeNT \Neptune ~does not require orders of magnitude more computational time than comparable 2-branch methods, as shown in Table \ref{tab:compute_time}. We find that models whose privileged input has the same shape as the primary input, such as those for SegPath, see a greater difference in similar computational speeds. For models with smaller inputs or fewer parameters in their privileged branch, such as those for the ALS-ST dataset, computational speeds are similar for TriDeNT \Neptune ~and unprivileged Siamese models.

\begin{figure*}[!t]
\centering
    \begin{subfigure}{0.4\textwidth}
    \scalebox{0.75}{\begin{tabular}{cccc}
    \toprule
        Change & Privileged & VICReg & InfoNCE \\
    \midrule
       \greycell - & \greycell\cmark & \greycell\B 0.9175 & \greycell\B 0.9075 \\
    \midrule
        Unprivileged & \xmark & 0.8361 & 0.9033 \\
    \midrule
        \multirow{2}{*}{\makecell{Weights not shared\\between any branches}} & \cmark & 0.8887 & 0.8910 \\
        & \xmark & 0.8193 & 0.8548 \\
    \midrule
         No projectors & \cmark & 0.8249 & 0.5155 \\
         % & Siamese & \cmark & \B 0.8410 & 0.4566 & 0.7177 & \B 0.7822 \\
         % & Siamese & \xmark & 0.7878 & \B 0.8478 & 0.7191 & \B 0.8972 \\
     \bottomrule
    \end{tabular}}
    \vspace{5mm}
    \caption{}
    \label{tab:model-ablations}
    \end{subfigure}
    \begin{subfigure}{0.48\textwidth}
    \scalebox{0.75}{\begin{tabular}{cccccc}
    \toprule
        Encoder & Loss & \greycell TriDeNT \Neptune & \makecell{Siamese\\(Privileged)} & \makecell{Siamese\\(Unprivileged)} \\
    \midrule
    \multirow{2}{*}{DenseNet121} & VICReg & \greycell \B 0.8991 & 0.8116 & 0.8373 \\
    & InfoNCE & \greycell\B 0.9299 & 0.8927 & 0.8711 \\
    \midrule
    \multirow{2}{*}{EfficientNetB0} & VICReg & \greycell \B 0.8101 & 0.5114 & 0.7659 \\
    & InfoNCE & \greycell\B 0.8381 & 0.5154 & 0.8257 \\
    \midrule
    \multirow{2}{*}{Xception} & VICReg & \greycell \B 0.9057 & 0.7513 & 0.8614 \\
    & InfoNCE & \greycell \B 0.9225 & 0.8601 & 0.8927 \\
    \midrule
    \multirow{2}{*}{ViT-L32} & VICReg & \greycell\B 0.8533 & 0.4680 & 0.8444\\
    & InfoNCE & \greycell\B 0.8937 & 0.5897 & 0.8527 \\
    \bottomrule
    \end{tabular}}
    \caption{}
    \label{tab:encoder-ablations}
    \end{subfigure}
    \caption{(a) TriDeNT \Neptune ~Model ablations trained on SegPath MNDA and tested on the NCT tissue classification task. (b) Encoder ablations trained on SegPath MNDA and tested on the NCT tissue classification task. Models are Keras implementations of DenseNet121 \cite{huang2017densely}, EfficientNetB0 \cite{tan2019efficientnet} Xception \cite{chollet2017xception}, and ViT-L32 \cite{dosovitskiy2020image}.}
    \label{fig:enter-label}
\end{figure*}

\subsection{Ablations}

We note from the ablations given in Figure \ref{tab:model-ablations} that performance is not significantly reduced by the model not sharing weights between encoders $f^1$ and $f^2$, and projectors $g^1$ and $g^2$, as these models will likely converge to similar minima. With the weights shared between these branches, but not with $f^*$ and $g^*$, the model learns primarily to extract features which are invariant to augmentation, while also finding features which are shared with the supplementary data source. Without weight sharing, this effect is weaker, as the models' hidden representations will differ, causing more sensitivity to the augmentations applied. Of course, there is also a significant increase in computational power required to train three encoders in parallel, so TriDeNT \Neptune~ is more efficient and better performing than this alternative formulation.

In Figure \ref{tab:encoder-ablations}, we demonstrate that TriDeNT \Neptune ~is robust to the choice of backbone encoder, with a set of common models. In Table \ref{tab:trident-unprivileged}, we provide extended ablation results for unprivileged TriDeNT \Neptune ~(the primary input supplied to all branches, with no privileged input).

\begin{table}[t]
    \centering
    \caption{SegPath dataset image characteristics. Non-Zero is the fraction of patches for a given stain which contain any non-zero pixels. A negligible number of patches were found to contain only non-zero values. Mean denotes the mean value for each patch, and StD denotes the standard deviation.}
    \label{tab:segpath-characteristics}
    \begin{tabular}{cccccc}
    \toprule
        \multirow{2}{*}{Stain} & \multicolumn{3}{c}{Image} & \multicolumn{2}{c}{Gradients} \\
         & Non-Zero & Mean & StD & Mean ($\times10^{-3}$) & StD ($\times10^{-3}$) \\
    \midrule
        MNDA & 0.3950 & 0.0066 & 0.0137 & 0.5787 & 1.2211 \\
        MIST1 & 0.1255 & 0.0029 & 0.0125 & 0.2520 & 1.0657 \\
        ERG & 0.2320 & 0.0032 & 0.0088 & 0.2691 & 0.7019 \\
        CD3CD20 & 0.3599 & 0.0112 & 0.0331 & 1.1471 & 3.3579 \\
        CD45RB & 0.4701 & 0.0154 & 0.0349 & 1.5070 & 3.3787 \\
        CD235A & 0.2703 & 0.0104 & 0.0434 & 0.8659 & 2.6670 \\
        pan-CK & 0.7015 & 0.1210 & 0.1920 & 5.5446 & 7.5578 \\
        $\alpha$SMA & 0.7743 & 0.2766 & 0.2989 & 7.4813 & 8.4139 \\
    \bottomrule
    \end{tabular}
\end{table}

\section{Differences in Performance Between SegPath Stains}
\label{sec:segpath-characteristics}

We observe significant differences in performance between SegPath stains on different tasks, although some models appear to achieve better performance across all tasks than others. This is in part due to some stains being more relevant to the task than others. For example, pan-CK is broadly relevant to most cancer-related tasks as it marks epithelial origin tumour cells. In contrast, MIST1 is a highly specific stain which will be less useful for tasks such as tissue classification and metastasis detection.

Additionally, we observe that there are major differences in performance between privileged Siamese models depending on the privileged stain. This can be as large as 46 percentage points difference between MNDA and $\alpha$SMA (see Figure \ref{fig:segpath_res_tri_priv} and Table \ref{tab:segpath}). This is despite lower performing stains containing a lot of useful semantic information. Therefore, it is useful to consider the content of these images to enable understanding of why there can be such large disparities between models of different stains, and why this is not the case with TriDeNT \Neptune. Table \ref{tab:segpath-characteristics} shows the characteristics of the privileged images in the subset for each stain. We see firstly that there is a large disparity between the proportion of images in each subset which are blank (not non-zero), and that this is distributed roughly along the lines of privileged Siamese performance on the tissue classification task. We have already seen in Section \ref{sec:harmful-privileged-info} that pairing with blank privileged information can seriously harm performance for privileged Siamese models, while TriDeNT \Neptune ~largely mitigates this. We also see similar differences in the mean pixel value. As information could be more accurately characterised as \emph{changes} between stained and unstained regions in an image, we also provide these values for gradients in the images, which show the same conclusions.

In \cite{jing2021understanding}, Jing \emph{et al.} demonstrate that that dimensional collapse is observed where the variance caused by data augmentation is greater than the variance caused by the data distribution. This implies that images with lower feature variance will be more prone to collapsed dimensions, resulting in worse downstream performance. In our case, we see that the stains with worse downstream performance are generally those with lower variance in their privileged images. Despite this, TriDeNT \Neptune ~is able to mitigate this shortcoming, as the objective discourages TriDeNT \Neptune ~from learning any of the collapsed dimensions from the privileged model branch. Instead TriDeNT \Neptune ~can retain only the useful information, meaning that stains with less content and/or lower variance can be used to train models with more content/higher variance. For example, on the TIL task, the CD3CD20 and CD45RB models outperform the pan-CK and $\alpha$SMA models, despite the former having much lower content/variance than the latter, because the privileged information is more relevant to the task.

\begin{table*}[t]
    \centering
    \caption{Performance for TriDeNT \Neptune ~with $N=M=2$ (2 primary branches and 2 privileged branches). Models trained on PanNuke using the VICReg loss as in Table \ref{tab:pannuke_results}.}
    \label{tab:vicreg-4-branch}
    \scalebox{1}{\begin{tabular}{ccccccc}
    \toprule
         Method & Privileged Branches & NCT & PanNuke & Singapore & MHIST \\
    \midrule
         4-Branch TriDeNT \Neptune & 2 & 0.7226 & 0.8760 & \B 0.7979 & 0.7144 \\
         \greycell TriDeNT \Neptune & \greycell 1 & \greycell\B 0.7337 & \greycell\B 0.9106 & \greycell 0.7975 & \greycell\B 0.7523 \\
         \multirow{2}{*}{Siamese} & 1 & 0.6000 & 0.8274 & 0.7106 & 0.6530 \\
         & 0 & 0.7301 & 0.8682 & 0.7754 & 0.7421 \\
     \bottomrule
    \end{tabular}}
\end{table*}

\section{Additional Privileged Branches May Harm Performance}
\label{sec:additional-privileged-branches}

We find that using more than one privileged branches for TriDeNT \Neptune ~may hinder performance. By introducing an additional privileged branch, there is an incentive for the model to learn features which are not present in the primary information, as these can be optimised to minimise invariance between both privileged branches. With only one privileged branch, this is not possible, and privileged branches are optimised only to find features which can be predicted from the primary information. Table \ref{tab:vicreg-4-branch} shows that this is the case with the PanNuke dataset. We find that adding an additional branch deteriorates performance on three out of four tasks, with a negligible increase in performance on the Singapore task. On the NCT and MHIST tasks, performance is reduced below the level of unprivileged Siamese learning. It may be the case that TriDeNT \Neptune ~can be adapted to accommodate multiple branches, such as by not imposing a loss term between privileged branches, although we leave this as an avenue for future work.

\setcounter{figure}{0}
\renewcommand{\figurename}{Fig.}
\renewcommand{\thefigure}{S\arabic{figure}}
\setcounter{table}{0}
\renewcommand{\tablename}{Table}
\renewcommand{\thetable}{S\arabic{table}}
\setcounter{algorithm}{0}
\renewcommand{\thealgorithm}{S\arabic{algorithm}}
\setcounter{section}{0}
\renewcommand{\thesection}{S\arabic{section}}

\clearpage

\begin{table*}[h]
    \centering
    \caption{Abbreviations used in the main text.}
    \label{tab:abbreviations}
    \begin{tabular}{ll}
    \toprule
    Abbreviation & Meaning \\
    \midrule
        BCI & Breast Cancer Immunohistochemisty \\
        CNN & Convolutional Neural Network \\
        GradCAM & Gradient-Weighted Class Activation Mapping \\
        H\&E & Haemotoxylin and Eosin \\
        HER2 & Human Epidermal growth factor Receptor 2 \\
        IF & ImmunoFluorescence \\
        IHC & ImmunoHistoChemistry \\
        InfoNCE & [Mutual Information] Noise-Contrastive Estimation \\
        LEEP & Loop Electrosurgical Excision Procedure \\
        LUPI & Learning Using Privileged Information \\
        MHIST & Minimal Histopathology Image Analysis Dataset \\
        mpp & microns per pixel \\
        NCT & National Center for Tumor diseases \\
        RNA & Ribonucleic Acid \\
        PANDA & Prostate cANcer graDe Assessment \\
        PGD & Projected Gradient Descent \\
        px & pixels \\
        SSL & Self-Supervised Learning \\
        SVM & Support Vector Machine \\
        TIL & Tumour Infiltrating Lymphocytes \\
        TriDeNT \Neptune ~& Triple Deep Network Training \\
        UMAP & Uniform Manifold Approximation and Projection \\
        VICReg & Variance Invariance Covariance Regularization \\
    \midrule
    SegPath Stains \\
    \midrule
    $\alpha$SMA & Alpha Smooth Muscle Actin (Smooth Muscle/Myofibroblasts) \\
    CD235a & Glycophorin A (Red Blood Cells) \\
    CD3CD20 & Cluster of differentiation 3/20 (Lymphocytes) \\
    CD45RB & Protein Tyrosine Phosphatase, Receptor Type C (Leukocytes) \\
    ERG & ETS-Related Gene (Endothelium) \\
    MIST1 & Muscle, Intestine, and Stomach Expression 1 (Plasma Cells) \\
    MNDA & Myeloid Cell Nuclear Differentiation Antigen (Myeloid Cells) \\
    pan-CK & Pan-Cytokeratin (Epithelium) \\
    \midrule
    NCT Tissue Types \\
    \midrule
    ADI & Adipose \\
    BACK & Background \\
    DEB & Debris \\
    LYM & Lymphocytes \\
    MUC & Mucus \\
    MUS & Smooth Muscle \\
    NORM & Normal Colon Mucosa \\
    STR & Cancer-Associated Stroma \\
    TUM & Colorectal Adenocarcinoma Epithelium \\
    \bottomrule
    \end{tabular}
\end{table*}

\clearpage

\begin{algorithm*}[h]
\caption{TriDeNT \Neptune ~Pseudocode}
\label{alg:pseudocode}
\begin{algorithmic}[1]
\LineComment{$x, x^*$: primary/privileged input}
\LineComment{$a, a^*$: primary/privileged augmentation operations}
\LineComment{$f, f^*$: primary/privileged encoders}
\LineComment{$g, g^*$: primary/privileged projectors}
\LineComment{$L$: two-branch self-supervised loss}
\LineComment{$M$: iterations}
\\
\For{$i=1,2,\ldots,M$}
\Comment{Encoder Pretraining}
\State $x^1, x^2 \gets a(x), a(x)$
\State $z^1, z^2 \gets f(x^1), f(x^2)$
\Comment{Representations}
\State $z^* \gets f^*(a^*(x^*))$
\State $e^1, e^2 \gets g(z^1), g(z^2)$
\Comment{Embeddings}
\State $e^* \gets g^*(z^*)$
\State Loss $\gets L(e^1, e^2) + L(e^1, e^*) + L(e^2, e^*)$
\EndFor
\LineComment{$c$: classifier head}
\LineComment{CE: cross entropy loss}
\For{$i=1,2,\ldots,M$}
\Comment{Classifier Head Training}
\State{$z \gets f(a(x))$}
\State{Predictions $\gets c(z)$}
\State{Loss $\gets$ CE(Predictions, Labels)}
\EndFor
\end{algorithmic}
\end{algorithm*}

\clearpage

\begin{table*}[!h]
    \caption{Image augmentations and probability of application. Augmentations are randomly applied at each training iteration, with the given probability.}
    \label{tab:image_augmentations}
    \centering
    \scalebox{1}{\begin{tabular}{lc}
    \toprule
    Augmentation & Probability\\
    \midrule
    Flip (left/right/up/down) & 1.0 \\
    Crop (scale 75\% to 100\%) & 1.0 \\
    Gaussian Background Noise & 0.3\\
    Rotation & 0.4 \\
    Solarize & 0.3 \\
    Colour Jitter & 1.0\\
    \bottomrule
    \end{tabular}}
\end{table*}

\begin{table*}[!h]
    \caption{Gene count augmentations and probability of application. \emph{Mask Genes} masks a proportion of genes in each sample with probability of masking any given gene randomly sampled at each iteration from the given range. \emph{Random Updates} randomly shuffles a subset of genes each iteration, with probability of any given gene being shuffled sampled at each iteration from the given range. Gaussian noise is applied with mean 0 and unit variance.}
    \label{tab:gene_augmentations}
    \centering
    \scalebox{1}{\begin{tabular}{lc}
    \toprule
    Augmentation & Probability\\
    \midrule
    Mask Genes & [0.0-0.2] \\
    Random Updates & [0.0-0.1] \\
    Gaussian Noise & 1.0 \\
    \bottomrule
    \end{tabular}}
\end{table*}

\clearpage

\begin{table*}[!ht]
    \centering
    \caption{Results on four tasks for models trained on each subset (stain) from the SegPath dataset. Each subset consists of H\&E patches paired with a binarised IF mask generated automatically from the corresponding stain. In unprivileged examples, the model was only trained on the H\&E patches from that subset, while in privileged examples, one branch received the H\&E patches while the other received the IF masks. Value marked $\dagger$ from \cite{farndale2024synthetic}.}
    \label{tab:segpath}
    \scalebox{1}{\begin{tabular}{cllabc|abc}
    \toprule
        & \multirow{3}{*}{\diagbox[height=3\line]{Stain}{}} & Loss: & \multicolumn{3}{c}{VICReg} & \multicolumn{3}{c}{InfoNCE} \\
        & & Method: & TriDeNT \Neptune & \multicolumn{2}{c}{Siamese} & TriDeNT \Neptune & \multicolumn{2}{c}{Siamese} \\
        Task & & Privileged: & \cmark & \cmark & \xmark & \cmark & \cmark & \xmark \\
    \midrule
        \multirow{8}{*}{\makecell{\B NCT:\\Tissue Type\\Classification\\ \\ \B Supervised:\\0.9245}} & MNDA & & \B 0.9169 & 0.4566 & 0.8616 & \B 0.9075 & 0.7822 & 0.8972 \\
        & MIST1 & & \B 0.8824 & 0.5098 & 0.8357 & \B 0.8644 & 0.8084 & 0.8547 \\
        & ERG & & \B 0.9019 & 0.5781 & 0.8749 & \B 0.9174 & 0.8173 & 0.8763 \\
        & CD3CD20 & & \B 0.8982 & 0.6625 & 0.8694 & \B 0.8681 & 0.8119 & 0.8524 \\
        & CD45RB & & \B 0.8914 & 0.5911 & 0.8322 & 0.8683 & 0.8554 & \B 0.8701 \\
        & CD235a & & 0.8887 & 0.7102 & \B 0.8929 & 0.8846 & 0.8221 & \B 0.8981 \\
        & $\alpha$SMA & & \B 0.9273 & 0.9186 & 0.8570 & \B 0.9438 & 0.9190 & 0.8845 \\
        & pan-CK & & \B 0.9179 & 0.8672 & 0.8423 & \B 0.9291 & 0.9074 & 0.8766 \\
    \midrule
        \multirow{8}{*}{\makecell{\B PanNuke:\\Neoplastic\\Cell\\Classification\\ \\ {\B Supervised:}\\0.8901}} & MNDA & & \B 0.9288 & 0.8327 & 0.9156 & \B 0.9369 & 0.8971 & 0.9277 \\
        & MIST1 & & \B 0.9206 & 0.8231 & 0.9156 & \B 0.9295 & 0.8730 & 0.9282 \\
        & ERG & & \B 0.9315 & 0.6821 & 0.9302 & \B 0.9391 & 0.8568 & 0.9308 \\
        & CD3CD20 & & \B 0.9261 & 0.8168 & 0.9161 & \B 0.9365 & 0.8760 & 0.9308 \\
        & CD45RB & & \B 0.9406 & 0.8680 & 0.9206 & \B 0.9396 & 0.9098 & 0.9369 \\
        & CD235a & & \B 0.9197 & 0.8050 & 0.9166 & 0.9198 & 0.8537 & \B 0.9273 \\   
        & $\alpha$SMA & & \B 0.9447 & 0.9088 & 0.9147 & \B 0.9374 & 0.9343 & 0.9308 \\
        & pan-CK & & \B 0.9469 & 0.9293 & 0.9229 & \B 0.9588 & 0.9400 & 0.9269 \\
    \midrule
        \multirow{8}{*}{\makecell{\B Singapore:\\Gland\\Malignancy\\Detection\\\\\B Supervised:\\0.9103}} & MNDA & & \B 0.8252 & 0.6679 & 0.7864 & \B 0.8578 & 0.7665 & 0.7665 \\
        & MIST1 & & \B 0.8491 & 0.6611 & 0.7198 & \B 0.8489 & 0.6412 & 0.7681 \\
        & ERG & & \B 0.8470 & 0.6381 & 0.8303 & 0.8512 & 0.6097 & \B 0.8597 \\
        & CD3CD20 & & \B 0.8019 & 0.6853 & 0.7749 & \B 0.8313 & 0.7057 & 0.8029 \\
        & CD45RB & & \B 0.7869 & 0.7212 & 0.7839 & \B 0.8416 & 0.7749 & 0.7714 \\
        & CD235a & & \B 0.8214 & 0.6893 & 0.7475 & 0.8315 & 0.7670 & \B 0.8679 \\
        & $\alpha$SMA & & \B 0.8869 & 0.8198 & 0.8261 & \B 0.8676 & 0.8641 & 0.8127 \\
        & pan-CK & & \B 0.8690 & 0.8559 & 0.8026 & \B 0.8601 & 0.8599 & 0.7860 \\
    \midrule
        \multirow{8}{*}{\makecell{\B MHIST:\\Precancerous\\Polyp\\Detection\\\\\B Supervised:\\0.7042}} & MNDA & & \B 0.7666 & 0.6766 & 0.7195 & \B 0.7748 & 0.7134 & 0.7431 \\
        & MIST1 & & \B 0.7400 & 0.5476 & 0.7083 & 0.7451 & 0.7124 & \B 0.7533 \\
        & ERG & & \B 0.7533 & 0.6223 & 0.7073 & \B 0.7789 & 0.6622 & 0.7544 \\
        & CD3CD20 & & \B 0.7554 & 0.6438 & 0.7318 & \B 0.7769 & 0.6909 & 0.7380 \\
        & CD45RB & & \B 0.7441 & 0.7114 & 0.7226 & \B 0.7697 & 0.7441 & 0.7544 \\
        & CD235a & & \B 0.7523 & 0.6684 & 0.7247 & \B 0.7799 & 0.6950 & 0.7646 \\
        & $\alpha$SMA & & \B 0.7861 & 0.7636 & 0.7021 & \B 0.8199 & 0.7687 & 0.7595 \\
        & pan-CK & & \B 0.8270 & 0.7953 & 0.7267 & 0.7963 & \B 0.8025 & 0.7595 \\
    \midrule
        \multirow{8}{*}{\makecell{\B TIL:\\Tumour Infiltrating\\Lymphocyte\\Detection\\\\\B Supervised:\\0.9216}} & MNDA & & \B 0.9329 & 0.8979 & 0.9125 & \B 0.9357 & 0.9271 & 0.9163 \\
        & MIST1 & & \B 0.9262 & 0.8880 & 0.9099 & \B 0.9347 & 0.9154 & 0.9215 \\
        & ERG & & \B 0.9195 & 0.8411 & 0.9077 & \B 0.9276 & 0.8919 & 0.9160 \\
        & CD3CD20 & & \B 0.9336 & 0.9084 & 0.9071 & \B 0.9380 & 0.9206 & 0.9196 \\
        & CD45RB & & \B 0.9389 & 0.9256 & 0.9001 & \B 0.9373 & 0.9340 & 0.9150 \\
        & CD235a & & \B 0.9008 & 0.8527 & 0.8995 & 0.9034 & 0.8666 & \B 0.9117 \\
        & $\alpha$SMA & & \B 0.9320 & 0.9201 & 0.8998 & \B 0.9302 & 0.9207 & 0.9081 \\
        & pan-CK & & \B 0.9324 & 0.9137 & 0.9015 & \B 0.9334 & 0.9257 & 0.9151 \\
    \midrule
        \multirow{8}{*}{\makecell{\B Camelyon:\\Out-of-Distribution\\Metastasis\\Detection\\\\\B Supervised:\\$0.8440^\dagger$}} & MNDA & & \B 0.7211 & 0.5340 & 0.5065 & \B 0.7689 & 0.6885 & 0.5460 \\
        & MIST1 & & \B 0.6601 & 0.5320 & 0.5430 & \B 0.6952 & 0.3655 & 0.5916 \\
        & ERG & & \B 0.6357 & 0.5052 & 0.5170 &  0.5756 & \B 0.5834 & 0.5429 \\
        & CD3CD20 & &  0.5996 & \B 0.6117 & 0.5325 &  0.6911 & \B 0.7368 & 0.5442 \\
        & CD45RB & & \B 0.7844 & 0.6376 & 0.4891 & \B 0.6646 & 0.6503 & 0.5866 \\
        & CD235a & &  0.6430 & \B 0.7950 & 0.5327 &  0.5920 & \B 0.6427 & 0.5470 \\
        & $\alpha$SMA & & \B 0.8027 & 0.6905 & 0.5091 & \B 0.6782 & 0.5972 & 0.5755 \\
        & pan-CK & & \B 0.8068 & 0.6373 & 0.5529 & \B 0.6856 & 0.6532 & 0.5225 \\
    \bottomrule
    \end{tabular}}
\end{table*}

\clearpage

\begin{table*}[!ht]
    \centering
    \caption{Slide-level tasks.}
    \label{tab:segpath-2}
    \scalebox{1}{\begin{tabular}{cllabc|abc}
    \toprule
        & \multirow{3}{*}{\diagbox[height=3\line]{Stain}{}} & Loss: & \multicolumn{3}{c}{VICReg} & \multicolumn{3}{c}{InfoNCE} \\
        & & Method: & TriDeNT \Neptune & \multicolumn{2}{c}{Siamese} & TriDeNT \Neptune & \multicolumn{2}{c}{Siamese} \\
        Task & & Privileged: & \cmark & \cmark & \xmark & \cmark & \cmark & \xmark \\
    \midrule
        \multirow{8}{*}{\makecell{\B PANDA:\\Prostate\\ISUP\\Grading\\\\\\}} & MNDA & & \B 0.7240 & 0.5828 & 0.6824 & \B 0.7194 & 0.6907 & 0.6830 \\
        & MIST1 & & \B 0.6800 & 0.5361 & 0.6688 & \B 0.7111 & 0.6432  & 0.6864 \\
        & ERG & & 0.6875 & 0.3773 & \B 0.6908 & \B 0.7005 & 0.5722 & 0.6990 \\
        & CD3CD20 & & \B 0.7040 & 0.5074 & 0.6821 & \B 0.7235 & 0.6641 & 0.6961 \\
        & CD45RB & & \B 0.7318 & 0.6392 & 0.6686 & \B 0.7375 & 0.7191 & 0.6845 \\
        & CD235a & & \B 0.7012 & 0.5392 & 0.6776 & \B 0.7174 & 0.5963 & 0.6810 \\
        & $\alpha$SMA & & \B 0.7287 & 0.6728 & 0.6625 & \B 0.7472 & 0.7261 & 0.6839 \\
        & pan-CK & & \B 0.7222 & 0.6855 & 0.6775 & \B 0.7379 & 0.7069 & 0.6842 \\
    \midrule
        \multirow{8}{*}{\makecell{\B IMP-1K:\\Colorectal\\Dysplasia\\Detection}} & MNDA & & \B 0.7411 & 0.7111 & 0.6700 & \B 0.7322 & 0.6811 & 0.6733 \\
        & MIST1 & & \B 0.7211 & 0.5222 & 0.6711 & \B 0.7078 & 0.5967 & 0.6800 \\
        & ERG & & \B 0.7367 & 0.5256 & 0.6878 & \B 0.7478 & 0.6978 & 0.6889 \\
        & CD3CD20 & & \B 0.6744 & 0.5611 & 0.6511 & \B 0.7478 & 0.6522 & 0.6289 \\
        & CD45RB & & \B 0.6811 & 0.6633 & 0.6744 & 0.7033 & \B 0.7700 & 0.7078 \\
        & CD235a & & \B 0.7478 & 0.6211 & 0.7000 & \B 0.7400 & 0.5833 & 0.6689 \\
        & $\alpha$SMA & & 0.7689 & \B 0.7733 & 0.6744 & \B 0.8233 & 0.8089 & 0.6789 \\
        & pan-CK & & \B 0.7733 & 0.7344 & 0.6933 & \B 0.7622 & 0.7589 & 0.7122 \\
    % \midrule
    %     & Average & & \B 0.7306 & 0.6390 & 0.6778 & \B 0.7456 & 0.6936 & 0.6799 \\
    \midrule
        \multirow{8}{*}{\makecell{\B IMP-4K:\\Colorectal\\Dysplasia\\Detection}} & MNDA & & \B 0.8578 & 0.7578 & 0.8367 & \B 0.8667 & 0.8489 & 0.8411 \\
        & MIST1 & & \B 0.8444 & 0.7022 & 0.8278 & \B 0.8689 & 0.7956 & 0.8456 \\
        & ERG & & \B 0.8722 & 0.6533 & 0.8289 & \B 0.8522 & 0.8211 & 0.8344 \\
        & CD3CD20 & & \B 0.8478 & 0.7244 & 0.8411 & \B 0.8478 & 0.8100 & 0.8467 \\
        & CD45RB & & \B 0.8389 & 0.8022 & \B 0.8389 & \B 0.8544 & 0.8433 & 0.8456 \\
        & CD235a & & 0.8178 & 0.7689 & \B 0.8400 & \B 0.8578 & 0.7978 & 0.8367 \\
        & $\alpha$SMA & & \B 0.8767 & 0.8467 & 0.8278 & \B 0.8844 & 0.8811 & 0.8633 \\
        & pan-CK & & \B 0.8878 & 0.8700 & 0.8244 & 0.8733 & \B 0.8867 & 0.8500 \\
    \midrule
        \multirow{8}{*}{\makecell{\B IMP-Cervix:\\Cervical\\Dysplasia\\Detection}} & MNDA & & \B 0.6667 & 0.6433 & 0.6550 & \B 0.7100 & 0.6650 & 0.6500 \\
        & MIST1 & & 0.6583 & 0.6617 & \B 0.6683 & 0.6617 & \B 0.6650 & 0.6600 \\
        & ERG & & \B 0.6950 & 0.6150 & 0.6867 & 0.6717 & 0.6083 & \B 0.6800 \\
        & CD3CD20 & & \B 0.6833 & 0.6367 & 0.6617 & \B 0.7067 & 0.6250 & 0.6933 \\
        & CD45RB & & 0.6650 & 0.6667 & \B 0.6767 & \B 0.6767 & 0.6633 & 0.6633 \\
        & CD235a & & \B 0.6817 & 0.6500 & 0.6800 & \B 0.7117 & 0.6550 & 0.6700 \\
        & $\alpha$SMA & & \B 0.7183 & 0.6467 & 0.6617 & \B 0.7333 & 0.7200 & 0.6633 \\
        & pan-CK & & \B 0.7150 & 0.6933 & 0.6617 & \B 0.7183 & 0.6917 & 0.6367 \\
    \bottomrule
    \end{tabular}}
\end{table*}

\clearpage

\begin{table*}[!ht]
    \centering
    \caption{Unprivileged TriDeNT \Neptune. As in Figure \ref{tab:segpath}, including unprivileged TriDeNT \Neptune ~results on NCT tasks for models trained on each subset (stain) from the SegPath dataset.}
    \label{tab:trident-unprivileged}
    \scalebox{0.95}{\begin{tabular}{cllacbc|acbc}
    \toprule
        & \multirow{3}{*}{\diagbox[height=3\line]{Stain}{}} & Loss: & \multicolumn{4}{c}{VICReg} & \multicolumn{4}{c}{InfoNCE} \\
        & & Method: & \multicolumn{2}{c}{Trident \Neptune} & \multicolumn{2}{c}{Siamese} & \multicolumn{2}{c}{Trident \Neptune} & \multicolumn{2}{c}{Siamese} \\
        Task & & Asymmetric: & \cmark & \xmark & \cmark & \xmark & \cmark & \xmark & \cmark & \xmark \\
    \midrule
        \multirow{8}{*}{\makecell{\B NCT:\\Tissue\\Type\\Classification\\ \\ \B Supervised:\\0.9245}} & MNDA & &  \B 0.9169 & 0.8361 & 0.4566 & 0.8616 & \B 0.9075 & 0.9033 & 0.7822 & 0.8972 \\
        & MIST1 & & \B 0.8824 & 0.8428 & 0.5098 & 0.8357 & 0.8644 & \B 0.8689 & 0.8084 & 0.8547 \\
        & ERG & & \B 0.9019 & 0.8378 & 0.5781 & 0.8749 & \B 0.9174 & 0.8679 & 0.8173 & 0.8763 \\
        & CD3CD20 & & \B 0.8982 & 0.8449 & 0.6625 & 0.8694 & \B 0.8681 & 0.8630 & 0.8119 & 0.8524 \\
        & CD45RB & & \B 0.8914 & 0.8484 & 0.5911 & 0.8322 & 0.8683 & \B 0.8869 & 0.8554 & 0.8701 \\
        & CD235a & & 0.8887 & 0.8625 & 0.7102 & \B 0.8929 & 0.8846 & 0.8969 & 0.8221 & \B 0.8981 \\
        & $\alpha$SMA & & \B 0.9273 & 0.8444 & 0.9186 & 0.8570 & \B 0.9438 & 0.8782 & 0.9190 & 0.8845 \\
        & pan-CK & & \B 0.9179 & 0.8455 & 0.8672 & 0.8423 & \B 0.9291 & 0.8814 & 0.9074 & 0.8766 \\
    \bottomrule
    \end{tabular}}
\end{table*}

\clearpage

\begin{table*}[!ht]
    \centering
    \caption{NCT results from Figure \ref{fig:dataset-sizes}. Models were the same as those evaluated in Figure \ref{tab:segpath}}
    \label{tab:nct-segpath-dataset-size-data}
    \scalebox{1}{\begin{tabular}{cllabc}
    \toprule
        & \multirow{3}{*}{\diagbox[height=3\line]{Stain}{}} & Loss: & \multicolumn{3}{c}{VICReg} \\
        & & Method: & TriDeNT \Neptune & \multicolumn{2}{c}{Siamese} \\
        Dataset \%ge & & Privileged: & \cmark & \cmark & \xmark \\
    \midrule
    \multirow{8}{*}{\makecell{\B 50\%\\\\\B Supervised:\\0.9245}} & MNDA & & \B 0.9284 & 0.7544 & 0.8901 \\
      & MIST1 & & \B 0.8888 & 0.7152 & 0.8766 \\
      & ERG & & \B 0.9242 & 0.5965 & 0.8770 \\
      & CD3CD20 & & \B 0.9193 & 0.7783 & 0.8890 \\
      & CD45RB & & \B 0.9131 & 0.8508 & 0.8720 \\
      & CD235a & & \B 0.9022 & 0.8122 & 0.8998 \\
      & $\alpha$SMA & & \B 0.9423 & 0.9296 & 0.8845 \\
      & pan-CK & & \B 0.9394 & 0.9093 & 0.8779 \\
    \midrule
    \multirow{8}{*}{\makecell{\B 20\%\\\\\B Supervised:\\0.6082}} & MNDA & & \B 0.9202 & 0.7456 & 0.8832 \\
      & MIST1 & & \B 0.8846 & 0.7093 & 0.8660 \\
      & ERG & & \B 0.9157 & 0.5887 & 0.8799 \\
      & CD3CD20 & & \B 0.8994 & 0.7661 & 0.8743 \\
      & CD45RB & & \B 0.9061 & 0.8289 & 0.8590 \\
      & CD235a & & \B 0.8975 & 0.8040 & 0.8824 \\
      & $\alpha$SMA & & \B 0.9359 & 0.9288 & 0.8753 \\
      & pan-CK & & \B 0.9335 & 0.9092 & 0.8713 \\
    \midrule
    \multirow{8}{*}{\makecell{\B 10\%\\\\\B Supervised:\\0.9413}} & MNDA & & \B 0.9207 & 0.7332 & 0.8799 \\
      & MIST1 & & \B 0.9020 & 0.6933 & 0.8622 \\
      & ERG & & \B 0.9093 & 0.5763 & 0.8779 \\
      & CD3CD20 & & \B 0.9080 & 0.7704 & 0.8683 \\
      & CD45RB & & \B 0.9085 & 0.8268 & 0.8559 \\
      & CD235a & & \B 0.8917 & 0.8009 & 0.8871 \\
      & $\alpha$SMA & & \B 0.9405 & 0.9282 & 0.8768 \\
      & pan-CK & & \B 0.9347 & 0.9047 & 0.8747 \\
    \midrule
    \multirow{8}{*}{\makecell{\B 5\%\\\\\B Supervised:\\0.92}} & MNDA & & \B 0.8998 & 0.7259 & 0.8605 \\
      & MIST1 & & \B 0.8968 & 0.6912 & 0.8405 \\
      & ERG & & \B 0.9093 & 0.5593 & 0.8587 \\
      & CD3CD20 & & \B 0.9001 & 0.7693 & 0.8568 \\
      & CD45RB & & \B 0.9062 & 0.8279 & 0.8339 \\
      & CD235a & & \B 0.8740 & 0.7877 & 0.8694 \\
      & $\alpha$SMA & & \B 0.9285 & 0.9131 & 0.8644 \\
      & pan-CK & & \B 0.9223 & 0.9015 & 0.8533 \\
    \midrule
    \multirow{8}{*}{\makecell{\B 1\%\\\\\B Supervised:\\0.9312}} & MNDA & & \B 0.8860 & 0.6926 & 0.8162 \\
      & MIST1 & & \B 0.8651 & 0.6677 & 0.8162 \\
      & ERG & & \B 0.8899 & 0.4980 & 0.8327 \\
      & CD3CD20 & & \B 0.8651 & 0.7301 & 0.8511 \\
      & CD45RB & & \B 0.8839 & 0.8136 & 0.8247 \\
      & CD235a & & 0.8350 & 0.7663 & \B 0.8364 \\
      & $\alpha$SMA & & \B 0.9055 & 0.8963 & 0.8388 \\
      & pan-CK & & \B 0.9140 & 0.8767 & 0.8465 \\
    \midrule
    \multirow{8}{*}{\makecell{\B 0.2\%\\\\\B Supervised:\\0.7354}} & MNDA & & \B 0.8910 & 0.2841 & 0.8250 \\
      & MIST1 & & \B 0.8657 & 0.2872 & 0.7955 \\
      & ERG & & \B 0.8644 & 0.2300 & 0.8176 \\
      & CD3CD20 & & \B 0.8570 & 0.5277 & 0.8380 \\
      & CD45RB & & \B 0.8842 & 0.4538 & 0.7523 \\
      & CD235a & & 0.7691 & 0.5194 & \B 0.8026 \\
      & $\alpha$SMA & & \B 0.8871 & 0.7387 & 0.8349 \\
      & pan-CK & & \B 0.8848 & 0.7892 & 0.8242 \\
    \bottomrule
    \end{tabular}}
\end{table*}

\clearpage

\begin{table*}[!ht]
    \centering
    \caption{PanNuke results from Figure \ref{fig:dataset-sizes}. Models were the same as those evaluated in Figure \ref{tab:segpath}}
    \label{tab:pannuke-segpath-dataset-size-data}
    \scalebox{1}{\begin{tabular}{cllabc}
    \toprule
        & \multirow{3}{*}{\diagbox[height=3\line]{Stain}{}} & Loss: & \multicolumn{3}{c}{VICReg} \\
        & & Method: & TriDeNT \Neptune & \multicolumn{2}{c}{Siamese} \\
        Dataset \%ge & & Privileged: & \cmark & \cmark & \xmark \\
    \midrule
    \multirow{8}{*}{\makecell{\B 50\%\\\\\B Supervised:\\0.8901}} & MNDA & & \B 0.9264 & 0.8331 & 0.8971 \\
      & MIST1 & & \B 0.9141 & 0.7937 & 0.8958 \\
      & ERG & & \B 0.9198 & 0.7127 & 0.9001 \\
      & CD3CD20 & & \B 0.9330 & 0.8042 & 0.8979 \\
      & CD45RB & & \B 0.9255 & 0.8774 & 0.9019 \\
      & CD235a & & \B 0.8958 & 0.7854 & 0.8870 \\
      & $\alpha$SMA & & \B 0.9439 & 0.9089 & 0.8909 \\
      & pan-CK & & \B 0.9488 & 0.9207 & 0.8984 \\
    \midrule
    \multirow{8}{*}{\makecell{\B 20\%\\\\\B Supervised:\\0.8791}} & MNDA & & \B 0.9032 & 0.8235 & 0.8747 \\
      & MIST1 & & \B 0.9067 & 0.7806 & 0.8752 \\
      & ERG & & \B 0.9006 & 0.6759 & 0.8756 \\
      & CD3CD20 & & \B 0.9177 & 0.7893 & 0.8787 \\
      & CD45RB & & \B 0.9168 & 0.8633 & 0.8914 \\
      & CD235a & & \B 0.8940 & 0.7687 & 0.8677 \\
      & $\alpha$SMA & & \B 0.9203 & 0.8901 & 0.8887 \\
      & pan-CK & & \B 0.9295 & 0.9063 & 0.8857 \\
    \midrule
    \multirow{8}{*}{\makecell{\B 10\%\\\\\B Supervised:\\0.8581}} & MNDA & & \B 0.8879 & 0.8195 & 0.8515 \\
      & MIST1 & & \B 0.8787 & 0.7797 & 0.8528 \\
      & ERG & & \B 0.8695 & 0.6715 & 0.8585 \\
      & CD3CD20 & & \B 0.9001 & 0.7735 & 0.8590 \\
      & CD45RB & & \B 0.9054 & 0.8463 & 0.8633 \\
      & CD235a & & \B 0.8703 & 0.7854 & 0.8388 \\
      & $\alpha$SMA & & \B 0.8984 & 0.8739 & 0.8603 \\
      & pan-CK & & \B 0.9120 & 0.8927 & 0.8673 \\
    \midrule
    \multirow{8}{*}{\makecell{\B 5\%\\\\\B Supervised:\\0.6776}} & MNDA & & \B 0.8563 & 0.6557 & 0.8095 \\
      & MIST1 & & \B 0.8458 & 0.7516 & 0.8108 \\
      & ERG & & 0.8033 & 0.6570 & \B 0.8222 \\
      & CD3CD20 & & \B 0.8752 & 0.6329 & 0.8173 \\
      & CD45RB & & \B 0.8651 & 0.8117 & 0.8235 \\
      & CD235a & & \B 0.8300 & 0.5576 & 0.7989 \\
      & $\alpha$SMA & & \B 0.8651 & 0.8208 & 0.8305 \\
      & pan-CK & & \B 0.8852 & 0.7718 & 0.8379 \\
    \midrule
    \multirow{8}{*}{\makecell{\B 1\%\\\\\B Supervised:\\0.5751}} & MNDA & & \B 0.7201 & 0.6693 & 0.7043 \\
      & MIST1 & & 0.7319 & 0.6426 & \B 0.7468 \\
      & ERG & & \B 0.7376 & 0.5804 & 0.7240 \\
      & CD3CD20 & & \B 0.7933 & 0.4472 & 0.7284 \\
      & CD45RB & & \B 0.6951 & 0.5523 & 0.6807 \\
      & CD235a & & \B 0.6728 & 0.5545 & 0.6697 \\
      & $\alpha$SMA & & 0.7499 & \B 0.7661 & 0.7587 \\
      & pan-CK & & \B 0.7613 & 0.5615 & 0.7464 \\
    \midrule
    \multirow{8}{*}{\makecell{\B 0.2\%\\\\\B Supervised:\\0.5756}} & MNDA & & \B 0.7166 & 0.5453 & 0.6229 \\
      & MIST1 & & \B 0.6868 & 0.4494 & 0.5843 \\
      & ERG & & \B 0.7100 & 0.4564 & 0.6562 \\
      & CD3CD20 & & \B 0.6180 & 0.4468 & 0.6294 \\
      & CD45RB & & 0.7367 & \B 0.7376 & 0.5655 \\
      & CD235a & & \B 0.6316 & 0.5519 & 0.6246 \\
      & $\alpha$SMA & & \B 0.7035 & 0.6588 & 0.5712 \\
      & pan-CK & & \B 0.7398 & 0.5125 & 0.6194 \\
    \bottomrule
    \end{tabular}}
\end{table*}

\clearpage

\begin{table*}[!ht]
    \centering
    \caption{Singapore dataset results from Figure \ref{fig:dataset-sizes}. Models were the same as those evaluated in Figure \ref{tab:segpath}}
    \label{tab:singapore-segpath-dataset-size-data}
    \scalebox{1}{\begin{tabular}{cllabc}
    \toprule
        & \multirow{3}{*}{\diagbox[height=3\line]{Stain}{}} & Loss: & \multicolumn{3}{c}{VICReg} \\
        & & Method: & TriDeNT \Neptune & \multicolumn{2}{c}{Siamese} \\
        Dataset \%ge & & Privileged: & \cmark & \cmark & \xmark \\
    \midrule
    \multirow{8}{*}{\makecell{\B 50\%\\\\\B Supervised:\\0.9103}} & MNDA & & \B 0.8038 & 0.6564 & 0.7893 \\
      & MIST1 & & \B 0.8280 & 0.6632 & 0.7433 \\
      & ERG & & 0.8352 & 0.6046 & \B 0.8428 \\
      & CD3CD20 & & 0.7841 & 0.6639 & \B 0.8078 \\
      & CD45RB & & \B 0.8057 & 0.7066 & 0.7618 \\
      & CD235a & & \B 0.8252 & 0.6801 & 0.7806 \\
      & $\alpha$SMA & & \B 0.8536 & 0.8223 & 0.8315 \\
      & pan-CK & & \B 0.8470 & 0.8409 & 0.7623 \\
    \midrule
    \multirow{8}{*}{\makecell{\B 20\%\\\\\B Supervised:\\0.8559}} & MNDA & & \B 0.8355 & 0.6337 & 0.8186 \\
      & MIST1 & & \B 0.8500 & 0.6341 & 0.7552 \\
      & ERG & & 0.8130 & 0.5970 & \B 0.8479 \\
      & CD3CD20 & & 0.7815 & 0.6947 & \B 0.8240 \\
      & CD45RB & & \B 0.8254 & 0.6508 & 0.7900 \\
      & CD235a & & 0.7714 & 0.6719 & \B 0.7855 \\
      & $\alpha$SMA & & \B 0.8721 & 0.7946 & 0.8317 \\
      & pan-CK & & \B 0.8543 & 0.8526 & 0.8045 \\
    \midrule
    \multirow{8}{*}{\makecell{\B 10\%\\\\\B Supervised:\\0.6759}} & MNDA & & \B 0.8367 & 0.5722 & 0.8057 \\
      & MIST1 & & \B 0.8571 & 0.5689 & 0.7501 \\
      & ERG & & 0.8139 & 0.5686 & \B 0.8367 \\
      & CD3CD20 & & \B 0.8437 & 0.7015 & 0.8188 \\
      & CD45RB & & \B 0.8561 & 0.7142 & 0.7979 \\
      & CD235a & & \B 0.8127 & 0.7252 & 0.7423 \\
      & $\alpha$SMA & & \B 0.8702 & 0.7479 & 0.8169 \\
      & pan-CK & & \B 0.8402 & 0.8223 & 0.8284 \\
    \midrule
    \multirow{8}{*}{\makecell{\B 5\%\\\\\B Supervised:\\0.6444}} & MNDA & & \B 0.8214 & 0.3633 & 0.7585 \\
      & MIST1 & & \B 0.8270 & 0.4921 & 0.7634 \\
      & ERG & & \B 0.7428 & 0.6222 & 0.7315 \\
      & CD3CD20 & & \B 0.8202 & 0.4234 & 0.7367 \\
      & CD45RB & & \B 0.8317 & 0.6630 & 0.7810 \\
      & CD235a & & \B 0.7878 & 0.6470 & 0.6973 \\
      & $\alpha$SMA & & \B 0.8214 & 0.7350 & 0.7580 \\
      & pan-CK & & \B 0.8287 & 0.7916 & 0.7747 \\
    \midrule
    \multirow{8}{*}{\makecell{\B 1\%\\\\\B Supervised:\\0.5295}} & MNDA & & \B 0.7597 & 0.3621 & 0.5916 \\
      & MIST1 & & \B 0.7179 & 0.6379 & 0.6956 \\
      & ERG & & 0.6696 & 0.6379 & \B 0.6834 \\
      & CD3CD20 & & \B 0.6998 & 0.6482 & 0.5914 \\
      & CD45RB & & \B 0.7259 & 0.6379 & 0.6818 \\
      & CD235a & & \B 0.6879 & 0.3621 & 0.5637 \\
      & $\alpha$SMA & & \B 0.7599 & 0.6513 & 0.6726 \\
      & pan-CK & & \B 0.7393 & 0.7327 & 0.6383 \\
    \midrule
    \multirow{8}{*}{\makecell{\B 0.2\%\\\\\B Supervised:\\0.3621}} & MNDA & & \B 0.7700 & 0.3774 & 0.6691 \\
      & MIST1 & & \B 0.6743 & 0.6379 & 0.6567 \\
      & ERG & & 0.6236 & 0.6114 & \B 0.7355 \\
      & CD3CD20 & & \B 0.7174 & 0.3621 & 0.7045 \\
      & CD45RB & & 0.6597 & \B 0.6991 & 0.6782 \\
      & CD235a & & \B 0.7174 & 0.5525 & 0.6923 \\
      & $\alpha$SMA & & \B 0.7320 & 0.6480 & 0.7238 \\
      & pan-CK & & \B 0.7273 & 0.6618 & 0.6541 \\
    \bottomrule
    \end{tabular}}
\end{table*}

\clearpage

\begin{table*}[!ht]
    \centering
    \caption{MHIST dataset results from Figure \ref{fig:dataset-sizes}. Models were the same as those evaluated in Figure \ref{tab:segpath}}
    \label{tab:mhist-segpath-dataset-size-data}
    \scalebox{1}{\begin{tabular}{cllabc}
    \toprule
        & \multirow{3}{*}{\diagbox[height=3\line]{Stain}{}} & Loss: & \multicolumn{3}{c}{VICReg} \\
        & & Method: & TriDeNT \Neptune & \multicolumn{2}{c}{Siamese} \\
        Dataset \%ge & & Privileged: & \cmark & \cmark & \xmark \\
    \midrule
    \multirow{8}{*}{\makecell{\B 50\%\\\\\B Supervised:\\0.7042}} & MNDA & & \B 0.7267 & 0.6714 & 0.7103 \\
      & MIST1 & & 0.6950 & 0.5742 & \B 0.7114 \\
      & ERG & & \B 0.7175 & 0.6080 & 0.7052 \\
      & CD3CD20 & & \B 0.7431 & 0.6581 & 0.7134 \\
      & CD45RB & & \B 0.7513 & 0.6817 & 0.7052 \\
      & CD235a & & \B 0.7144 & 0.6499 & 0.7093 \\
      & $\alpha$SMA & & \B 0.7584 & 0.7308 & 0.6755 \\
      & pan-CK & & \B 0.8076 & 0.7738 & 0.7144 \\
    \midrule
    \multirow{8}{*}{\makecell{\B 20\%\\\\\B Supervised:\\0.6049}} & MNDA & & \B 0.7185 & 0.6203 & 0.6950 \\
      & MIST1 & & \B 0.6991 & 0.6049 & 0.6878 \\
      & ERG & & \B 0.7021 & 0.5210 & 0.6950 \\
      & CD3CD20 & & 0.6919 & 0.6520 & \B 0.7001 \\
      & CD45RB & & \B 0.7329 & 0.6776 & 0.7001 \\
      & CD235a & & \B 0.6960 & 0.6725 & 0.6602 \\
      & $\alpha$SMA & & \B 0.7574 & 0.7083 & 0.6766 \\
      & pan-CK & & \B 0.7615 & 0.7523 & 0.6786 \\
    \midrule
    \multirow{8}{*}{\makecell{\B 10\%\\\\\B Supervised:\\0.6008}} & MNDA & & \B 0.7257 & 0.6418 & 0.6981 \\
      & MIST1 & & 0.6223 & 0.4964 & \B 0.6684 \\
      & ERG & & 0.6714 & 0.6285 & \B 0.6888 \\
      & CD3CD20 & & \B 0.6684 & 0.6059 & 0.6612 \\
      & CD45RB & & \B 0.6827 & 0.6315 & 0.6561 \\
      & CD235a & & 0.6469 & 0.6346 & \B 0.7052 \\
      & $\alpha$SMA & & \B 0.7574 & 0.6346 & 0.6745 \\
      & pan-CK & & \B 0.7421 & 0.7195 & 0.6469 \\
    \midrule
    \multirow{8}{*}{\makecell{\B 5\%\\\\\B Supervised:\\0.4637}} & MNDA & & 0.5527 & 0.6315 & \B 0.6653 \\
      & MIST1 & & \B 0.6725 & 0.3685 & \B 0.6725 \\
      & ERG & & 0.6029 & 0.3685 & \B 0.6745 \\
      & CD3CD20 & & 0.6407 & 0.6325 & \B 0.6469 \\
      & CD45RB & & \B 0.6479 & 0.6315 & 0.6264 \\
      & CD235a & & 0.6182 & 0.6315 & \B 0.6540 \\
      & $\alpha$SMA & & \B 0.6837 & 0.6315 & 0.6673 \\
      & pan-CK & & \B 0.7277 & 0.4432 & 0.6807 \\
    \midrule
    \multirow{8}{*}{\makecell{\B 1\%\\\\\B Supervised:\\0.6315}} & MNDA & & 0.6346 & 0.6315 & \B 0.6489 \\
      & MIST1 & & 0.6213 & 0.3685 & \B 0.6366 \\
      & ERG & & 0.6192 & \B 0.6315 & 0.5230 \\
      & CD3CD20 & & \B 0.6448 & 0.3900 & 0.6203 \\
      & CD45RB & & 0.6336 & 0.3685 & \B 0.6540 \\
      & CD235a & & \B 0.6315 & 0.3685 & 0.5885 \\
      & $\alpha$SMA & & 0.5210 & \B 0.6540 & 0.6530 \\
      & pan-CK & & \B 0.6663 & 0.4862 & 0.5803 \\
    \midrule
    \multirow{8}{*}{\makecell{\B 0.2\%\\\\\B Supervised:\\0.3685}} & MNDA & & 0.3808 & \B 0.5343 & 0.5097 \\
      & MIST1 & & \B 0.5599 & 0.3685 & 0.5241 \\
      & ERG & & 0.3961 & \B 0.6315 & 0.4739 \\
      & CD3CD20 & & 0.5834 & \B 0.6264 & 0.5937 \\
      & CD45RB & & 0.4371 & \B 0.5220 & 0.4340 \\
      & CD235a & & 0.4145 & \B 0.6315 & 0.4125 \\
      & $\alpha$SMA & & \B 0.5814 & 0.3838 & 0.4903 \\
      & pan-CK & & \B 0.5486 & 0.4954 & 0.4800 \\
    \bottomrule
    \end{tabular}}
\end{table*}

\clearpage

\begin{table*}[h]
    \centering
    \caption{PanNuke three-fold cross-validation.}
    \label{tab:pannuke_cross_validation}
    \begin{tabular}{ccccc}
    \toprule
    Stain & Fold & TriDeNT \Neptune & \makecell{Siamese\\(Privileged)} & \makecell{Siamese\\(Unprivileged)} \\
    \midrule
         \multirow{3}{*}{MNDA} & 0 & \greycell\B 0.9288 & 0.8327 & 0.9156 \\
         & 1 & \greycell\B 0.9263 & 0.8349 & 0.9081 \\
         & 2 & \greycell\B 0.9360 & 0.8528 & 0.9102 \\
    \midrule
        \multirow{3}{*}{MIST1} & 0 & \greycell\B 0.9206 & 0.8231 & 0.9156 \\
         & 1 & \greycell\B 0.9134 & 0.8215 & 0.9105 \\
         & 2 & \greycell\B 0.9282 & 0.8204 & 0.9159 \\
    \midrule
        \multirow{3}{*}{ERG} & 0 & \greycell\B 0.9315 & 0.6821 & 0.9302 \\
         & 1 & \greycell\B 0.9268 & 0.7057 & 0.9144 \\
         & 2 & \greycell\B 0.9299 & 0.7197 & 0.9194 \\
    \midrule
        \multirow{3}{*}{CD3CD20} & 0 & \greycell\B 0.9261 & 0.8168 & 0.9161 \\
         & 1 & \greycell\B 0.9292 & 0.8268 & 0.9096 \\
         & 2 & \greycell\B 0.9356 & 0.8292 & 0.9137 \\
    \midrule
        \multirow{3}{*}{CD45RB} & 0 & \greycell\B 0.9406 & 0.8680 & 0.9206 \\
         & 1 & \greycell\B 0.9278 & 0.8718 & 0.9110 \\
         & 2 & \greycell\B 0.9330 & 0.8984 & 0.9216 \\
    \midrule
        \multirow{3}{*}{CD235a} & 0 & \greycell\B 0.9197 & 0.8050 & 0.9166 \\
         & 1 & \greycell\B 0.9191 & 0.7990 & 0.9000 \\
         & 2 & \greycell\B 0.9141 & 0.8178 & 0.9067 \\
    \midrule
        \multirow{3}{*}{$\alpha$SMA} & 0 & \greycell\B 0.9447 & 0.9088 & 0.9147 \\
         & 1 & \greycell\B 0.9368 & 0.9100 & 0.9077 \\
         & 2 & \greycell\B 0.9466 & 0.9255 & 0.9137 \\
    \midrule
        \multirow{3}{*}{pan-CK} & 0 & \greycell\B 0.9469 & 0.9293 & 0.9229 \\
         & 1 & \greycell\B 0.9522 & 0.9263 & 0.9129 \\
         & 2 & \greycell\B 0.9509 & 0.9334 & 0.9190 \\
    \bottomrule
    \end{tabular}
\end{table*}

\clearpage

\begin{table*}[!ht]
    \centering
    \caption{Five-fold cross validation results for IMP-Cervix task.}
    \label{tab:cervix-cross-validation}
    \scalebox{0.9}{\begin{tabular}{cllabc|abc}
    \toprule
        & \multirow{3}{*}{\diagbox[height=3\line]{Stain}{}} & Loss: & \multicolumn{3}{c}{VICReg} & \multicolumn{3}{c}{InfoNCE} \\
        & & Method: & TriDeNT \Neptune & \multicolumn{2}{c}{Siamese} & TriDeNT \Neptune & \multicolumn{2}{c}{Siamese} \\
        Fold & & Privileged: & \cmark & \cmark & \xmark & \cmark & \cmark & \xmark \\
    \midrule
    \multirow{8}{*}{1} & MNDA & & \B 0.7570 & 0.7477 & 0.7477 & \B 0.7944 & 0.7103 & 0.7290 \\
    & MIST1 & & \B 0.7664 & 0.7477 & \B 0.7664 & \B 0.7477 & 0.6729 & 0.7383 \\
    & ERG & & \B 0.8037 & 0.7477 & 0.7664 & \B 0.7570 & 0.6822 & 0.7477 \\
    & CD3CD20 & & 0.7757 & 0.7477 & \B 0.7850 & \B 0.7570 & 0.7477 & 0.7196 \\
    & CD45RB & & \B 0.7757 & 0.7664 & 0.7290 & \B 0.7757 & 0.7290 & 0.6822 \\
    & CD235a & & 0.7009 & \B 0.7850 & 0.7009 & 0.7477 & 0.7383 & \B 0.7944 \\
    & $\alpha$SMA & & 0.7477 & 0.7103 & \B 0.7757 & \B 0.7383 & 0.7196 & 0.7196 \\
    & pan-CK & & 0.7009 & \B 0.7477 & 0.7103 & \B 0.7850 & 0.7196 & 0.7757 \\
    \midrule
    & Average & & \B 0.7535 & 0.7500 & 0.7477 & \B 0.7629 & 0.7150 & 0.7383 \\
    \midrule
    \multirow{8}{*}{2} & MNDA & & \B 0.7664 & 0.7570 & 0.7103 & \B 0.8131 & 0.8037 & 0.7570 \\
    & MIST1 & & \B 0.7477 & 0.7290 & 0.7290 & \B 0.7570 & 0.7290 & \B 0.7570 \\
    & ERG & & 0.7383 & 0.6822 & \B 0.7757 & \B 0.7944 & 0.7196 & 0.7850 \\
    & CD3CD20 & & 0.7664 & \B 0.7757 & 0.7290 & \B 0.7664 & 0.7570 & 0.7477 \\
    & CD45RB & & \B 0.7850 & \B 0.7850 & 0.7757 & \B 0.7850 & \B 0.7850 & 0.7664 \\
    & CD235a & & \B 0.7850 & 0.7477 & 0.7103 & 0.7009 & \B 0.7757 & 0.7570 \\
    & $\alpha$SMA & & 0.7477 & 0.7196 & \B 0.8037 & 0.7850 & \B 0.8131 & 0.7944 \\
    & pan-CK & & 0.7850 & \B 0.8131 & 0.7477 & \B 0.8224 & 0.8131 & 0.7757 \\
    \midrule
    & Average & & \B 0.7652 & 0.7512 & 0.7477 & \B 0.7780 & 0.7745 & 0.7675 \\
    \midrule
    \multirow{8}{*}{3} & MNDA & & \B 0.7570 & 0.7103 & 0.7383 & \B 0.7477 & 0.6916 & 0.7196 \\
    & MIST1 & & 0.7477 & \B 0.7570 & 0.7103 & \B 0.7383 & 0.6916 & 0.6729 \\
    & ERG & & 0.7383 & 0.6916 & \B 0.7570 & \B 0.7383 & 0.6075 & \B 0.7383 \\
    & CD3CD20 & & \B 0.7009 & 0.6916 & 0.6916 & \B 0.7664 & 0.6916 & \B 0.7664 \\
    & CD45RB & & 0.7103 & 0.7477 & \B 0.7570 & \B 0.7383 & 0.6916 & 0.7103 \\
    & CD235a & & \B 0.7570 & 0.7290 & 0.7103 & \B 0.7009 & \B 0.7009 & 0.6822 \\
    & $\alpha$SMA & & \B 0.7757 & 0.7383 & 0.7290 & \B 0.7944 & 0.7009 & 0.6822 \\
    & pan-CK & & \B 0.7570 & 0.7477 & 0.7103 & \B 0.7664 & 0.7570 & 0.7290 \\
    \midrule
    & Average & & \B 0.7430 & 0.7266 & 0.7255 & \B 0.7488 & 0.6916 & 0.7126 \\
    \midrule
    \multirow{8}{*}{4} & MNDA & & 0.6729 & \B 0.7290 & 0.7103 & \B 0.8037 & 0.7290 & 0.6822 \\
    & MIST1 & & \B 0.7570 & 0.7196 & 0.7196 & \B 0.7477 & 0.6822 & 0.7196 \\
    & ERG & & \B 0.7383 & 0.6916 & 0.7103 & \B 0.7664 & 0.6636 & 0.6822 \\
    & CD3CD20 & & 0.6636 & 0.7196 & \B 0.7383 & \B 0.7383 & 0.6355 & 0.6822 \\
    & CD45RB & & 0.7196 & \B 0.7290 & 0.6916 & \B 0.7477 & 0.6729 & 0.7290 \\
    & CD235a & & 0.6262 & 0.7009 & \B 0.7290 & 0.6822 & 0.7290 & \B 0.7570 \\
    & $\alpha$SMA & & \B 0.8037 & 0.6916 & 0.7290 & \B 0.8037 & 0.7664 & 0.7477 \\
    & pan-CK & & \B 0.7477 & 0.7383 & 0.6916 & 0.7757 & \B 0.7850 & 0.7383 \\
    \midrule
    & Average & & \B 0.7161 & 0.7150 & 0.7150 & \B 0.7582 & 0.7079 & 0.7173 \\
    \midrule
    \multirow{8}{*}{5} & MNDA & & \B 0.7547 & 0.6698 & 0.7264 & \B 0.7925 & 0.7830 & 0.7453 \\
    & MIST1 & & \B 0.7453 & 0.6509 & 0.7170 & 0.7736 & \B 0.7925 & 0.7736 \\
    & ERG & & \B 0.7642 & 0.6887 & 0.7358 & 0.7264 & 0.6887 & \B 0.7358 \\
    & CD3CD20 & & \B 0.7736 & 0.6981 & 0.7358 & \B 0.7925 & 0.7642 & 0.7358 \\
    & CD45RB & & 0.6981 & \B 0.7358 & \B 0.7358 & \B 0.7170 & 0.6509 & \B 0.7170 \\
    & CD235a & & \B 0.7170 & \B 0.7170 & 0.7075 & \B 0.7547 & 0.7075 & 0.7358 \\
    & $\alpha$SMA & & 0.7170 & \B 0.7264 & \B 0.7264 & \B 0.7736 & 0.7170 & 0.7453 \\
    & pan-CK & & \B 0.7642 & 0.7264 & 0.6321 & \B 0.7736 & \B 0.7736 & 0.7075 \\
    \midrule
    & Average & & \B 0.7417 & 0.7017 & 0.7146 & \B 0.7630 & 0.7347 & 0.7370 \\
    \midrule
    \multirow{8}{*}{Averages}
    & MNDA & & \B 0.7416 & 0.7227 & 0.7266 & \B 0.7903 & 0.7435 & 0.7266 \\
    & MIST1 & & \B 0.7528 & 0.7208 & 0.7284 & \B 0.7528 & 0.7136 & 0.7323 \\
    & ERG & & \B 0.7566 & 0.7004 & 0.7490 & \B 0.7565 & 0.6723 & 0.7378 \\
    & CD3CD20 & & \B 0.7360 & 0.7265 & \B 0.7360 & \B 0.7641 & 0.7192 & 0.7303 \\
    & CD45RB & & 0.7378 & \B 0.7528 & 0.7378 & \B 0.7527 & 0.7059 & 0.7210 \\
    & CD235a & & 0.7172 & \B 0.7359 & 0.7116 & 0.7173 & 0.7303 & \B 0.7453 \\
    & $\alpha$SMA & & \B 0.7583 & 0.7172 & 0.7528 & \B 0.7790 & 0.7434 & 0.7378 \\
    & pan-CK & & 0.7510 & \B 0.7546 & 0.6984 & \B 0.7846 & 0.7697 & 0.7452 \\
    \midrule
    & Overall Average & & \B 0.7439 & 0.7289 & 0.7301 & \B 0.7622 & 0.7247 & 0.7346 \\
    \bottomrule
    \end{tabular}}
\end{table*}

\clearpage

\begin{table*}[!ht]
\centering
    \caption{Results for adversarial attack for values $\epsilon=0.001,...,0.05$.}
    \label{tab:full-adversarial-0.001-0.05}
    \scalebox{0.9}{\begin{tabular}{cllabc|abc}
    \toprule
        & \multirow{3}{*}{\diagbox[height=3\line]{Stain}{}} & Loss: & \multicolumn{3}{c}{VICReg} & \multicolumn{3}{c}{InfoNCE} \\
        & & Method: & TriDeNT \Neptune & \multicolumn{2}{c}{Siamese} & TriDeNT \Neptune & \multicolumn{2}{c}{Siamese} \\
        Fold & & Privileged: & \cmark & \cmark & \xmark & \cmark & \cmark & \xmark \\
    \midrule
    \multirow{8}{*}{0.001} & MNDA & & 0.0007 & 0.0088 & 0.0010 & 0.0004 & 0.0018 & 0.0014 \\
     & MIST1 & & 0.0013 & 0.0141 & 0.0013 & 0.0008 & 0.0024 & 0.0028 \\
     & ERG & & 0.0006 & 0.0061 & 0.0007 & 0.0011 & 0.0014 & 0.0022 \\
     & CD3CD20 & & 0.0017 & 0.0068 & 0.0006 & 0.0011 & 0.0015 & 0.0018 \\
     & CD45RB & & 0.0017 & 0.0065 & 0.0018 & 0.0018 & 0.0013 & 0.0040 \\
     & CD235a & & 0.0014 & 0.0046 & 0.0007 & 0.0025 & 0.0021 & 0.0020 \\
     & pan-CK & & 0.0007 & 0.0008 & 0.0010 & 0.0020 & 0.0006 & 0.0020 \\
     & $\alpha$SMA & & 0.0010 & 0.0011 & 0.0015 & 0.0010 & 0.0013 & 0.0020 \\
    \midrule
    \multirow[c]{8}{*}{0.002} & MNDA & & 0.0007 & 0.0086 & 0.0010 & 0.0004 & 0.0018 & 0.0014 \\
     & MIST1 & & 0.0013 & 0.0138 & 0.0013 & 0.0008 & 0.0024 & 0.0028 \\
     & ERG & & 0.0006 & 0.0061 & 0.0008 & 0.0011 & 0.0014 & 0.0022 \\
     & CD3CD20 & & 0.0017 & 0.0067 & 0.0006 & 0.0010 & 0.0017 & 0.0018 \\
     & CD45RB & & 0.0020 & 0.0064 & 0.0017 & 0.0018 & 0.0013 & 0.0040 \\
     & CD235a & & 0.0013 & 0.0046 & 0.0007 & 0.0024 & 0.0021 & 0.0020 \\
     & pan-CK & & 0.0007 & 0.0008 & 0.0010 & 0.0020 & 0.0004 & 0.0020 \\
     & $\alpha$SMA & & 0.0010 & 0.0011 & 0.0015 & 0.0011 & 0.0013 & 0.0020 \\
    \midrule
    \multirow[c]{8}{*}{0.005} & MNDA & & 0.0018 & 0.0144 & 0.0025 & 0.0018 & 0.0043 & 0.0025 \\
     & MIST1 & & 0.0020 & 0.0247 & 0.0018 & 0.0028 & 0.0057 & 0.0042 \\
     & ERG & & 0.0014 & 0.0132 & 0.0024 & 0.0014 & 0.0043 & 0.0035 \\
     & CD3CD20 & & 0.0031 & 0.0163 & 0.0021 & 0.0024 & 0.0045 & 0.0029 \\
     & CD45RB & & 0.0040 & 0.0120 & 0.0039 & 0.0035 & 0.0031 & 0.0061 \\
     & CD235a & & 0.0031 & 0.0086 & 0.0021 & 0.0045 & 0.0046 & 0.0032 \\
     & pan-CK & & 0.0021 & 0.0024 & 0.0026 & 0.0032 & 0.0020 & 0.0033 \\
     & $\alpha$SMA & & 0.0025 & 0.0028 & 0.0029 & 0.0020 & 0.0022 & 0.0033 \\
    \midrule
    \multirow[c]{8}{*}{0.01} & MNDA & & 0.0029 & 0.0238 & 0.0045 & 0.0028 & 0.0063 & 0.0029 \\
     & MIST1 & & 0.0035 & 0.0337 & 0.0029 & 0.0040 & 0.0093 & 0.0063 \\
     & ERG & & 0.0024 & 0.0194 & 0.0035 & 0.0029 & 0.0070 & 0.0049 \\
     & CD3CD20 & & 0.0045 & 0.0220 & 0.0031 & 0.0038 & 0.0070 & 0.0040 \\
     & CD45RB & & 0.0050 & 0.0164 & 0.0052 & 0.0059 & 0.0047 & 0.0074 \\
     & CD235a & & 0.0054 & 0.0134 & 0.0029 & 0.0063 & 0.0077 & 0.0052 \\
     & pan-CK & & 0.0032 & 0.0042 & 0.0039 & 0.0050 & 0.0032 & 0.0050 \\
     & $\alpha$SMA & & 0.0033 & 0.0050 & 0.0046 & 0.0031 & 0.0039 & 0.0045 \\
    \midrule
    \multirow[c]{8}{*}{0.02} & MNDA & & 0.0071 & 0.0419 & 0.0086 & 0.0049 & 0.0132 & 0.0071 \\
     & MIST1 & & 0.0075 & 0.0648 & 0.0057 & 0.0102 & 0.0210 & 0.0118 \\
     & ERG & & 0.0057 & 0.0361 & 0.0081 & 0.0072 & 0.0125 & 0.0089 \\
     & CD3CD20 & & 0.0091 & 0.0439 & 0.0079 & 0.0084 & 0.0155 & 0.0093 \\
     & CD45RB & & 0.0110 & 0.0283 & 0.0107 & 0.0116 & 0.0124 & 0.0118 \\
     & CD235a & & 0.0127 & 0.0298 & 0.0061 & 0.0121 & 0.0174 & 0.0093 \\
     & pan-CK & & 0.0063 & 0.0110 & 0.0082 & 0.0086 & 0.0072 & 0.0082 \\
     & $\alpha$SMA & & 0.0061 & 0.0106 & 0.0082 & 0.0053 & 0.0096 & 0.0082 \\
    \midrule
    \multirow[c]{8}{*}{0.05} & MNDA & & 0.0145 & 0.0922 & 0.0167 & 0.0125 & 0.0305 & 0.0173 \\
     & MIST1 & & 0.0178 & 0.1407 & 0.0150 & 0.0213 & 0.0482 & 0.0249 \\
     & ERG & & 0.0170 & 0.0801 & 0.0162 & 0.0166 & 0.0318 & 0.0206 \\
     & CD3CD20 & & 0.0184 & 0.0996 & 0.0153 & 0.0199 & 0.0404 & 0.0223 \\
     & CD45RB & & 0.0238 & 0.0691 & 0.0219 & 0.0252 & 0.0304 & 0.0247 \\
     & CD235a & & 0.0270 & 0.0706 & 0.0141 & 0.0265 & 0.0450 & 0.0187 \\
     & pan-CK & & 0.0155 & 0.0226 & 0.0181 & 0.0192 & 0.0176 & 0.0191 \\
     & $\alpha$SMA & & 0.0132 & 0.0244 & 0.0176 & 0.0146 & 0.0209 & 0.0201 \\
    \bottomrule
    \end{tabular}
    }
\end{table*}

\clearpage

\begin{table*}[!ht]
\centering
    \caption{Results for adversarial attack for values $\epsilon=0.1,...,10$.}
    \label{tab:full-adversarial-0.1-10}
    \scalebox{0.9}{\begin{tabular}{cllabc|abc}
    \toprule
        & \multirow{3}{*}{\diagbox[height=3\line]{Stain}{}} & Loss: & \multicolumn{3}{c}{VICReg} & \multicolumn{3}{c}{InfoNCE} \\
        & & Method: & TriDeNT \Neptune & \multicolumn{2}{c}{Siamese} & TriDeNT \Neptune & \multicolumn{2}{c}{Siamese} \\
        Fold & & Privileged: & \cmark & \cmark & \xmark & \cmark & \cmark & \xmark \\
    \midrule
    \multirow[c]{8}{*}{0.1} & MNDA & & 0.0315 & 0.2195 & 0.0375 & 0.0304 & 0.0889 & 0.0379 \\
     & MIST1 & & 0.0483 & 0.2982 & 0.0382 & 0.0486 & 0.1049 & 0.0510 \\
     & ERG & & 0.0376 & 0.1679 & 0.0382 & 0.0359 & 0.0701 & 0.0496 \\
     & CD3CD20 & & 0.0397 & 0.2154 & 0.0390 & 0.0474 & 0.1130 & 0.0581 \\
     & CD45RB & & 0.0521 & 0.1640 & 0.0472 & 0.0637 & 0.0779 & 0.0627 \\
     & CD235a & & 0.0580 & 0.1491 & 0.0291 & 0.0581 & 0.1130 & 0.0463 \\
     & pan-CK & & 0.0380 & 0.0563 & 0.0379 & 0.0450 & 0.0449 & 0.0451 \\
     & $\alpha$SMA & & 0.0410 & 0.0567 & 0.0425 & 0.0330 & 0.0521 & 0.0465 \\
    \midrule
    \multirow[c]{8}{*}{0.2} & MNDA & & 0.0764 & 0.4007 & 0.0871 & 0.0810 & 0.2136 & 0.0911 \\
     & MIST1 & & 0.1212 & 0.4643 & 0.0882 & 0.1115 & 0.2285 & 0.1127 \\
     & ERG & & 0.0935 & 0.3014 & 0.0903 & 0.0915 & 0.1586 & 0.1189 \\
     & CD3CD20 & & 0.0931 & 0.3790 & 0.0876 & 0.1101 & 0.2522 & 0.1395 \\
     & CD45RB & & 0.1233 & 0.3266 & 0.1005 & 0.1572 & 0.2114 & 0.1480 \\
     & CD235a & & 0.1445 & 0.3334 & 0.0733 & 0.1364 & 0.2537 & 0.1156 \\
     & pan-CK & & 0.0847 & 0.1535 & 0.0901 & 0.1067 & 0.1297 & 0.1014 \\
     & $\alpha$SMA & & 0.1027 & 0.1354 & 0.0986 & 0.0904 & 0.1325 & 0.1010 \\
    \midrule
    \multirow[c]{8}{*}{0.5} & MNDA & & 0.3090 & 0.6263 & 0.2862 & 0.3340 & 0.5132 & 0.2998 \\
     & MIST1 & & 0.3897 & 0.6192 & 0.2762 & 0.3539 & 0.5091 & 0.3319 \\
     & ERG & & 0.3164 & 0.4992 & 0.2593 & 0.3315 & 0.3953 & 0.3649 \\
     & CD3CD20 & & 0.3142 & 0.5562 & 0.2526 & 0.3628 & 0.5536 & 0.3965 \\
     & CD45RB & & 0.4198 & 0.6303 & 0.2756 & 0.4620 & 0.5877 & 0.4431 \\
     & CD235a & & 0.4091 & 0.6182 & 0.2504 & 0.4467 & 0.5284 & 0.3901 \\
     & pan-CK & & 0.3083 & 0.4236 & 0.2714 & 0.3775 & 0.4485 & 0.3350 \\
     & $\alpha$SMA & & 0.3384 & 0.3540 & 0.2716 & 0.3352 & 0.4042 & 0.3150 \\
    \midrule
    \multirow[c]{8}{*}{1} & MNDA & & 0.6046 & 0.7163 & 0.5038 & 0.6359 & 0.7354 & 0.5615 \\
     & MIST1 & & 0.6630 & 0.7059 & 0.5338 & 0.6271 & 0.6741 & 0.6440 \\
     & ERG & & 0.5803 & 0.6075 & 0.4838 & 0.6060 & 0.5948 & 0.6377 \\
     & CD3CD20 & & 0.5682 & 0.6613 & 0.4634 & 0.6230 & 0.7422 & 0.6898 \\
     & CD45RB & & 0.6943 & 0.7988 & 0.5091 & 0.7302 & 0.7929 & 0.7004 \\
     & CD235a & & 0.6868 & 0.7511 & 0.5058 & 0.7330 & 0.6886 & 0.6643 \\
     & pan-CK & & 0.6492 & 0.7227 & 0.5441 & 0.7045 & 0.7636 & 0.6338 \\
     & $\alpha$SMA & & 0.6235 & 0.6023 & 0.5282 & 0.6466 & 0.7307 & 0.6216 \\
    \midrule
    \multirow[c]{8}{*}{2} & MNDA & & 0.9404 & 0.8184 & 0.8276 & 0.9356 & 0.9128 & 0.9053 \\
     & MIST1 & & 0.9184 & 0.8269 & 0.8633 & 0.9143 & 0.8938 & 0.8693 \\
     & ERG & & 0.8764 & 0.7406 & 0.8347 & 0.8984 & 0.8576 & 0.8842 \\
     & CD3CD20 & & 0.9142 & 0.8648 & 0.8735 & 0.9338 & 0.9117 & 0.8839 \\
     & CD45RB & & 0.9267 & 0.9125 & 0.8329 & 0.9228 & 0.9338 & 0.8427 \\
     & CD235a & & 0.9050 & 0.8633 & 0.8541 & 0.9171 & 0.8648 & 0.8490 \\
     & pan-CK & & 0.9409 & 0.9242 & 0.8818 & 0.9544 & 0.9404 & 0.8798 \\
     & $\alpha$SMA & & 0.9585 & 0.9301 & 0.8398 & 0.9582 & 0.9373 & 0.8676 \\
    \midrule
    \multirow[c]{8}{*}{5} & MNDA & & 0.9477 & 0.8531 & 0.9216 & 0.9387 & 0.9229 & 0.9353 \\
     & MIST1 & & 0.9294 & 0.8384 & 0.9107 & 0.9236 & 0.8997 & 0.9218 \\
     & ERG & & 0.9448 & 0.8929 & 0.9144 & 0.9519 & 0.9085 & 0.9306 \\
     & CD3CD20 & & 0.9404 & 0.8828 & 0.9140 & 0.9434 & 0.9170 & 0.9285 \\
     & CD45RB & & 0.9324 & 0.9238 & 0.9210 & 0.9260 & 0.9359 & 0.9256 \\
     & CD235a & & 0.9282 & 0.8651 & 0.9326 & 0.9448 & 0.8825 & 0.9349 \\
     & pan-CK & & 0.9457 & 0.9262 & 0.9096 & 0.9553 & 0.9416 & 0.9334 \\
     & $\alpha$SMA & & 0.9628 & 0.9491 & 0.9146 & 0.9634 & 0.9455 & 0.9341 \\
    \midrule
    \multirow[c]{8}{*}{10} & MNDA & & 0.9498 & 0.8575 & 0.9249 & 0.9397 & 0.9248 & 0.9394 \\
     & MIST1 & & 0.9321 & 0.8438 & 0.9165 & 0.9271 & 0.9046 & 0.9298 \\
     & ERG & & 0.9477 & 0.9072 & 0.9182 & 0.9553 & 0.9146 & 0.9340 \\
     & CD3CD20 & & 0.9415 & 0.8902 & 0.9174 & 0.9444 & 0.9190 & 0.9335 \\
     & CD45RB & & 0.9340 & 0.9305 & 0.9260 & 0.9298 & 0.9415 & 0.9306 \\
     & CD235a & & 0.9298 & 0.8667 & 0.9367 & 0.9462 & 0.8887 & 0.9406 \\
     & pan-CK & & 0.9475 & 0.9268 & 0.9131 & 0.9589 & 0.9429 & 0.9363 \\
     & $\alpha$SMA & & 0.9639 & 0.9498 & 0.9193 & 0.9647 & 0.9489 & 0.9409 \\
    \bottomrule
    \end{tabular}
    }
\end{table*}

\clearpage

\begin{figure*}
    \centering
    \includegraphics[width=0.9\textwidth]{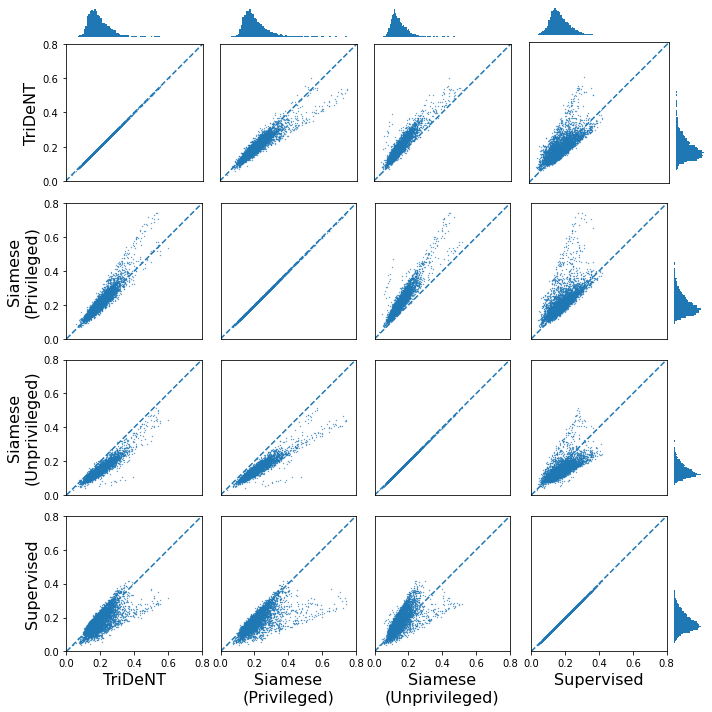}
    \caption{Mouse ALS-ST dataset - Greatest absolute correlation between gene counts and elements of each model's representations. We compare the correlation strength for the expression of each gene, with histograms appended to illustrate the distribution for each model. Dashed line is the identity. Above the line implies a greater correlation for the model listed on the y-axis, while below the line implies a greater correlation for the model listed on the x-axis.}
    \label{fig:counts-correlation-plot-mouse}
\end{figure*}

\clearpage

\begin{figure*}
    \centering
    \includegraphics[width=0.9\textwidth]{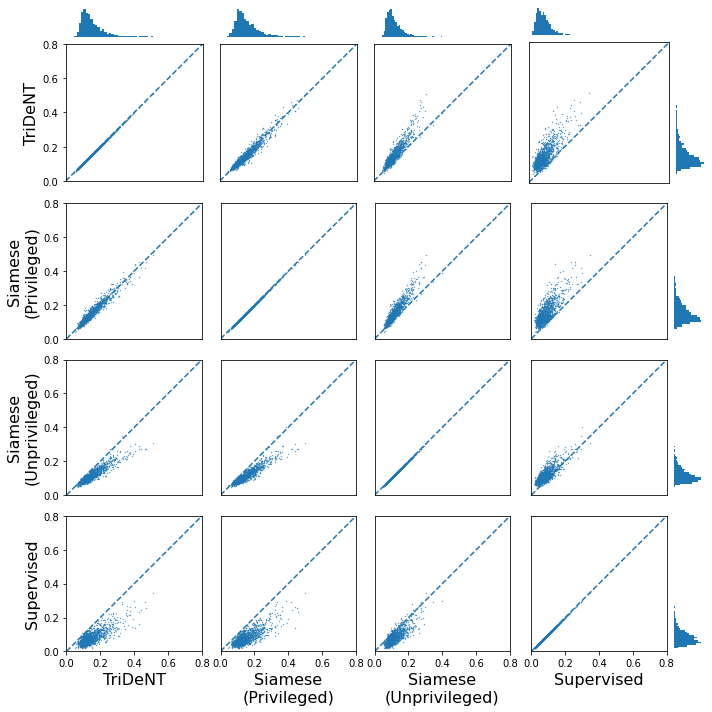}
    \caption{Human ALS-ST dataset - Greatest absolute correlation between gene counts and elements of each model's representations. We compare the correlation strength for the expression of each gene, with histograms appended to illustrate the distribution for each model. Dashed line is the identity. Above the line implies a greater correlation for the model listed on the y-axis, while below the line implies a greater correlation for the model listed on the x-axis.}
    \label{fig:counts-correlation-plot-human}
\end{figure*}

\clearpage

\begin{figure*}
    \centering
    \includegraphics[width=\textwidth]{fig/wikipathways-heatmap.pdf}
    \caption{WikiPathways-2019-Mouse geneset enrichment analysis heatmap. Representations were clustered using Leiden clustering \cite{traag2019louvain} with 7 nearest neighbours, and counts $c$ were normalised to $10^4$ counts per spot, and scaled by the rule $c'=\log(1+c)$. Genes were ranked by the Wilcoxon rank-sum test, and genes with a minimum log2 fold change of 0.25 and maximum adjusted p-value of 0.05 are used for the following calculations. Overexpression analysis was then performed using Enrichr \cite{kuleshov2016enrichr} with a background of all genes used during training. The pathways dataset used was "WikiPathways 2019 Mouse" \cite{pico2008wikipathways}. Enrichment scores are plotted on a log10 colour scale, with clusters with no genes enriched for a given pathway rendered in black.}
    \label{fig:wiki-heatmap}
\end{figure*}

\clearpage

\begin{figure*}
    \centering
    \includegraphics[width=0.8\textwidth]{fig/kegg-heatmap.pdf}
    \caption{KEGG-2019-Mouse geneset enrichment analysis heatmap. As in Figure \ref{fig:kegg-heatmap} but using the pathways from "KEGG 2019 Mouse" \cite{kanehisa2000kegg}.}
    \label{fig:kegg-heatmap}
\end{figure*}

\clearpage

\begin{figure*}[h]
    \includegraphics[width=\textwidth]{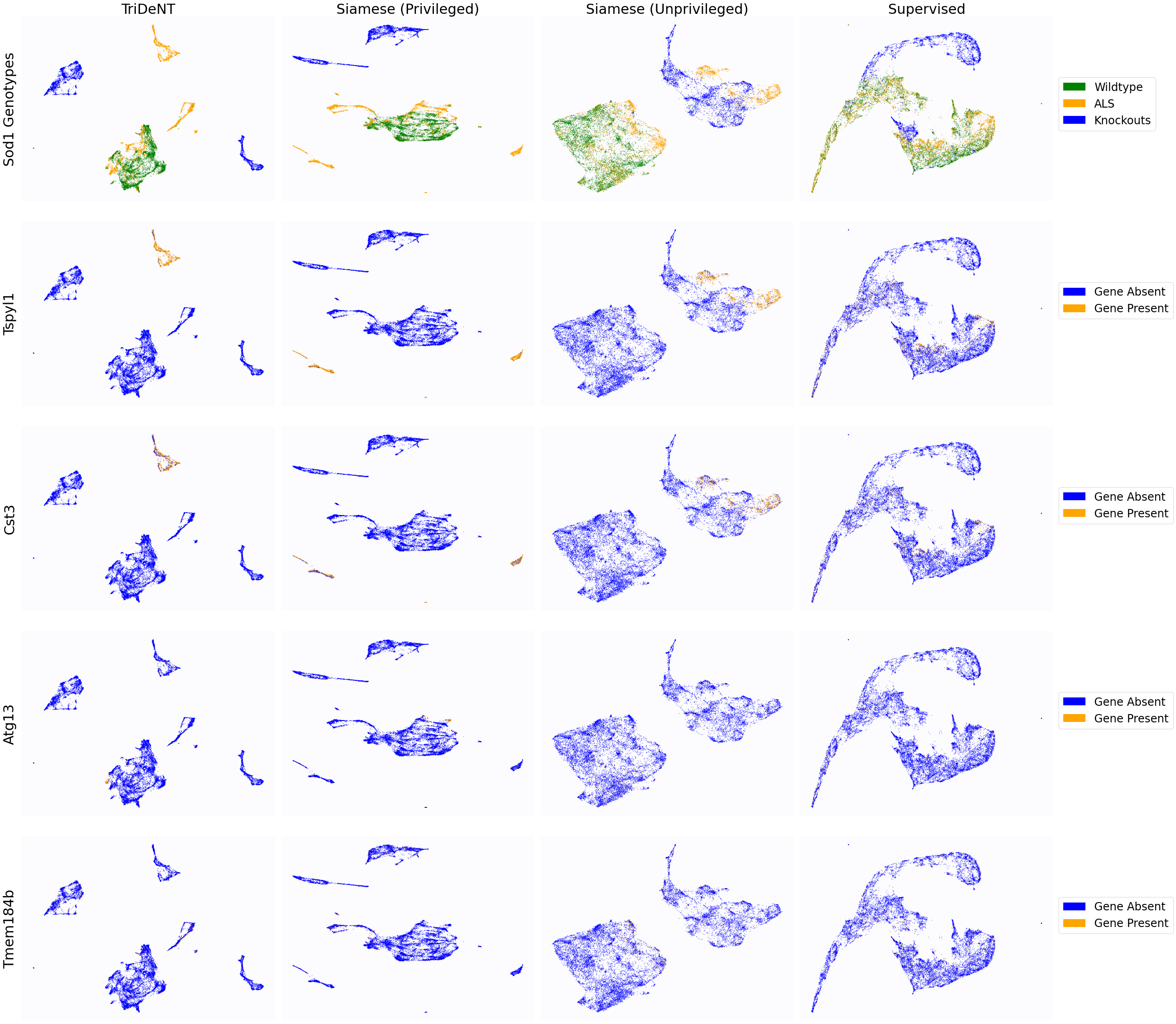}
    \caption{UMAP projections demonstrating the distribution in representation space of (a) mouse Sod1 genotypes, and expression of (b) Tspyl1, (c) Cst3, (d) Atg13, and (e) Tmem184b. Tspyl1 was selected as the gene with the highest absolute correlation (0.60) of its counts with the representations of the TriDeNT \Neptune ~model, while Cst3 was selected as the gene with the highest absolute correlation (0.51) of its counts with the representations of the unprivileged Siamese model. Note that the absolute correlation of Cst3 was higher (0.53) with the TriDeNT \Neptune ~model's representations. Atg13 was selected as the gene with the greatest difference in absolute correlation between TriDeNT \Neptune ~(0.39) and unprivileged Siamese (0.11). For comparison, Tmem184b was selected as the gene with the greatest difference in absolute correlation between unprivileged Siamese (0.16) and TriDeNT \Neptune ~(0.13).}
    \label{fig:als-st-umaps}
\end{figure*}

\clearpage

\begin{figure*}[h]
    \centering
    \includegraphics[width=\textwidth]{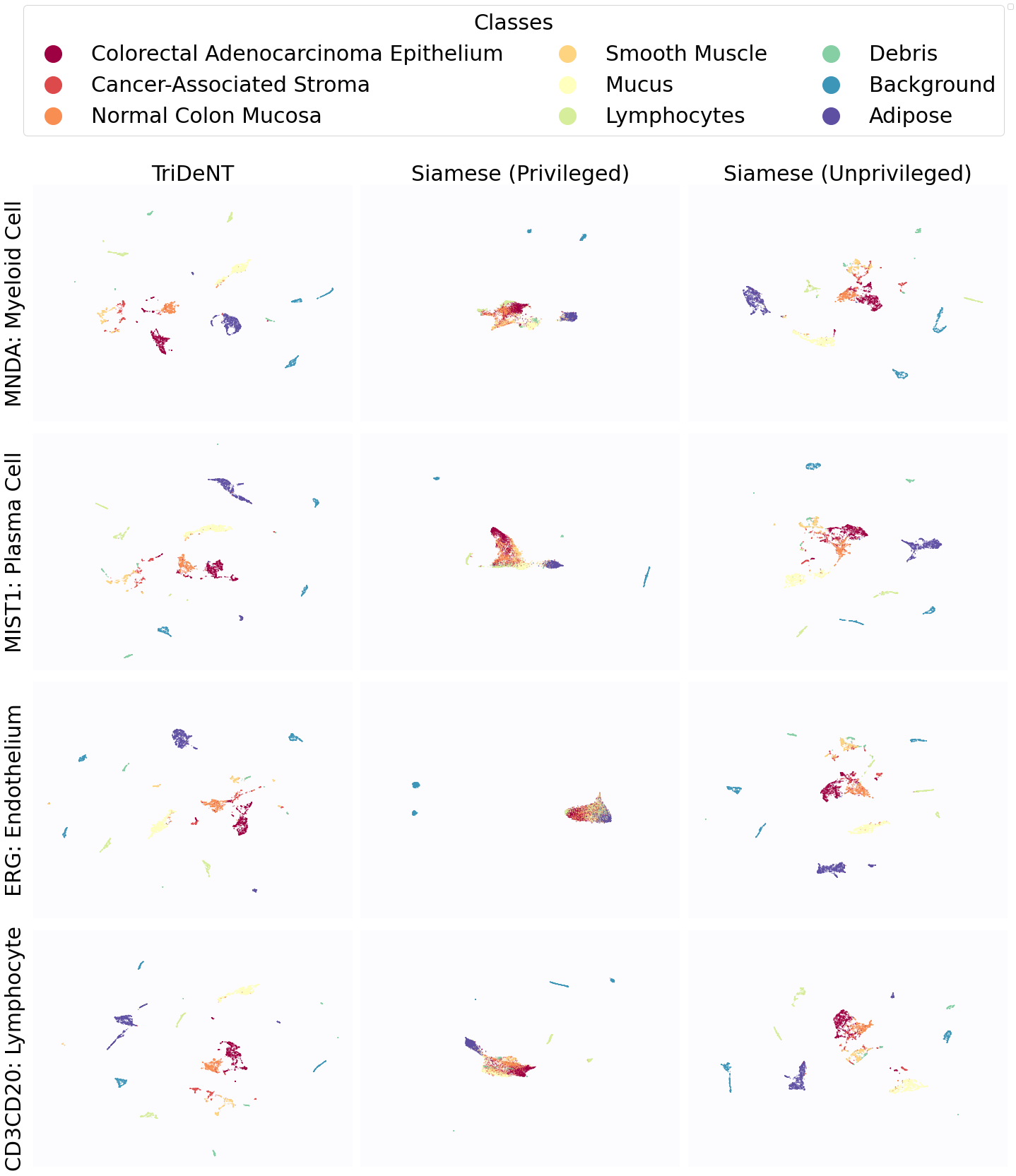}
    \caption{UMAPs for MNDA, MIST1, ERG and CD3CD20 evaluated on the NCT test set and coloured by tissue type.}
    \label{fig:supp_umaps_1}
\end{figure*}

\begin{figure*}[h]
    \centering
    \includegraphics[width=\textwidth]{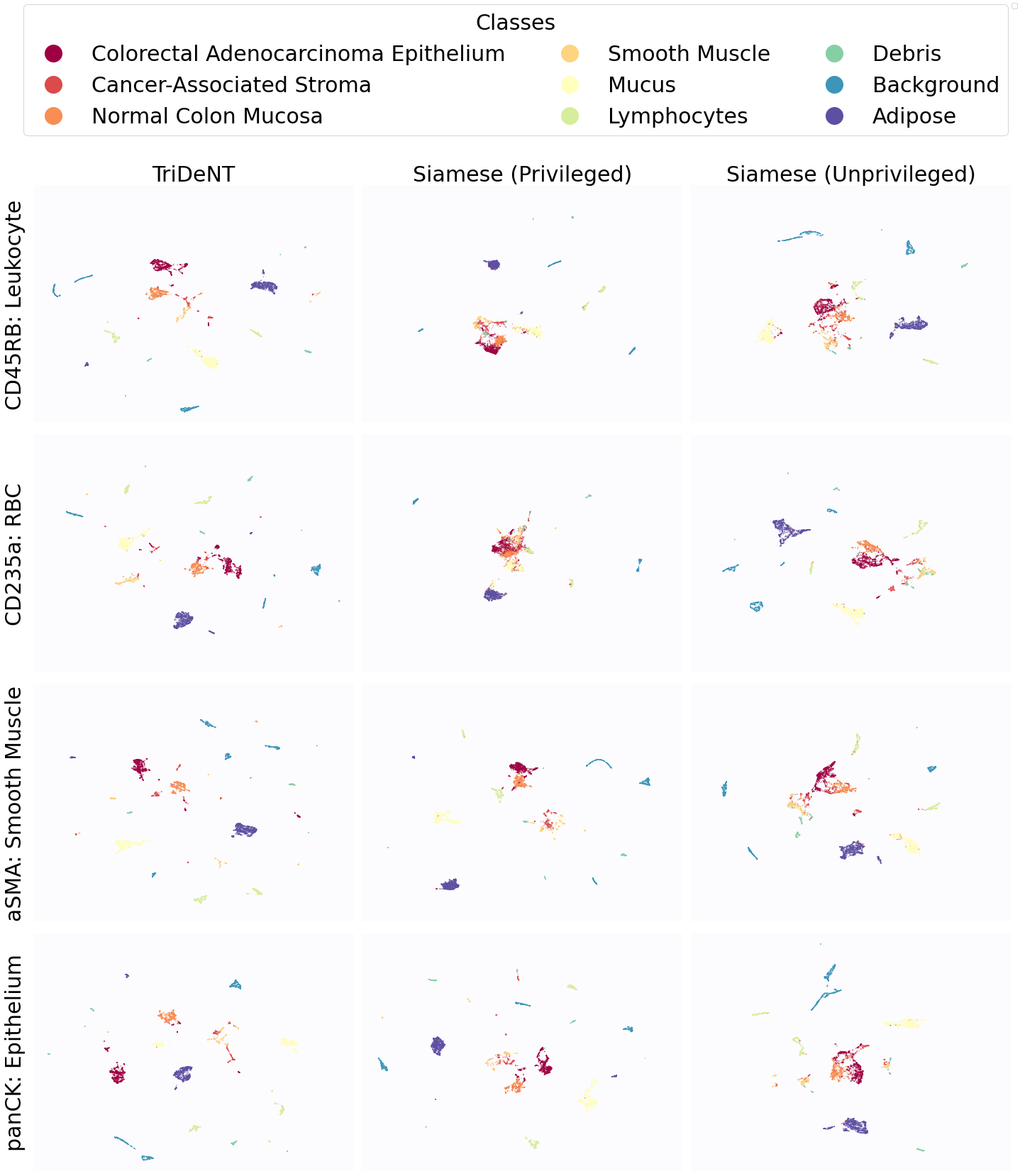}
    \caption{UMAPs for CD45RB, CD235a, $\alpha$SMA and pan-CK evaluated on the NCT test set and coloured by tissue type.}
    \label{fig:supp_umaps_2}
\end{figure*}

\foreach \M/\m/\s/\T in {VICReg/vicreg/vicreg/asymm3.png,InfoNCE/infonce/simclr/asymm3.png}{
\foreach \N/\n/\J in {MNDA_MyeloidCell/mnda/MNDA,MIST1_PlasmaCell/mist1/MIST1,ERG_Endothelium/erg/ERG,CD235a_RBC/cd235a/CD235a,CD45RB_Leukocyte/cd45rb/CD45RB,CD3CD20_Lymphocyte/cd3cd20/CD3CD20,aSMA_SmoothMuscle/asma/$\alpha$SMA,panCK_Epithelium/panCK/pan-CK}{
\clearpage
\begin{figure*}[h]
    \centering
    \begin{subfigure}{0.49\textwidth}
       \includegraphics[width=\textwidth, trim={1cm 1cm 1cm 1cm}, clip]{fig/new_gradcam_figures/\s/\N-he.png}
        \caption{H\&E}
        \label{fig:supp-grad-\n-he-\m}
    \end{subfigure}
    ~
    \centering
    \begin{subfigure}{0.49\textwidth}
        \includegraphics[width=\textwidth, trim={1cm 1cm 1cm 1cm}, clip]{fig/new_gradcam_figures/\s/\N-\T}
        \caption{TriDeNT \Neptune}
        \label{fig:supp-grad-\n-trident-\m}
    \end{subfigure}

    \begin{subfigure}{0.49\textwidth}
       \includegraphics[width=\textwidth, trim={1cm 1cm 1cm 1cm}, clip]{fig/new_gradcam_figures/\s/\N-symm2.png}
        \caption{Unprivileged Siamese}
        \label{fig:supp-grad-\n-symm-\m} 
    \end{subfigure}
    ~
    \centering
    \begin{subfigure}{0.49\textwidth}
        \includegraphics[width=\textwidth, trim={1cm 1cm 1cm 1cm}, clip]{fig/new_gradcam_figures/\s/\N-asymm2.png}
        \caption{Privileged Siamese}
        \label{fig:supp-grad-\n-asymm-\m}
    \end{subfigure}
    \caption{\J ~GradCAM examples for models with \M ~loss. (a) H\&E patches used as input, (b) TriDeNT \Neptune, (c) Unprivileged Siamese, and (d) Privileged Siamese.}
    \label{fig:supp-grad-\n-\m}
\end{figure*}
}
}

% \printbibliography[title=Supplementary References]

\end{document}